%% file: main.tex
\documentclass[10pt,twocolumn,letterpaper]{article}

\usepackage{cvpr}              % To produce the CAMERA-READY version
\input{preamble}
\definecolor{cvprblue}{rgb}{0.21,0.49,0.74}
\usepackage[pagebackref,breaklinks,colorlinks,allcolors=cvprblue]{hyperref}
\usepackage{subcaption}
\usepackage{multirow}
\usepackage{adjustbox}
\usepackage{wrapfig}
\usepackage{amsmath}
\usepackage{bm}
\usepackage{booktabs}
\usepackage{tcolorbox}
\usepackage{xcolor}
\usepackage{siunitx}
\usepackage{float}
\usepackage[table]{xcolor}
\usepackage{algorithm}
\usepackage{algorithmic}
\def\paperID{1177} % *** Enter the Paper ID here
\def\confName{CVPR}
\def\confYear{2026}

\title{Mind the Discriminability Trap in Source-Free Cross-domain Few-shot Learning}

\author
  {Zhenyu Zhang$^{1}$,\quad Yixiong Zou$^{1}$\thanks{Corresponding author.},\quad Yuhua Li$^{1}$\footnotemark[1],\quad Ruixuan Li$^{1}$,\quad Guangyao Chen$^{2}$\\ 
  $^{1}$ School of Computer Science and Technology, Huazhong University of Science and Technology \\
  $^{2}$ National Key Laboratory for Multimedia Information Processing,  Peking University\\ 
  {\tt\small \{d202481641, yixiongz, idcliyuhua, rxli\}@hust.edu.cn, gy.chen@pku.edu.cn}
  }
\begin{document}
\maketitle
\input{./Tex/00_abs_yxz}
\input{./Tex/01_intro_new_yxz}

\input{./Tex/0x_related_yxz}
\input{./Tex/02_preliminary_yxz}

\input{./Tex/03_analyze_yxz_Cpic_yxz}
\input{./Tex/04_method_yxz}
\input{./Tex/05_exp}
\input{./Tex/06_conclusion}

\section{Acknowledgments}
This work is supported by the National Key Research and Development Program of China under grant 2024YFC3307900; the National Natural Science Foundation of China under grants 62436003, 62206102, 62376103, 62302184, and 62402015; the Major Science and Technology Project of Hubei Province under grant 2025BAB011 and 2024BAA008; the Hubei Science and Technology Talent Service Project under grant 2024DJC078; Ant Group through the CCF–Ant Research Fund; the Postdoctoral Fellowship Program of the China Postdoctoral Science Foundation under grant GZB20230024; and the China Postdoctoral Science Foundation under grant 2024M750100. Computations were performed on the HPC Platform of Huazhong University of Science and Technology.
\newpage
{
    \small
    \bibliographystyle{ieeenat_fullname}
    \bibliography{main}
}
\onecolumn
\noindent{\Large \textbf{Appendix}}
\tableofcontents
\input{./Tex/07_supp}

% WARNING: do not forget to delete the supplementary pages from your submission 
% \input{sec/X_suppl}

\end{document}

%% file: Tex/00_abs_yxz.tex
\begin{abstract}

Source-Free Cross-Domain Few-Shot Learning (SF-CDFSL) focuses on fine-tuning with limited training data from target domains (e.g., medical or satellite images), where Vision-Language Models (VLMs) such as CLIP and SigLIP have shown promising results. Current works in traditional visual models suggest that improving visual discriminability enhances performance. However, in VLM-based SF-CDFSL tasks, we find that \textbf{strengthening visual-modal discriminability actually suppresses VLMs’ performance}. 
In this paper, we aim to delve into this phenomenon for an interpretation and a solution.
By both theoretical and experimental proofs, our study reveals that fine-tuning with the typical cross-entropy loss ($\mathcal{L}_{\mathrm{vlm}}$) inherently includes a visual learning part and a cross-modal learning part, where the cross-modal part is crucial for rectifying the heavily disrupted modality misalignment in SF-CDFSL.
However, we find that the visual learning essentially acts as a shortcut that encourages the model to reduce $\mathcal{L}_{\mathrm{vlm}}$ without considering the cross-modal part, therefore hindering the cross-modal alignment and harming the performance.
Based on this interpretation, we further propose an approach to address this problem: first, we perturb the visual learning to guide the model to focus on the cross-modal alignment. Then, we use the visual-text semantic relationships to gradually align the visual and textual modalities during the fine-tuning. 
Extensive experiments on various settings, backbones (CLIP, SigLip, PE-Core), and tasks (4 CDFSL datasets and 11 FSL datasets) show that we consistently set new state-of-the-art results. Code is available at https://github.com/zhenyuZ-HUST/CVPR26-Mind-the-Discriminability-Trap.

\end{abstract}

%% file: Tex/01_intro_new_yxz.tex
%%%\vspace{-0.4cm}
\section{Introduction}

\begin{figure*}[t]
%%\vspace{-0.5cm}
\medskip
\subfloat[c][]{
\begin{minipage}[b]{.29\linewidth}
    \centering
    %\subfloat[b][]{
    \label{fig:intro_a}\includegraphics[width=1\linewidth]{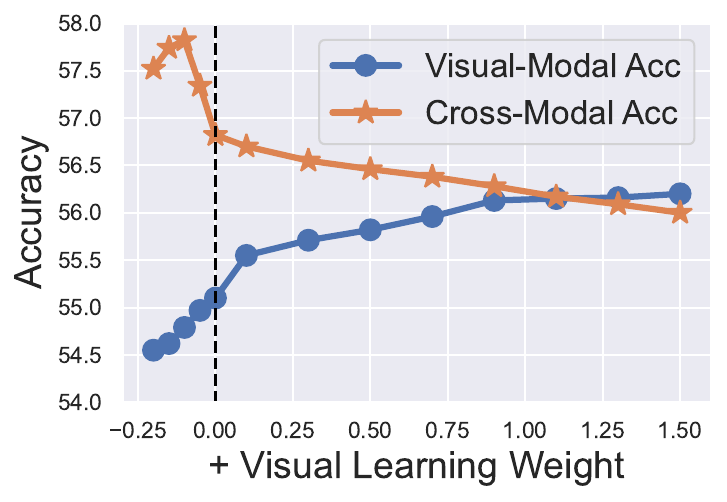} \\
\end{minipage}}
\subfloat[c][]{
\begin{minipage}[b]{.30\linewidth}
    \centering
\label{fig:intro_b}\includegraphics[width=1\linewidth]{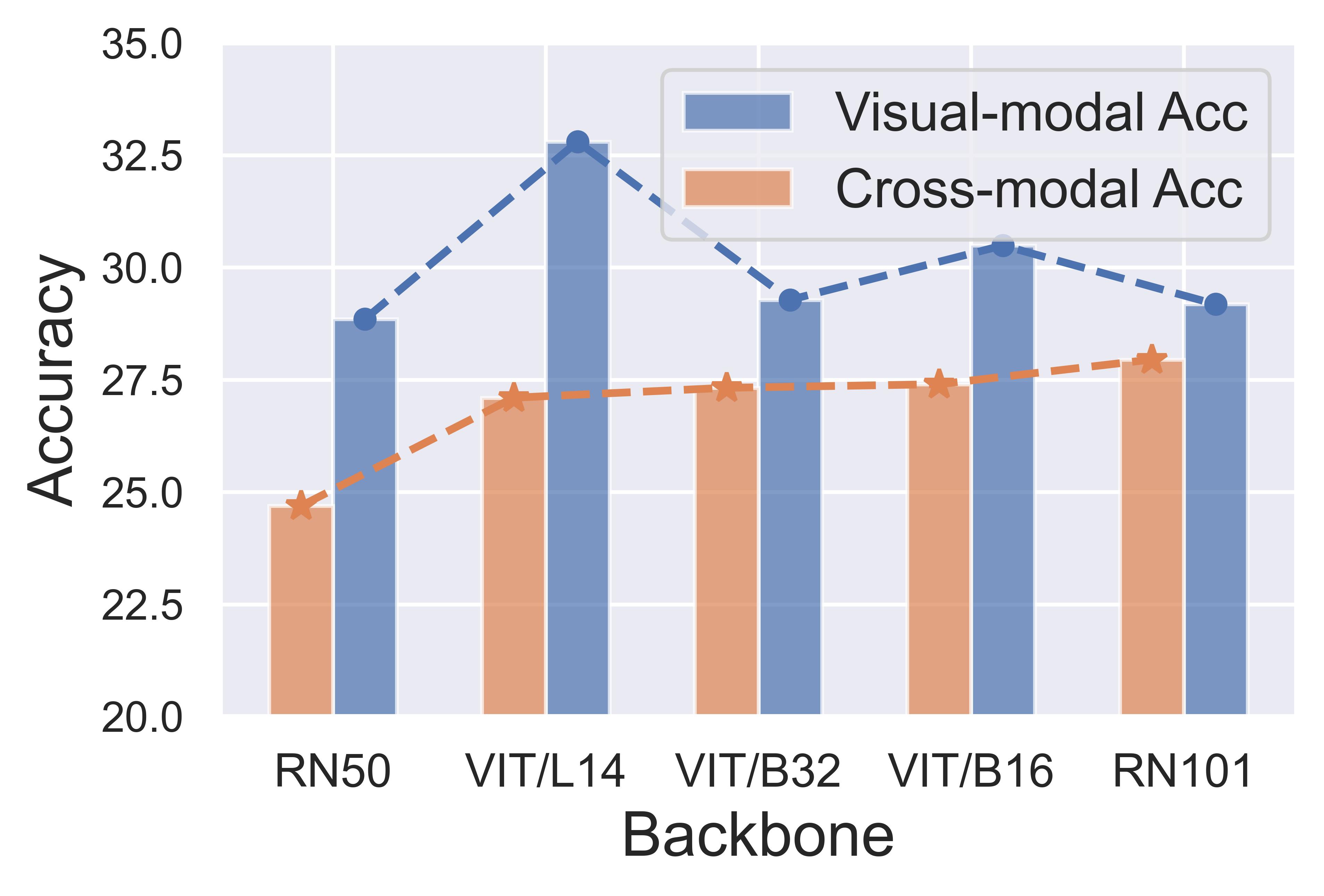}
    %}
\end{minipage}}
\hspace{0.01\textwidth}
\subfloat[c][Raw Fine-tuning]{
\begin{minipage}[b]{.2\linewidth}
    \centering
    %\subfloat[c][]{
\label{fig:intro_c_base}\includegraphics[width=1\linewidth]{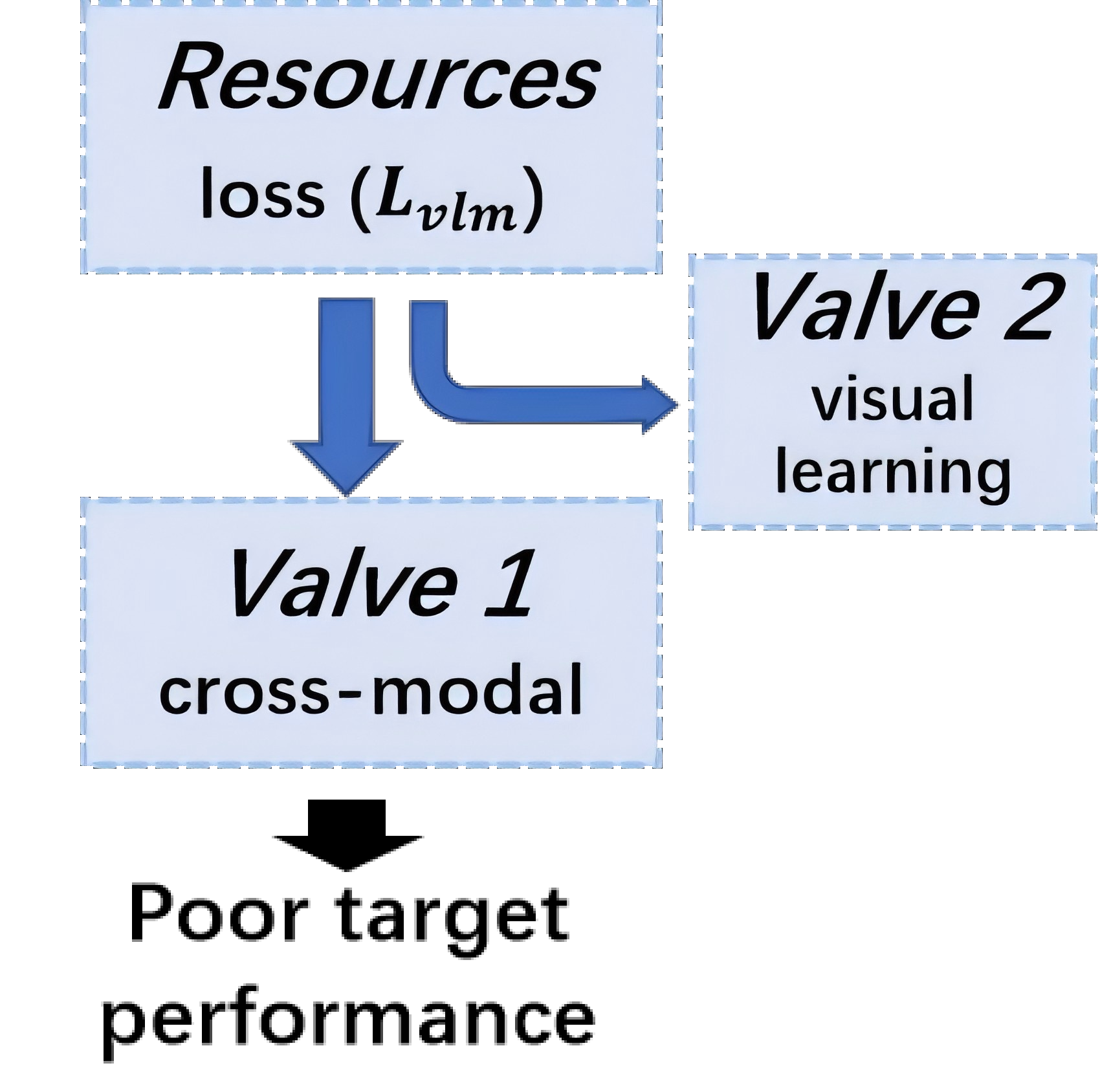}\\
    %}
\end{minipage}}
\subfloat[c][Ours]{
\begin{minipage}[b]{.2\linewidth}
    \centering
    %\subfloat[c][]{
    \label{fig:intro_d_ours}\includegraphics[width=1\linewidth]{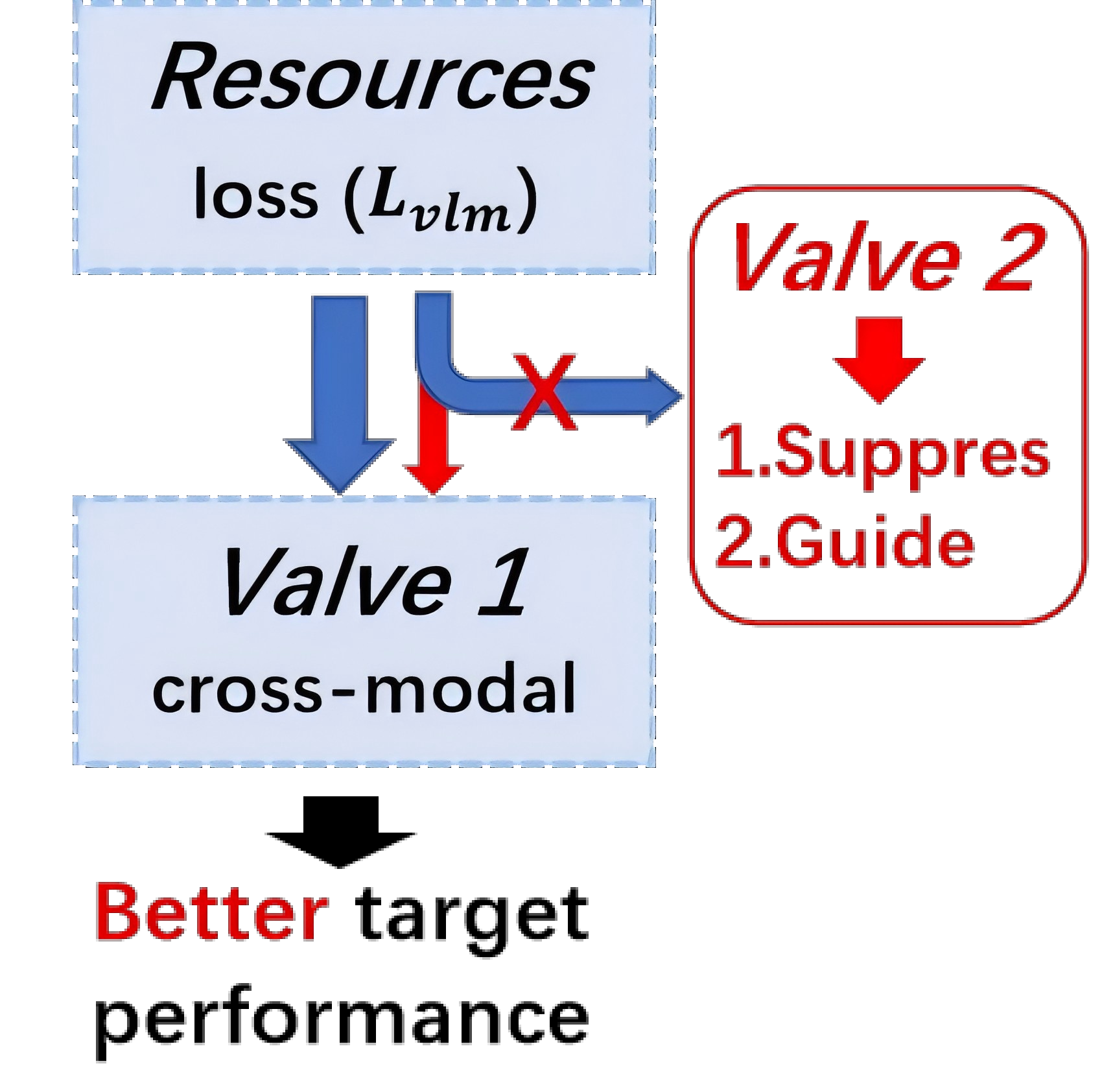}\\
\end{minipage}}
%\vspace{-0.1cm} 
\caption{
    (a) Unlike traditional visual models, we find that enhancing visual learning consistently decreases the performance of the VLM-based CDFSL, although the visual-modal accuracy increases. Appropriately inhibiting visual learning improves cross-modal performance. (b) We also find that this phenomenon widely exists in cross-domain scenarios, where the best-performing zero-shot CLIP model does not necessarily extract the best visual features. (c) In this paper, we explore the phenomenon for an interpretation, finding that fine-tuning involves two learning directions: the visual and the cross-modal ones. We interpret visual learning as a shortcut that reduces the VLM's classification loss but disrupts cross-modal learning. (d) Based on this, we propose a method to suppress and guide visual learning, enhancing the model's cross-modal learning.
    %(a) Enhancing discriminability disrupts the true semantic relationships between categories. (b) Gradual enhancement of feature discriminability learning (for intra-modal loss, refer to Section 2.3) leads to a consistent decline in the fine-tuned performance of CLIP on four CDFSL datasets. (c) The accuracy of different CLIP models on the ISIC dataset under two classification methods: using CLIP (CLIP mode) and using only the visual features extracted by CLIP (Intra-modal mode). The models are ranked from left to right by their accuracy in CLIP mode, from lowest to highest.
}
\label{fig:intro}
%\vspace{-0.1cm}
\end{figure*}

Cross-domain few-shot learning (CDFSL)~\cite{guo2020broader,zouattention,zou2025closer,xu2024step,fu2023styleadv} aims to address target-domain tasks (such as medical or satellite image recognition) with limited data by leveraging knowledge transferred from large-scale source-domain datasets (such as ImageNet). Traditional CDFSL works 
%tackle the target domain few-shot learning (FSL) tasks using vast amounts of data and carefully designed training or adaptation strategies from alternative (source) domains. However, these methods require source data to be accessible and rely on specifically designed training strategies. 
focus on the source-domain training to obtain a generalizable model for various target domains.
However, in real-world scenarios, issues such as computational burden and data privacy make source-domain training infeasible. 
%In such contexts where source domain data is unavailable, 
%some studies have shown that pre-trained models with strong generalization capabilities can effectively address 
As a result, the source-free CDFSL task (SF-CDFSL)~\cite{radford2021learning,yazdanpanah2022visual,zhuo2024prompt,xu2024step} is proposed to focus solely on the finetuning strategy on the scarce training data of target domains. 
Currently, VLMs, such as CLIP~\cite{radford2021learning}, SigLIP~\cite{tschannen2025siglip}, and PE-Core~\cite{bolya2025perception}, have achieved promising results on in-domain few-shot learning tasks, where the classification is carried out by computing the similarity between image and text features.

% Large models, such as LAVVA \cite{} and CLIP \cite{}, often incorporate multiple modality branches. As multimodal models, these systems can transfer to downstream cross-domain tasks in a zero-shot manner. Specifically, they perform classification by calculating the similarity between image and text features. In this process, the alignment between modalities—specifically, how well image features correspond to their textual category descriptions—directly influences the model's performance on downstream tasks.

In traditional visual models, it is generally believed that the more discriminative the visual features extracted by the model, the better the classification performance~\cite{mistretta2025cross,li2020prototypical,wu2018unsupervised,yang2025clip} would be. However, in VLMs, we discover an opposite phenomenon: 1) enhancing the visual learning actually suppresses the VLMs' performance on SF-CDFSL tasks, as in Fig.~\ref{fig:intro_a} (Visual learning weight $>$ 0), and 2) \textbf{appropriately inhibiting the visual learning can achieve better fine-tuning performance on SF-CDFSL tasks}, as in Figure~\ref{fig:intro_a} (Visual learning weight $<$ 0). Moreover, we also observe this phenomenon in zero-shot settings, where the best-performing CLIP model does not necessarily extract the most discriminative visual features (Fig.~\ref{fig:intro_b}).

In this paper, we aim to explore this phenomenon for an interpretation. Specifically, typical few-shot fine-tuning methods~\cite{li2025logits,huang2024lp++,tang2024amu,zhang2021tip,khattak2023maple,zanella2024low} take the cross-entropy-based loss ($\mathcal{L}_{\mathrm{vlm}}$, Eq.~\ref{eq:vlm_loss}) between class texts and each image, which we demonstrate through mathematical derivation that involves a visual learning part and a cross-modal learning part. 
Recent studies \cite{shangguan2025cross,jing2020cross,ji2022information,wang2022toward} indicate that in cross-domain scenarios, VLMs' visual-text (cross-modal) alignment is heavily disrupted, making the cross-modal learning necessary in cross-domain few-shot fine-tuning. 
However, as both theoretically and empirically proved, we find that the visual learning actually acts as a shortcut of $\mathcal{L}_{\mathrm{vlm}}$ that hinders the cross-modal learning, harming the performance. 
For an intuitive understanding, we consider the fine-tuning with $\mathcal{L}_{\mathrm{vlm}}$ as \textbf{a two-valve bucket draining process} (Fig.~\ref{fig:intro_c_base}), 
where $\mathcal{L}_{\mathrm{vlm}}$ is the water (resources) to drain, and the visual learning and the cross-modal learning act as two valves, jointly controlling the draining process. 
The more resources are drained from the visual-learning-valve, the fewer resources are drained from the cross-modal-valve, therefore harming the cross-modal learning.

% In this paper, we combine theoretical derivation with multiple experimental designs to prove our hypothesis: \textbf{during the fine-tuning process of VLM for downstream SF-CDFSL tasks, visual discriminative learning diverts some of the model's learning resources, thus inhibiting the model's ability to learn better textual-visual cross-modal relationships for improved performance}. 
Based on these interpretations, we first propose the Suppressing Visual Learning (SVL) module to suppress visual learning during the fine-tuning process, guiding it to focus more on cross-modal alignment. Subsequently, we introduce the Relationship Alignment (RA) module to steer the direction of visual learning, promoting cross-modal alignment by visual-text semantic relationships. Our method is simple and effective, delivering significant performance improvements across multiple tasks with nearly zero additional overhead. Additionally, it is plug-and-play, requiring only a few lines of code to implement. Extensive experiments across various settings, backbones, and tasks demonstrate our state-of-the-art performance and generality.
%Additionally, in other task such as few-shot fine-tuning, introducing our methods also leads to performance improvements, proving the effectiveness and generality of our approach. 

In summary, we make the following contributions:
%%%\vspace{-0.05cm}

% $\bullet$ We identified that the VLM fine-tuning process inherently includes a visual discriminative learning branch, an aspect overlooked and underexplored in previous research.

% \item To the best of our knowledge, we are the first to identify that the CLIP model exhibits misalignment between modalities in CDFSL scenarios, and enhancing visual-modal discriminability exacerbates this misalignment. 
$\bullet$  We are the first to reveal that during the cross-entropy-based fine-tuning process, the visual learning acts as a shortcut that disrupts the important cross-modal alignment. 

$\bullet$  We propose two simple, effective, and plug-and-play methods to constrain and correctly guide visual learning, steering the model fine-tuning process to focus on the more critical cross-modal alignment.

$\bullet$  We applied our methods to various VLMs and achieved significant improvements in classification performance on 4 CDFSL datasets and 11 few-shot learning datasets, achieving new state-of-the-art performance.

% \begin{figure*}[!h]
% \centering
% \subfloat[CLIP]{
% \begin{minipage}[b]{0.31\linewidth}
% \begin{adjustbox}{max width=1\linewidth}
% \label{fig:intro_a}\includegraphics[]{./Figs/intro/both_modal_acc_64_both.jpg}
% \end{adjustbox}
% \end{minipage}
% } % 不空行表示不换行
% \subfloat[CLIP]{
% \begin{minipage}[b]{0.31\linewidth}
% \begin{adjustbox}{max width=1\linewidth}
% \label{fig:intro_b}\includegraphics[]{./Figs/intro/both_modal_acc_64_both.jpg}
% \end{adjustbox}
% \end{minipage}
% } % 不空行表示不换行
% \subfloat[Ours]{
% \begin{minipage}[b]{0.23\linewidth}
% \label{fig:intro_lost_base}\includegraphics[width=1\linewidth]{./Figs/intro/intro3_base.png}
% \end{minipage}

% \subfloat[Ours]{
% \begin{minipage}[b]{0.23\linewidth}
% \label{fig:intro_lost_base}\includegraphics[width=1\linewidth]{./Figs/intro/intro3_ours.png}
% \end{minipage}
% }
% %%\vspace{-0.3cm}
% \caption{(a) CLIP has two branches: a visual encoder and a text encoder. However, we find that removing certain layers of the text encoder can significantly enhance its performance in SF-CDFSL tasks. (b) Performance of 5-way 1-shot fine-tuned CLIP after removing the i-th layer (x-axis) of the text encoder. The horizontal dashed line represents the performance achieved using the full text encoder. Masking certain layers results in better performance. (c) After applying our method, the optimal performance is achieved using the full text encoder (dashed line), indicating that the lost layer no longer exists.}
% \label{fig:intro}
% %%\vspace{-0.3cm}
% \end{figure*}

%% file: Tex/0x_related_yxz.tex
%\vspace{-0.15cm}
\section{Related Work}
%\vspace{-0.15cm}
%\subsection{Source-Free Cross-domain Few shot Learning}
 \textbf{Cross-Domain Few-Shot Learning (CDFSL)} aims to train a model on a source domain that can generalize effectively to a target domain with limited examples. Existing methods are typically categorized into two types: meta-learning-based approaches~\cite{fu2022wave,guo2020broader,hu2022adversarial,wang2021cross} and transfer learning-based approaches~\cite{guo2020broader,liang2021boosting,zhou2023revisiting,zou2024flatten,zou2025closer,zouattention}. Source-Free CDFSL (SF-CDFSL) introduces a stronger constraint by making source domain data inaccessible~\cite{yazdanpanah2022visual,zhuo2024prompt,xu2024step}. However, the influence of visual learning on VLM-based SF-CDFSL tasks remains underexplored.
 
%\subsection{Fine-tuning in CLIP}
%\vspace{0.1cm}
\noindent\textbf{In Fine-tuning}, a common strategy is parameter-efficient fine-tuning (PEFT), which uses only a few samples from the target task.  PEFT methods can be grouped into three main types: prompt learning~\cite{zhou2022learning,zhou2022conditional,khattak2023maple,chen2022plot,zhu2023prompt,khattak2023self,yao2023visual,lu2023beyond,li2024promptkd}, adapters~\cite{gao2024clip,zhang2022tip,li2025logits,huang2024lp++,tang2024amu}, and LoRA (and its variants)~\cite{hu2021lora,zanella2024low}. And,~\cite{mistretta2025cross} use optimization-based modality inversion techniques to map representations from input modality to the complementary one.

%\subsection{Modality Gap and Misalignment}
%\vspace{0.1cm}
\noindent\textbf{Modality gap and misalignment} is currently a key area of focus in multimodal research. \cite{liang2022mind} was the first to identify the existence of a modality gap in multimodal models. Subsequent studies\cite{ji2022information,hao2023uncertainty} further analyzed this gap in multimodal models.~\cite{tian2025mind} pointed out a similar gap between prototypes and instances. Some research has noted that the alignment of these modalities is disrupted in cross-domain scenarios~\cite{shangguan2025cross,jing2020cross,ji2022information,wang2022toward}. These existing works identified the issue of modality misalignment in cross-domain scenarios and considered fine-tuning an effective method for realignment. However, our work demonstrates that fine-tuning alone is insufficient for effective realignment in cross-domain scenarios, especially for the large domain gaps in the CDFSL task, e.g., general domains vs. medical domains, which we analyze from the perspective of shortcut learning~\cite{geirhos2020shortcut,song2024shortcut,yuan2024llms,ma2023rectify} and handle in this paper.

%% file: Tex/02_preliminary_yxz.tex
%\vspace{-0.15cm}
\section{Preliminary}
%\vspace{-0.15cm}

% (a) (b): Cosine similarity heatmaps of visual features before and after fine-tuning with Eq.~\ref{eq:cross_loss}. After fine-tuning, the visual features demonstrate strong discriminability, 
% %with intra-class similarities approaching 1 and inter-class similarities decreasing, 
% even without visual-modal loss. 

\textbf{Source-Free Cross-Domain Few-Shot Learning}:
Given a target domain dataset $D_T$, an episode $P = \{(S, Q), Y\}$ is randomly sampled for meta-testing. This meta-testing is formulated as an $N$-way $K$-shot problem. Specifically, for each episode $P_i$ from $D_T$, $N$ classes are sampled with $K$ labeled images to form the support set $S$, and the same $N$ classes with $M$ different images form the query set $Q$. The label set for these $N$ classes is $Y = \{ c_i \}_{i=1}^N$. The support set $S$ is for training, while the query set $Q$ is for testing. 

%\vspace{0.1cm}
\noindent\textbf{Fine-tuning in VLMs}:
In classification tasks using VLMs, a textual description, known as a prompt, is generated for each class, such as ``a photo of a cat". Let \( r_k \) denote the tokenized prompt for the \( k \)-th class. The language encoder \( \theta_t \) processes \( r_k \) to produce the normalized text embedding \( t_k = \theta_t(r_k) \). Similarly, each image \( x_i \) is processed by the visual encoder \( \theta_v \) to obtain the normalized visual embedding \( f_i = \theta_v(x_i) \). The classification probability is: 
{\small
\begin{equation}\label{eq:cross_sim_and_p}
\setlength{\abovedisplayskip}{3pt}
\setlength{\belowdisplayskip}{3pt}
s_{i, j} = Sim(f_i,t_j), \quad  p_{i, j} = \frac{\exp\left(s_{i, j} / \tau\right)}{\sum_{j=1}^K \exp\left(s_{i, j} / \tau\right)},
\end{equation}}where $s_{i, j}$ represents the cosine similarity between image sample $i$ and text prompt $j$, and $\tau$ is the temperature coefficient. 
%In CLIP~\cite{radford2021learning}, the value of $\tau$ is set to 0.01.
%Many works explored fine-tuning in VLMs~\cite{zhou2022learning,khattak2023maple,hu2021lora,zanella2024low}. Despite differences in learning strategies and parameter update mechanisms, most methods fine-tune by minimizing the cross-entropy loss, which can be expressed as:
The typical cross-entropy-based fine-tuning loss~\cite{zhou2022learning,khattak2023maple,hu2021lora,zanella2024low} is
{\small
\begin{equation}\label{eq:vlm_loss}
\setlength{\abovedisplayskip}{3pt}
\setlength{\belowdisplayskip}{3pt}
\mathcal{L}_{\mathrm{vlm}}=-\frac{1}{N} \sum_i \log \frac{\exp \left(s_{i, i} / \tau\right)}{\sum_j \exp \left(s_{i, j} / \tau\right)}.
\end{equation}}

%%\vspace{-0.1cm}
\noindent\textbf{Visual Learning} trains a classifier to perform classification~\cite{chen2019closer} of visual features as follows:
{\small
\begin{equation}\label{eq:intra_sim}
\setlength{\abovedisplayskip}{3pt}
\setlength{\belowdisplayskip}{3pt}
p_{i, j}=\frac{e^{f_i w_j / \tau}}{\sum_{j=1}^K e^{f_i w_j / \tau}}, \quad \mathcal{L}_{\mathrm{v}}=-\frac{1}{N} \sum_i \log \frac{e^{f_i w_i / \tau}}{\sum_{j=1}^K e^{f_i w_j / \tau}},
\end{equation}}where $w$ represents the weights of the classifier. 
%The cross-entropy loss is used to train the model:
 % When performing this process on a pre-trained feature extractor, the learning rate for the classifier is typically set higher, while the feature extractor uses a lower learning rate. 
Upon completion of visual learning, the feature extractor can extract discriminative image features, and the classifier weights can be viewed as the centroid of each category.

%% file: Tex/03_analyze_yxz_Cpic_yxz.tex
\vspace{-0.2cm}
\section{Analysis of the Discriminability Trap}
\vspace{-0.1cm}
In VLMs, there are two types of relationships: cross-modal relationships (visual-text relationships) and intra-modal relationships (specifically, visual-modal relationships, which are the focus of this study). Among these, cross-modal relationships are more crucial as they directly determine the performance of VLMs~\cite{shangguan2025cross,ji2022information,wang2022toward}. 

During visual model classification tasks, the model is trained through visual learning (Eq.~\ref{eq:intra_sim}) to extract discriminative features. In contrast, we propose an opposite learning direction, which we call Anti-Visual Learning :

\begin{tcolorbox}[colback=gray!10, colframe=gray!30, title=\textcolor{black}{\textbf{Anti-Visual Learning}}]
% \textbf{Target:} The learning of visual-text relationships, specifically the image-text classification tasks.

We replace $w$ in Eq.~\ref{eq:intra_sim} with \textbf{randomly selected samples'} visual features, forming a random classifier $w'$. Then Anti-Visual Loss ($\mathcal{L}_{\mathrm{ad}}$) is:
{\small
\begin{equation}
\label{eq:anti_visual_loss}
\setlength{\abovedisplayskip}{3pt}
\setlength{\belowdisplayskip}{3pt}
\mathcal{L}_{\mathrm{ad}}=-\frac{1}{N} \sum_i \log \frac{e^{f_i w^{'}_i / \tau}}{\sum_{j=1}^K e^{f_i w^{'}_j / \tau}}.
\end{equation}}This loss introduces perturbations, making the visual features indistinguishable and thereby suppressing visual learning.
\end{tcolorbox}

\subsection{Phenomenon}
\vspace{-0.1cm}

We incorporate $\mathcal{L}_{\mathrm{v}}$ and $\mathcal{L}_{\mathrm{ad}}$ into the fine-tuning of VLMs to control visual learning, either enhancing or inhibiting it. The results are shown in Fig.~\ref{fig:intro_a}. We find that visual learning (Eq.~\ref{eq:intra_sim}) improves visual-modal performance, but reduces cross-modal performance. In contrast, anti-visual learning (Eq.~\ref{eq:anti_visual_loss}) suppresses visual-modal performance but effectively enhances cross-modal performance. 
%\vspace{-0.1cm}

% \begin{figure*}[!t]
% %%\vspace{-1.2cm}
% \centering
% %\includegraphics[width=3in]{fig5}
% \subfloat[Delta Cos]{
% % 	\label{fig:disc_in_clip_a}\includegraphics[width=0.18\linewidth]{./Figs/disc_in_clip/zero_headmap.jpg}}
% % \subfloat[]{
% % 		\label{fig:disc_in_clip_b}\includegraphics[width=0.18\linewidth]{./Figs/disc_in_clip/after_headmap.jpg}}
% % \subfloat[]{
% \label{fig:clip_delta_cos}\includegraphics[width=0.25\linewidth]{./Figs/disc_in_clip/clip_deta_cos.pdf}\vspace{-0.05cm}} 
% \subfloat[CLIP]{
% \label{fig:pe_delta_cos}\includegraphics[width=0.25\linewidth]{./Figs/disc_in_clip/pe_deta_cos.pdf}\vspace{-0.05cm}} 
% \subfloat[CLIP]{
% \centering
% \label{fig:disc_loss_drop_clip}\includegraphics[width=0.25\linewidth]{./Figs/disc_in_clip/CLIP_dis_loss.pdf}\vspace{-0.05cm}} %\vspace{-0.2cm}
% \subfloat[PE-Core]{
% \centering
% \label{fig:disc_loss_drop_pe}\includegraphics[width=0.25\linewidth]{./Figs/disc_in_clip/PE_dis_loss.pdf}\vspace{-0.05cm}}
% \vspace{-0.35cm}
% \caption{(a)(b) When sample $i$ and sample $k$ belong to different classes, $\Delta \cos(\theta_{ik})$ in 5-way 1-shot fine-tuning is always less than 0. (b)(c) At any stage of fine-tuning (each epoch), visual learning can effectively reduce the loss value ($\mathcal{L}_{\mathrm{vlm}}$). }
% \vspace{-0.3cm}
% \label{fig:disc_ana}
% \end{figure*}

\begin{figure*}[!t]
%%\vspace{-1.2cm}
\centering
\subfloat[$\Delta \cos(\theta_{ik})$ in fine-tuning]{
\label{fig:delta_cos}
\vspace{-0.1cm}
\centering
\includegraphics[width=0.245\linewidth]{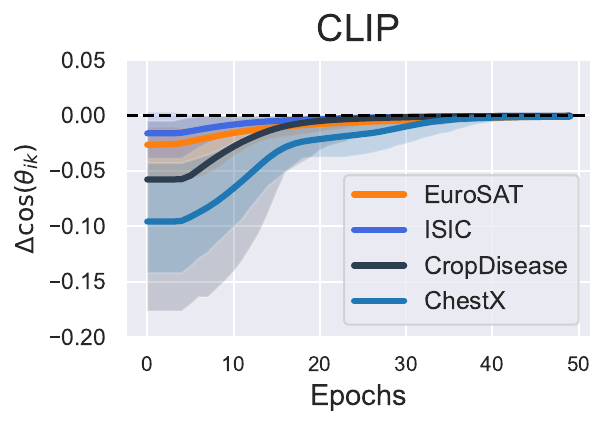}
\includegraphics[width=0.245\linewidth]{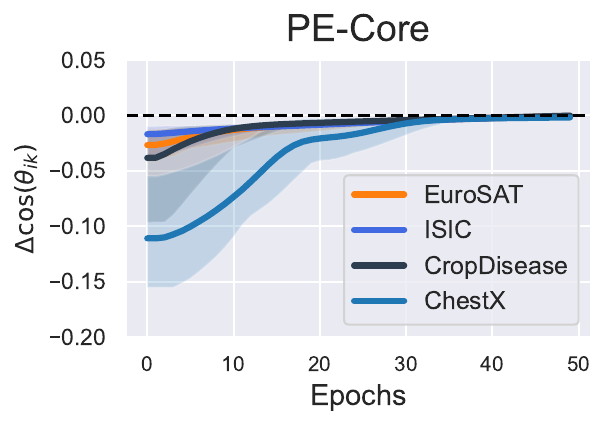}
} 
\medskip
\subfloat[Reduced loss by visual learning]{
\label{fig:disc_loss_drop}
\vspace{-0.1cm}
\centering
\includegraphics[width=0.25\linewidth]{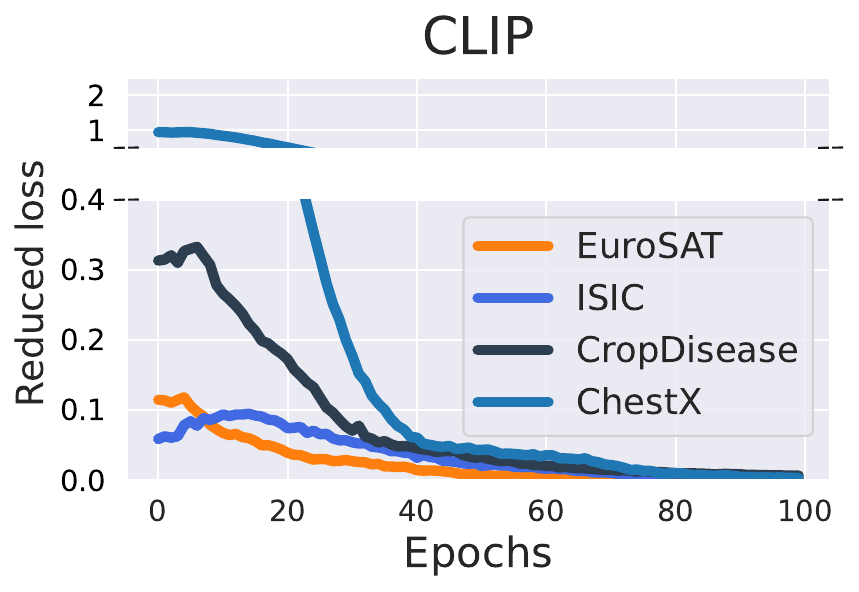} \includegraphics[width=0.25\linewidth]{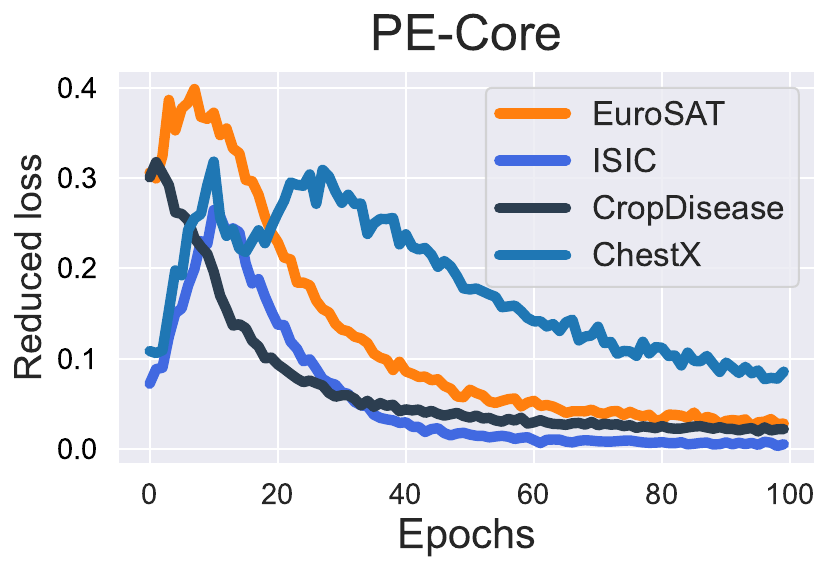}
}
\vspace{-0.3cm}
\caption{(a) When sample $i$ and sample $k$ belong to different classes, $\Delta \cos(\theta_{ik})$ in 5-way 1-shot fine-tuning is always less than 0. (b) At any stage of fine-tuning (each epoch), visual learning can effectively reduce the loss value ($\mathcal{L}_{\mathrm{vlm}}$). }
%\vspace{-0.3cm}
\label{fig:disc_ana}
\end{figure*}

\vspace{-0.1cm}
\subsection{Hypothesis}
\vspace{-0.1cm}
\subsubsection{Visual learning is part of the fine-tuning process.}
To understand the reason behind this phenomenon, we first investigate the role of visual learning during the fine-tuning of VLMs. We find that during the fine-tuning process, VLM models naturally engage in visual learning:
% Firstly, we plot the  Fig.~\ref{fig:disc_in_clip_a}, \ref{fig:disc_in_clip_b}, the cosine similarity map of image features before and after fine-tuning. The rows and columns represent the sample IDs. As observed, even without explicitly incorporating visual discriminative learning, the visual features, fine-tuned with cross-modal loss (Eq.~\ref{eq:cross_loss}), become highly discriminative. Furthermore, we validate this finding through mathematical derivations:
%\vspace{-0.2cm}
\begin{tcolorbox}[
  colback=purple!2!white, % 背景颜色
  colframe=purple, % 边框颜色为淡紫色
  boxrule=0.8pt, % 边框线条粗细
  arc=4pt, % 边框圆角半径
  boxsep=1pt, % 内容与边框间的距离
  left=0mm, % 左边内边距
  right=0.05mm, % 右边内边距
  toptitle=0.05mm, % 标题与框内上边的距离
  %drop fuzzy shadow % 阴影效果
]
\textbf{Theorem 4.1.} Let $f_i$ and $t_i$ represent the visual and text features of sample $i$, where $\|f_i\| = 1$, $\|t_i\| = 1$, and $f_i^{\text{new}}$ represents the visual feature after gradient update. $p_{ij}$ denotes the classification probability (Eq.~\ref{eq:cross_sim_and_p}). The difference in cosine similarity between two samples, $f_i$ and $f_k$, before and after one step of training using $\mathcal{L}_{\mathrm{vlm}}$ (Eq.~\ref{eq:vlm_loss}) is $\Delta \cos(\theta_{ik}) = f_i^{\text{new}}\cdot f_k^{\text{new}} -  f_i \cdot f_k$, there is:

\begin{scriptsize}
\begin{equation}
\setlength{\abovedisplayskip}{0pt}
\setlength{\belowdisplayskip}{1pt}
\begin{aligned}
    \Delta \cos(\theta_{ik}) &=  \eta \frac{1}{\tau} \left( f_i \cdot t_k - \sum_{j=1}^{N} p_{kj} f_i \cdot t_j + f_k \cdot t_i - \sum_{j=1}^{N} p_{ij} f_k \cdot t_j \right) \\ &\quad + O(\eta^2),
\end{aligned}
\end{equation}
\end{scriptsize}

where $\eta$ is the learning rate and $\tau$ is the temperature coefficient. Detailed proof in the Appendix.
\end{tcolorbox}
%\vspace{-0.2cm}
\noindent{Learning objective of VLMs' cross-entropy loss function satisfies: $f_i\cdot t_i > f_i\cdot t_j, \forall j \neq i$. Therefore: (1) When $f_i$ and $f_k$ belong to the same class, $t_i = t_k$, there is $\Delta \cos(\theta_{ik}) > \eta \frac{1}{\tau} \left( f_i \cdot t_i - 1\cdot f_i \cdot t_i + f_k \cdot t_k - 1\cdot f_k \cdot t_k \right) =0 $. (2) When $f_i$ and $f_k$ belong to different classes, we analyze the values of $\Delta \cos(\theta_{ik})$ during the training process across 200 episodes and observe that $\Delta \cos(\theta_{ik}) <0$, as shown in Fig.~\ref{fig:delta_cos}. 
% \begin{wrapfigure}{r}{.45\linewidth}
% \centering
% % \setlength{\abovecaptionskip}{0cm}
% \includegraphics[width=1 \linewidth]{./Figs/disc_in_clip/delta_cos.jpg}
% \caption{.}
% \label{fig:delta_cos}
% %\vspace{-0.2cm}  
% \end{wrapfigure}
From points (1) and (2), that is, during fine-tuning, visual features of the same category cluster together while those of different categories are pushed apart, which means visual learning is consistently present in the fine-tuning process. Refer to the Appendix for detailed proof.

%\vspace{-0.2cm}
\subsubsection{Fine-tuning is a dual-valve drainage process.}
In Theorem 4.1, we demonstrated that during the fine-tuning process of VLMs based on cross-modal loss (Eq.~\ref{eq:vlm_loss}), there is an inherent direction towards discriminative learning of visual features, even without explicitly guiding the model to learn such features. Therefore, fine-tuning actually drives the VLM into two learning directions: the first is the learning of cross-modal relationships, which is the primary goal of most existing VLM fine-tuning efforts~\cite{li2025logits,huang2024lp++,tang2024amu,zanella2024low}, while the second is visual learning, an aspect overlooked and underexplored in previous research.

With this understanding, we can visualize the VLM fine-tuning as a dual-valve drainage system (Fig.~\ref{fig:intro_c_base}). During fine-tuning, the loss value represents the resources (water) consumed, while the alignment of visual-text features (the target goal) and visual learning (the unintended goal) act as two contributing valves in this process:

\begin{tcolorbox}[colback=gray!10, colframe=gray!30, title=\textcolor{black}{\textbf{Fine-tuning is a dual-valve drainage process.}}]
% \textbf{Target:} The learning of visual-text relationships, specifically the image-text classification tasks.

%\vspace{-0.2cm}
\textbf{Resources:} Loss value ($\mathcal{L}_{\mathrm{vlm}}$). The decrease in loss value during training is analogous to the drainage of resources (water) in a drainage process.

\textbf{Valve1:} The branch of learning of the visual-text relationship. (True target)

\textbf{Valve2:} The visual learning branch.

\textbf{Risk:} Valve 2 acts as a shortcut, reducing resource flow to Valve 1, the primary target.
\end{tcolorbox}
%\vspace{-0.1cm}

Existing research~\cite {geirhos2020shortcut,song2024shortcut,tang2021augmented} in traditional visual models suggests that such dual-branch networks have an inherent risk: \textbf{the diversion of resources will suppress the performance of the primary objective branch}. 
Therefore, although no prior works on VLM-based fine-tuning suggest such a risk, we hypothesize that \textit{visual learning during the fine-tuning can also become a shortcut that diverts resources from the visual-text alignment learning}, thereby harming the model's learning of visual-text relationships. 
% To demonstrate the existence of this phenomenon, \textbf{in the following sections, we show that during the fine-tuning process: 1) Visual discriminative learning (valve2) provides a way to reduce the cross-modal loss (consuming resources). 2) While it reduces the loss, it degrades the alignment of visual-text features (valve1, the main target of fine-tuning).}

\vspace{-0.2cm}
\subsection{Experimental Proof}
\vspace{-0.1cm}
In this section, we prove our hypothesis from two aspects:

\noindent$\bullet$\textbf{Visual learning can reduce the object loss.} During the fine-tuning of VLMs, visual learning can effectively reduce $\mathcal{L}_{\mathrm{vlm}}$, meaning it is valve 2 that consumes resources.

\noindent$\bullet$\textbf{Visual learning hinders cross-modal alignment.} Opening valve 2 (the visual learning) weakens cross-modal alignment learning, while closing it improves alignment.

\subsubsection{Visual learning can reduce the object loss.}

\begin{figure}[t]
\centering
\vspace{-0.2cm} 
\includegraphics[width=0.99 \linewidth]{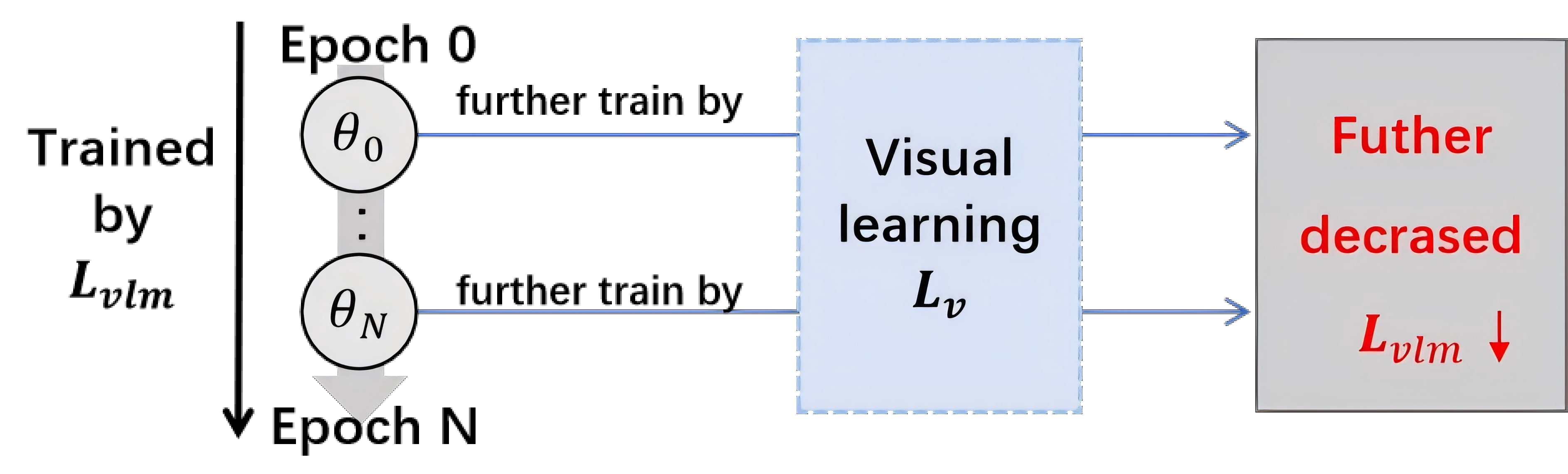}
%\vspace{-0.2cm}
\caption{Recording the model at each epoch during training and further training these models using visual learning, the target loss value $\mathcal{L}_{\mathrm{vlm}}$ can be effectively reduced even further.}
\label{fig:exp_prof_1}
\vspace{-0.2cm}  
\end{figure}

\noindent\textbf{Experiment Design.}
The fine-tuning objective of VLM is to reduce $\mathcal{L}_{\mathrm{vlm}}$ (Eq.~\ref{eq:vlm_loss}). To demonstrate that visual learning can also reduce $\mathcal{L}_{\mathrm{vlm}}$, we save the model at each epoch during the fine-tuning process and use visual learning (Eq.~\ref{eq:intra_sim}) to further train the model (Fig.~\ref{fig:exp_prof_1}). 

\noindent\textbf{Experiment Setup.}
We select four CDFSL datasets~\cite{mohanty2016using,helber2019eurosat,codella2019skin,wang2017chestx}, and consider Vit-B/16 from CLIP~\cite{radford2021learning} and PE-Core~\cite{bolya2025perception} as the base model, and fine-tuning by LoRA~\cite{hu2021lora}.

\noindent\textbf{Experiment Results.}
We record the decrease in the loss ($\mathcal{L}_{\mathrm{vlm}}$) during the visual learning. The results, shown in Fig.~\ref{fig:disc_loss_drop}, indicate that at each epoch of the fine-tuning process, the VLM's loss can be reduced by visual learning that disregards the visual-text relationships. In other words, visual learning, acting as the valve 2, can consume resources and reduce the loss value.

\subsubsection{Visual learning hinders cross-modal alignment.}
\begin{figure}[H]
\centering
\vspace{-0.2cm} 
\includegraphics[width=0.92 \linewidth]{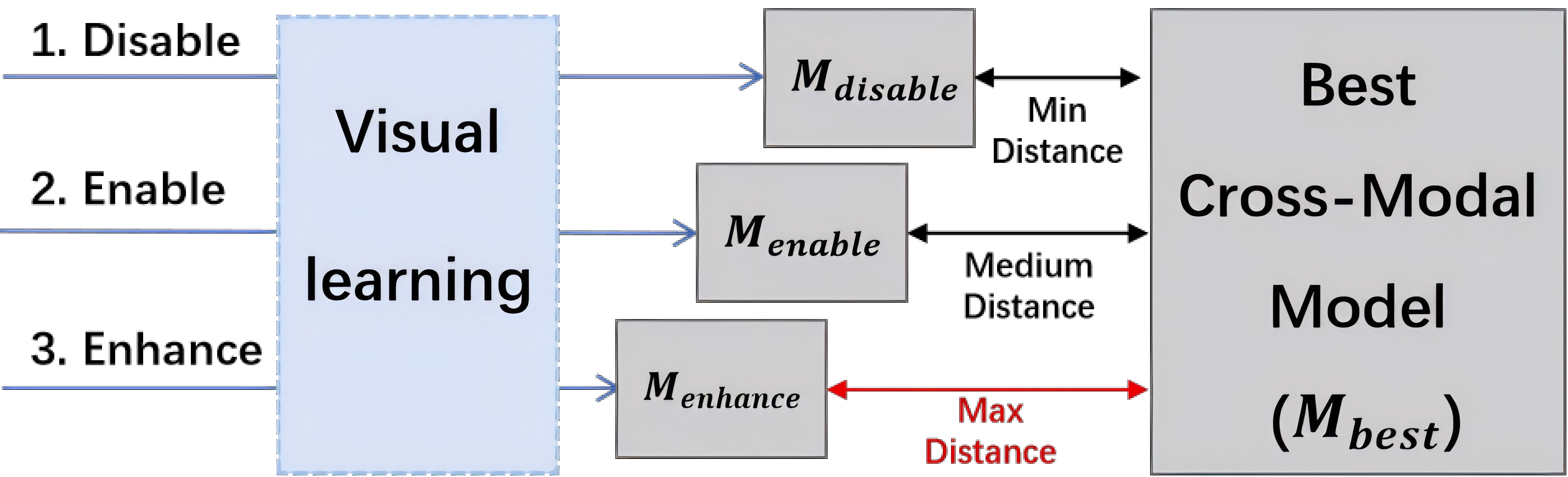}
\caption{We employ three strategies to restrict visual learning during the VLM fine-tuning process and measure the distance of the resulting models from the optimal cross-modal model. We find that visual learning increases the distance from the optimal model, which hinders visual-text alignment.}
\label{fig:exp_prof_2}
\vspace{-0.2cm}  
\end{figure}
In VLMs, cross-modal alignment is essential. One of the most direct indicators of this alignment is classification accuracy. Furthermore, we use the modality embedding shift experiment from previous work~\cite{liang2022mind,tian2025mind} to measure the gap between the trained model and the optimal modality alignment model, as shown in Fig.~\ref{fig:exp_prof_2}. This allows us to assess the degree of modality alignment in the model.

\begin{figure*}[!t]
    \centering
    % \subfloat[\tiny{Relationship in CLIP}]{\label{fig:opt_dir_anlyze}\includegraphics[width=0.16\textwidth]{./Figs/opt_dir_anlyze_half.png}}
    % \hfill
    \subfloat[Gap Shift]{\label{fig:gap_shift_exp_a}\includegraphics[width=0.19\textwidth]{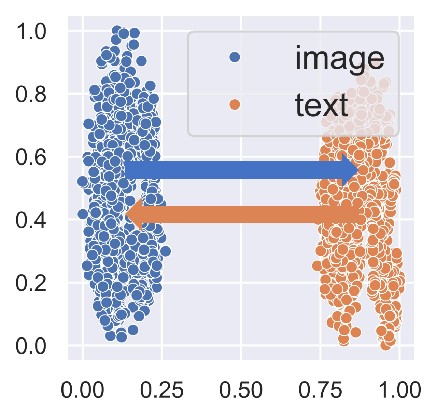}\vspace{-0.1cm}}
    \hfill
    % \subfloat[\scriptsize{MiniImageNet}]{\label{fig:gap_shift_exp_b}\includegraphics[width=0.20\textwidth]{./Figs/gap_shift_exp/mini_zero.jpg}}
    \subfloat[zero shot]{\label{fig:gap_shift_exp_b}\includegraphics[width=0.19\textwidth]{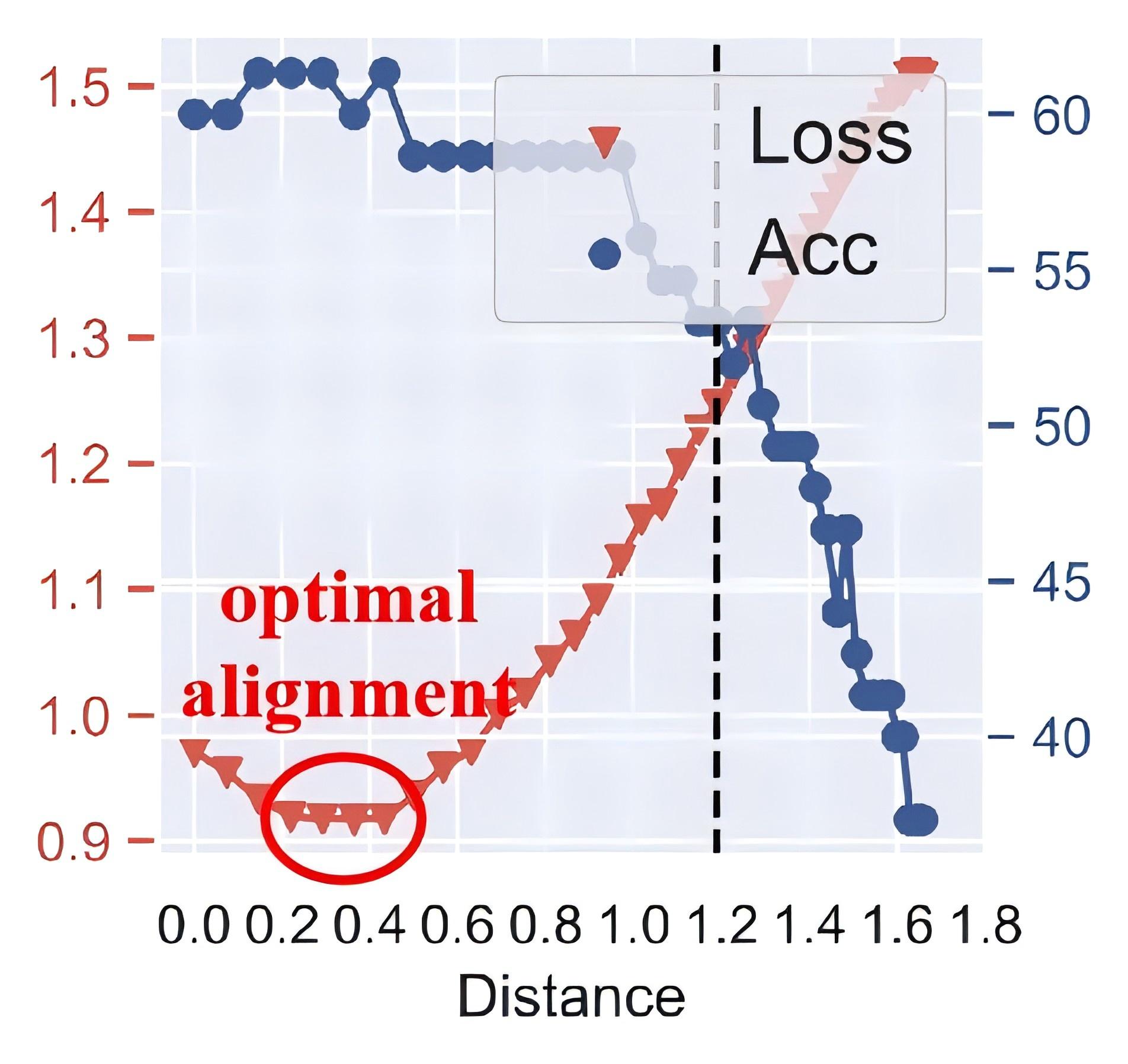}\vspace{-0.1cm}}
    \hfill
    \subfloat[$\mathcal{L}_{\mathrm{vlm}}$]{\label{fig:gap_shift_exp_c}\includegraphics[width=0.205\textwidth]{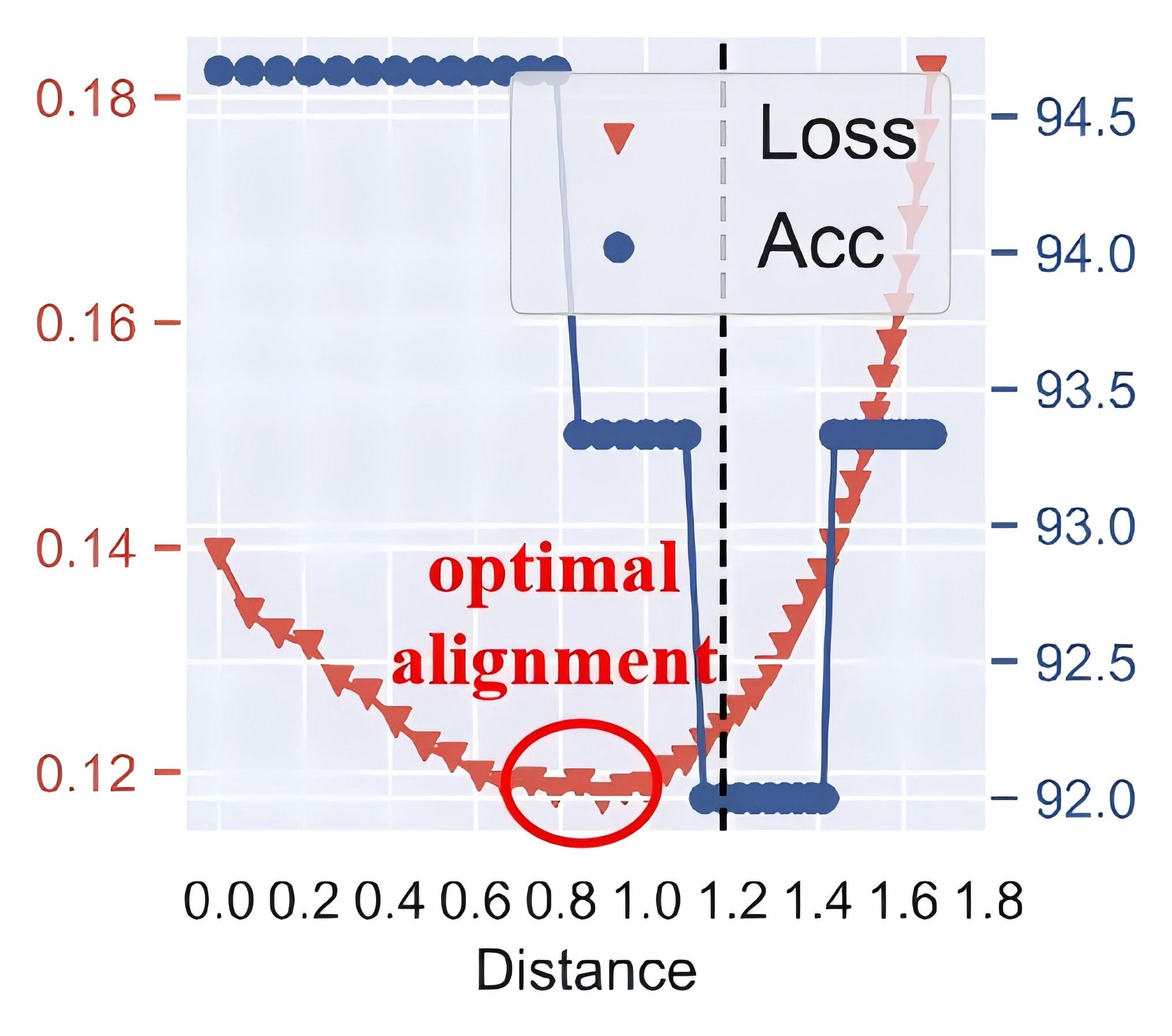}\vspace{-0.1cm}}
    \hfill
    \subfloat[$\mathcal{L}_{\mathrm{vlm}} + \mathcal{L}_{\mathrm{v}}$]{\label{fig:gap_shift_exp_d}\includegraphics[width=0.21\textwidth]{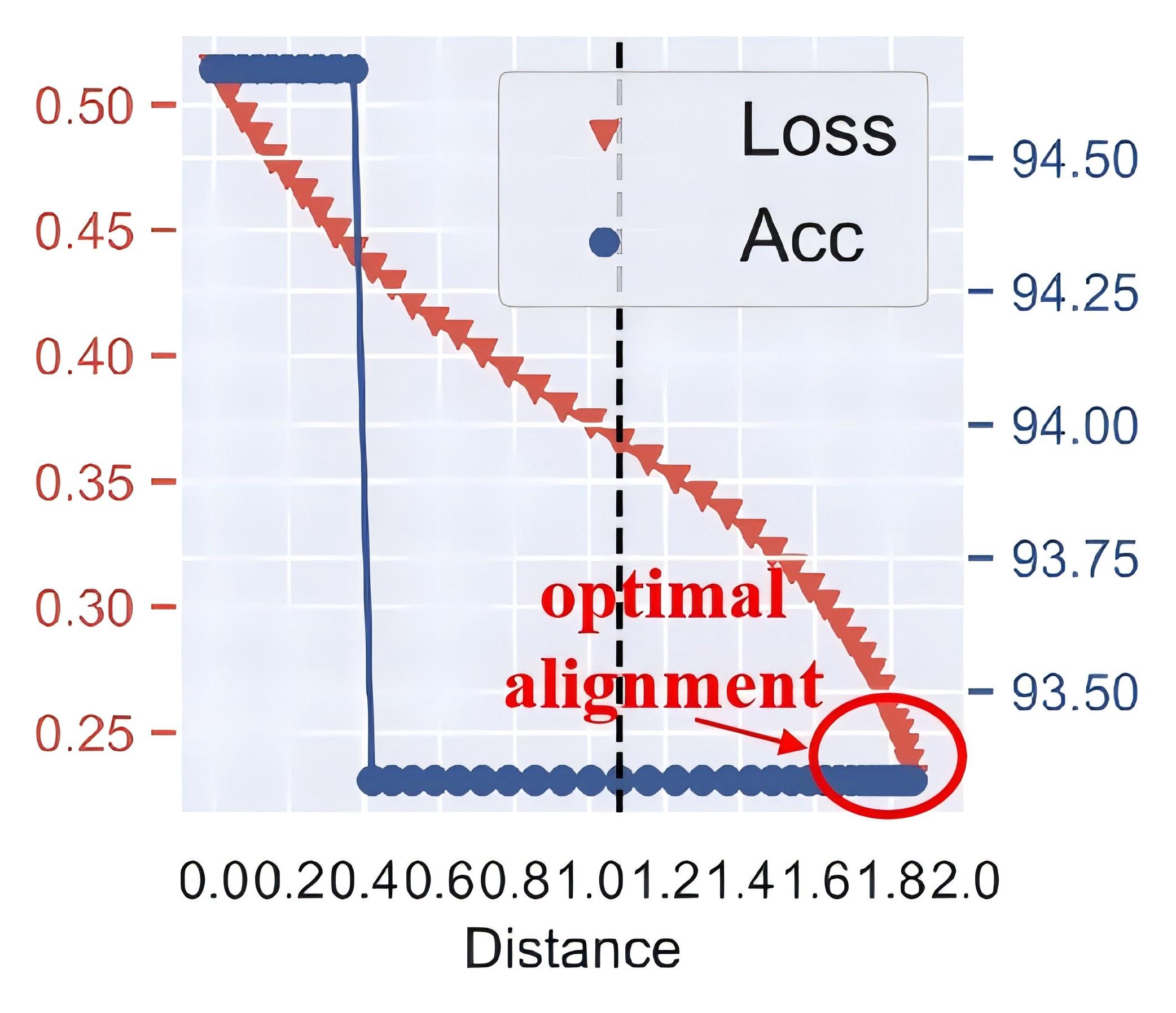}\vspace{-0.1cm}}
    \hfill
    \subfloat[$\mathcal{L}_{\mathrm{vlm}} + \mathcal{L}_{\mathrm{ad}}$]{\label{fig:gap_shift_exp_e}\includegraphics[width=0.20\textwidth]{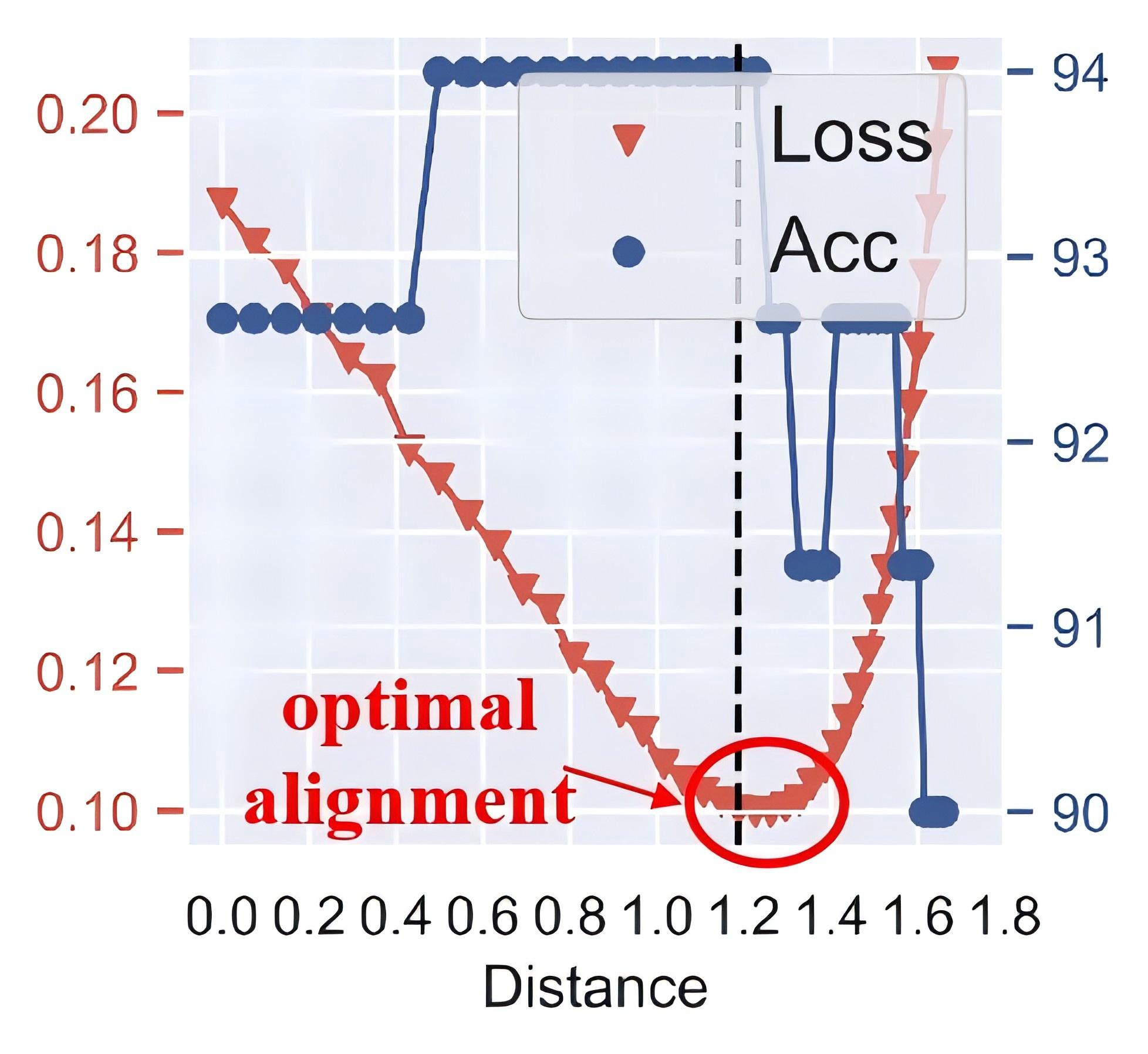}\vspace{-0.1cm}}
    \vspace{-0.3cm}
    \caption{
    (a) Gap shift operation~\cite{liang2022mind}. 
    % (b) In the natural domain of \textit{mini}ImageNet, without any learning or gap shift operation, CLIP is naturally in an optimal alignment state, which has the lowest loss and highest accuracy. The model's initial state is marked by the vertical dashed line. 
    (b) The model's initial state is marked by the vertical dashed line. In the cross-domain dataset EuroSAT, CLIP's modalities are misaligned, meaning that through gap shift, lower loss and higher accuracy can be achieved. (c) Fine-tuning does not effectively re-align the modalities. (d) When visual learning is enhanced ($\mathcal{L}_{\mathrm{v}}$), the misalignment becomes more pronounced. (e) When visual learning is suppressed ($\mathcal{L}_{\mathrm{ad}}$), the misalignment is mitigated.}
    \label{fig:gap_shift_exp}
    \vspace{-0.1cm}
\end{figure*}

\begin{table*}[!t]
%\vspace{-0.4cm}
\centering
\caption{Experiments with 800 5-way 5-shot episodes. A smaller Gap indicates better modality alignment in the model's original state. For comparison, in \textit{mini}ImageNet~\cite{deng2009imagenet}, Acc and Gap is 98.56 / 0.007. The modality alignment is poor in cross-domain scenarios (even after fine-tuning), which is exacerbated by visual learning ($\mathcal{L}_{v}$). In contrast, our methods ($\mathcal{L}_{ad}$ and $\mathcal{L}_{ra}$), enhances modality alignment.}\vspace{-0.2cm}
\label{tab:gap_story} 
\begin{adjustbox}{max width=0.95\linewidth}
\begin{tabular}{c|*{8}{c}|*{8}{c}} 
    \toprule 
    \multicolumn{1}{c|}{\multirow{3}{*}{Method}}  & \multicolumn{8}{c|}{CLIP}  & \multicolumn{8}{c}{PE-Core} \\ 
    \cmidrule(lr){2-9} \cmidrule(lr){10-17}
       & \multicolumn{2}{c}{CropDisease}  & \multicolumn{2}{c}{EuroSAT}  & \multicolumn{2}{c}{ISIC}   & \multicolumn{2}{c|}{ChestX} & \multicolumn{2}{c}{CropDisease}  & \multicolumn{2}{c}{EuroSAT}  & \multicolumn{2}{c}{ISIC}   & \multicolumn{2}{c}{ChestX} \\ 
   \cmidrule(lr){2-3}\cmidrule(lr){4-5}\cmidrule(lr){6-7}\cmidrule(lr){8-9}\cmidrule(lr){10-11}\cmidrule(lr){12-13}\cmidrule(lr){14-15}\cmidrule(lr){16-17}
     & Acc$\uparrow$ & Gap$\downarrow$ & Acc$\uparrow$ & Gap$\downarrow$ & Acc$\uparrow$ & Gap$\downarrow$ & Acc$\uparrow$ & Gap$\downarrow$  & Acc$\uparrow$ & Gap$\downarrow$ & Acc$\uparrow$ & Gap$\downarrow$ & Acc$\uparrow$ & Gap$\downarrow$ & Acc$\uparrow$ & Gap$\downarrow$ \\
    \midrule 
    Fine-tune (FT, $\mathcal{L}_{vlm}$) & 95.8 & 0.014 &93.3 &0.048 &52.2 & 0.406 & 24.6 & 0.356 & 97.2 & 0.011 & 94.0 & 0.060 & 59.4 & 0.538 & 24.4 & 0.661\\ \cmidrule(lr){1-17}
    FT + $\mathcal{L}_{v}$ (Eq.~\ref{eq:intra_sim})  & 95.6 & 0.022 &92.7 &0.072 &51.9 & 0.626 & 23.2 & 0.742 & 94.2 & 0.118 & 89.0 & 0.211 & 49.4 & 1.312 & 23.0 & 1.752\\ \cmidrule(lr){1-17}
    \textbf{FT + \bm{$\mathcal{L}_{ad}$} (Eq.~\ref{eq:anti_visual_loss}) }  & 96.1 & 0.012 &93.3 &\textbf{0.024} &55.4 & 0.191 & 25.5 & 0.249 & 97.4 & 0.007 & 94.2 & 0.024 & 60.6 & 0.254 & 25.8 & 0.480\\
    \cmidrule(lr){1-17}
    \textbf{FT + \bm{$\mathcal{L}_{ad}$} + \bm{$\mathcal{L}_{ra}$}}  & \textbf{96.6} & \textbf{0.009} &\textbf{94.2} &0.027 &\textbf{56.1} & \textbf{0.171} & \textbf{26.6} & \textbf{0.238} & \textbf{98.1} & \textbf{0.005} & \textbf{94.8} & \textbf{0.022} & \textbf{61.4} & \textbf{0.193} & \textbf{26.8} & \textbf{0.438}\\
    \bottomrule 
\end{tabular}
\end{adjustbox}
\vspace{-0.3cm}
\end{table*}

\begin{figure*}[t]
\centering
\begin{minipage}[b]{.48\linewidth}
    \centering
    \centering
    \subfloat[“a photo of a Strawberry\_healthy.”]{
    \includegraphics[width=0.24\linewidth]{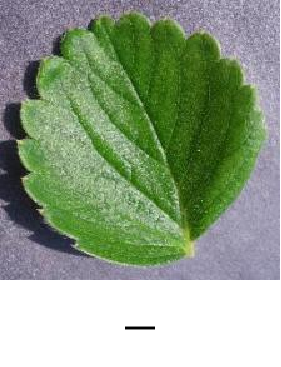} 
    \includegraphics[width=0.24\linewidth]{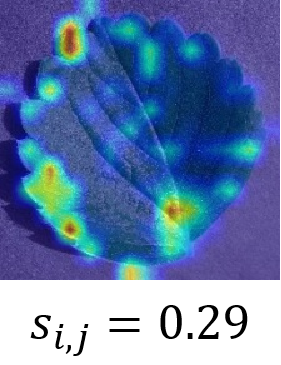} 
    \includegraphics[width=0.24\linewidth]{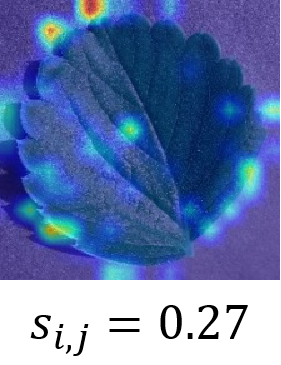} 
    \includegraphics[width=0.24\linewidth]{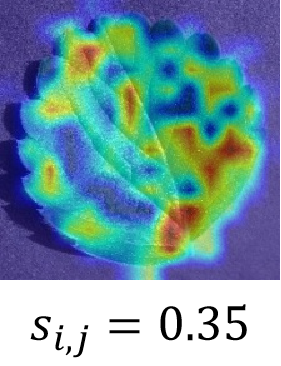} 
    }
\end{minipage}
\hspace{2mm}
\begin{minipage}[b]{.48\linewidth}
    \centering
    \centering
    \subfloat[“a photo of a Benign Keratosis.”]{
    \includegraphics[width=0.24\linewidth]{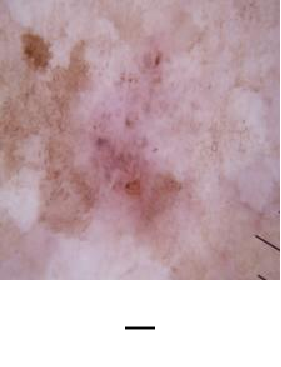} 
    \includegraphics[width=0.24\linewidth]{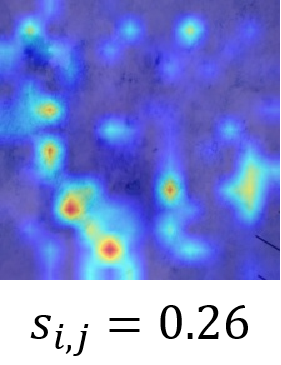} 
    \includegraphics[width=0.24\linewidth]{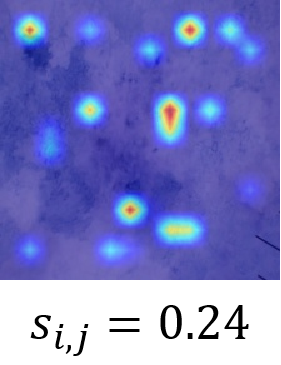} 
    \includegraphics[width=0.24\linewidth]{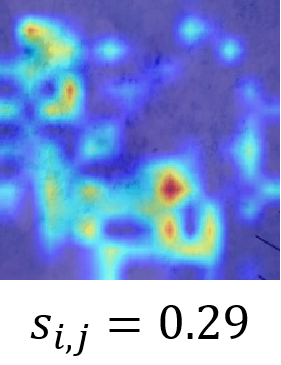} 
    }
\end{minipage}
\vspace{-0.1cm}
\caption{(a)(b) The Attention maps of the three models. From left to right: the original image, strategy 2 result (trained with $\mathcal{L}_{\mathrm{vlm}}$), strategy 3 result (trained with $\mathcal{L}_{\mathrm{vlm}} + \mathcal{L}_{\mathrm{v}}$), and strategy 1 result (trained with $\mathcal{L}_{\mathrm{vlm}} + \mathcal{L}_{\mathrm{ad}}$). $s_{i, j}$ represents the cosine similarity between the image features and the text features. A higher similarity indicates better alignment. 
}
\label{fig:heat_map}
%\vspace{-0.1cm}
\end{figure*}

\noindent\textbf{Experiment Design.} 
We employ three strategies: 1. Disable visual learning (\(\mathcal{L}_{\mathrm{vlm}} + \mathcal{L}_{\mathrm{ad}}\)), 2. Enable visual learning (\(\mathcal{L}_{\mathrm{vlm}}\)), and 3. Enhance visual learning (\(\mathcal{L}_{\mathrm{vlm}} + \mathcal{L}_{\mathrm{v}}\)) to train and obtain models \(M_{\text{disable}}\), \(M_{\text{enable}}\), and \(M_{\text{enhance}}\), respectively. We then use these models to extract text and visual features. Next, we manually adjust both the text and visual features by modifying the gap between the two sets of features, narrowing or enlarging the gap by \(\alpha\) (Fig.~\ref{fig:gap_shift_exp_a}):
\begin{equation}
\setlength{\abovedisplayskip}{2pt}
\setlength{\belowdisplayskip}{1pt}
\vec{\Delta}_{\mathrm{gap}}=\frac{1}{n} \sum_{i=1}^n f_i-\frac{1}{n} \sum_{i=1}^n t_i, 
\end{equation}
\begin{equation}
\setlength{\abovedisplayskip}{1pt}
\setlength{\belowdisplayskip}{1pt}
f_i^{\text {shift }}=\mathcal{N}\left(f_i-\alpha \vec{\Delta}_{\mathrm{gap}}\right), \quad 
t_i^{\text {shift }}=\mathcal{N}\left(t_i+\alpha \vec{\Delta}_{\mathrm{gap}}\right),
\end{equation}
here, $f$ and $t$ represent the visual and text features, and $\mathcal{N}$ denotes the L2 normalization operation. We then track the corresponding changes in validation loss and accuracy. If the model's modality alignment is strong, it will exhibit high classification (as Eq.~\ref{eq:cross_sim_and_p}) accuracy and low validation loss (as Eq.~\ref{eq:vlm_loss}) at its initial state. Conversely, if the alignment is weak, artificially modifying the cross-modal relationship should result in improved performance. 

\noindent\textbf{Experiment Setup.}
We select four CDFSL datasets~\cite{mohanty2016using,helber2019eurosat,codella2019skin,wang2017chestx}, and consider Vit-B/16 from CLIP~\cite{radford2021learning} and PE-Core~\cite{bolya2025perception} as the base model, and fine-tuning by LoRA~\cite{hu2021lora}.

\noindent\textbf{Experiment Results.}
The results are presented in Fig.~\ref{fig:gap_shift_exp}, where the X-axis represents the modal gap, calculated as the Euclidean distance between the mean of the text features and the mean of the visual features: $\left\|\vec{\Delta}_{\text {gap }}\right\|$. The initial state of the model is indicated by the black vertical line.

Without any fine-tuning (Fig.~\ref{fig:gap_shift_exp_b}, zero shot), modality alignment is poor (the initial state does not correspond to the lowest Loss and highest Acc). However, gap shifting can achieve higher accuracy and reduce validation loss. Using strategy 2, which fine-tunes the model without disabling visual learning (Fig.~\ref{fig:gap_shift_exp_c}), does not result in satisfactory re-alignment (not the lowest Loss and highest Acc). Moreover, when using strategy 3, which further enhances visual learning (Fig.~\ref{fig:gap_shift_exp_d}), the misalignment issue becomes even more pronounced (the initial state is far from the lowest Loss and highest Acc). In contrast, when we use strategy 1 to disable visual learning during the fine-tuning process (Fig.~\ref{fig:gap_shift_exp_e}), the model achieves near-optimal alignment (lowest Loss and highest Acc) without any adjustments.

\begin{figure*}[t]
\centering
\includegraphics[width=0.95 \linewidth]{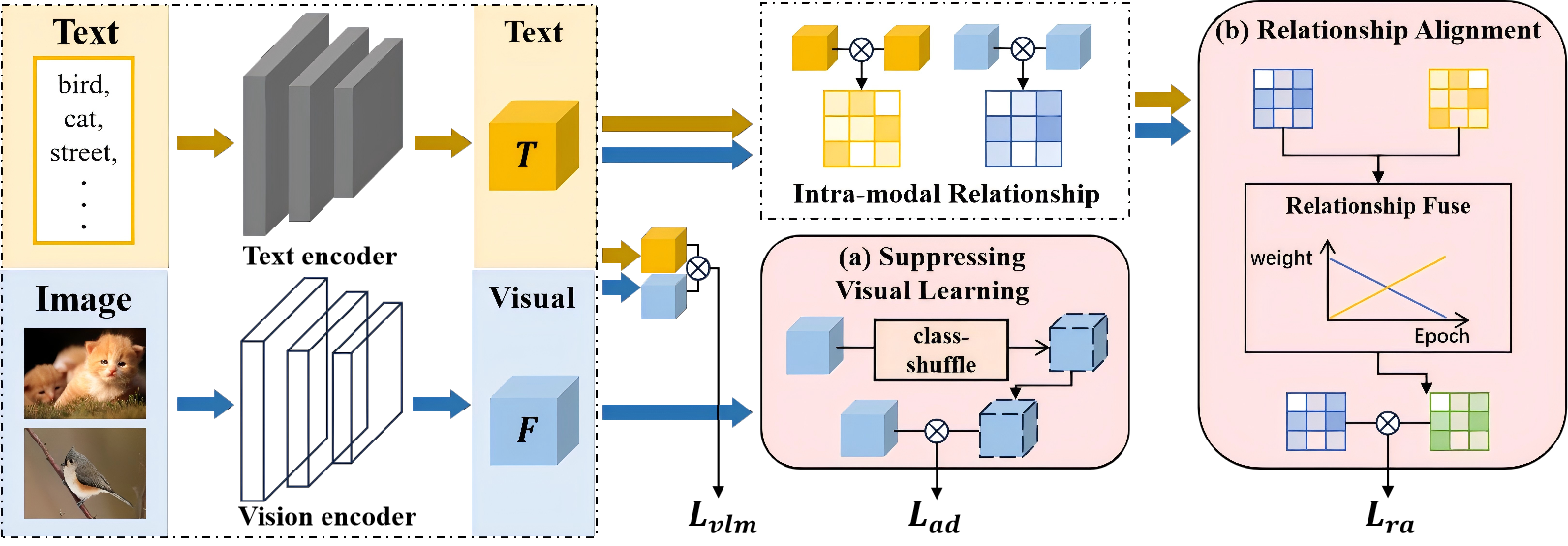}\vspace{-0.2cm}
\caption{The overall architecture of the model, with the pink region highlighting our proposed method. (a) Firstly, we propose suppressing the visual learning to resist its shortcut during the finetuning, which constrains the model to focus more on the cross-modal alignment.
(b) Then, we propose using semantic relationships to gradually align the internal relationships of the visual modality with those of the text modality, thereby enhancing cross-modal alignment.}
\label{fig:method_full2}
\vspace{-0.2cm}  
\end{figure*}

To further illustrate this phenomenon, we present numerical results in Tab.~\ref{tab:gap_story}. The metrics include classification accuracy (Acc) and the loss gap (Gap), which measures the difference between the optimal validation loss achieved via gap shifting and the initial state’s validation loss. A small loss gap indicates that the model's initial state is nearly optimal ($M\approx M_{best}$), suggesting good modality alignment. The results (FT → FT + $\mathcal{L}_{v}$) corroborate our findings: \textbf{visual learning hinders visual-text alignment.} 

\vspace{0.1cm}
\noindent\textbf{Case Study.}
The visualization results in Fig.~\ref{fig:heat_map} provide a detailed explanation of how visual learning impacts modality alignment. When visual learning is emphasized (column 2 → column 3), the model focuses more on discriminative features. However, discriminative features alone cannot align with text such as “a photo of a Strawberry\_healthy”. This reduction in attention scope hinders generalization, resulting in lower image-text similarity and poorer alignment.

% Unlike~\cite{shangguan2025cross,jing2020cross,hao2023uncertainty,wang2022toward}, we consider the scenario of cross-domain few-shot fine-tuning and highlight the limitations of fine-tuning for modality alignment in this context. Additionally, we provide a clear visual analysis of modality misalignment in these scenarios using metrics (Acc and Loss Gap). 
\vspace{-0.2cm}
\subsection{Conclusion and Discussion}
\vspace{-0.1cm}
Based on the analysis above, we offer the following interpretation. In classification tasks on VLMs, cross-entropy loss (Eq.~\ref{eq:vlm_loss}) is used as the loss function. However, during the fine-tuning process, visual learning provides a shortcut to reducing the loss function without considering cross-modal relationships, hindering the learning of the crucial cross-modal alignment.

\noindent\textbf{Our insight:} It is essential to appropriately constrain visual learning during the fine-tuning of the VLM model to guide the process toward prioritizing cross-modal learning.

%% file: Tex/04_method_yxz.tex
\section{Method}
%\vspace{-0.2cm}
% On the visual-modal side, we perturb the visual-modal learning to resist its shortcut during the CLIP finetuning, which constrains the model to focus more on the cross-modal alignment during the finetuning.
% On the text-modal side, we construct a bridge between the general semantics of CLIP's pretraining data and the target-domain expert semantics by mapping target-domain semantics to an intermediate basic semantic space, which reduces the shift between general and target-domain text semantics.

Based on interpretation, we further propose a method to assist the fine-tuning process of VLMs in SF-CDFSL (Fig.~\ref{fig:method_full2}). Firstly, we propose the SVL module to suppress the model's visual learning, encouraging the fine-tuning to prioritize the learning of cross-modal relationships. Then, we propose the RA module, using a visual-text-fused relationship matrix to guide visual-modal relationship learning, promoting alignment learning between modalities.

%\vspace{-0.2cm}
\subsection{Suppressing Visual Learning (SVL)}
%\vspace{-0.1cm}
Let \( A^v \) denote the visual feature similarity matrix and \( A^t \) denote the text feature similarity matrix. \(\mathcal{F}\) and \(\mathcal{T}\) are the sets of visual and text features. Then, we have:
\begin{equation}
\setlength{\abovedisplayskip}{3pt}
\setlength{\belowdisplayskip}{3pt}
A^v = \mathcal{F}\mathcal{F}^T, A^t = \mathcal{T}\mathcal{T}^T.
\end{equation}
Visual learning makes visual features of the same class similar and those of different classes distant, meaning that for any $i$, $j$ where $i \ne j$, $\frac{A^v_{i, i}}{A^v_{i,j}}$ and $\frac{A^v_{j,j}}{A^v_{i,j}}$ are maximized. However, our previous analysis indicated that in VLMs, visual learning can actually harm cross-modal performance by excessively focusing on discriminative learning within the visual modality. Therefore, we propose $\mathcal{L}_{\mathrm{ad}}$ (Eq.~\ref{eq:anti_visual_loss}) to suppress visual learning. Specifically, in contrast to the classifier weights that correspond one-to-one to the categories in Eq.~\ref{eq:intra_sim}, we generate classifier weights $w^{\prime}$ by using visual features of randomly selected samples from the support set and compute their cross-entropy loss, as illustrated in Fig.~\ref{fig:method_full2}a. 

% where $x^{\prime}_i$ represents the $i$-th sample randomly selected from the support set, irrespective of class. We refer to $w^{\prime}$ as the class-shuffle weights. By using class-shuffle weights for visual-modal training, we suppress the learning of discriminative features. Therefore, $L_{ad}$ serves as a disturbance that disrupts the visual-modal shortcut in CLIP fine-tuning.

% It is important to note, as shown in Fig.~\ref{fig:method_full2}, that Anti-Discriminative Loss ($L_{ad}$) is applied only during the early epochs of fine-tuning, and a smaller hyperparameter, $\lambda$, is used to control the contribution of $L_{ad}$ during the training process, ensuring that $L_{ad}$ as a disturbance without dominating the entire training procedure. In subsequent epochs, we remove $L_{ad}$ to allow the model to learn visual-modal relationships that are crucial for multi-modal tasks (refer to Section 6.4 for more discussion). As a result, the model's fine-tuning loss function can be expressed as:
% %%\vspace{-0.1cm}
% \begin{equation}
% \label{eq:final_loss}
% %\setlength{\abovedisplayskip}{1pt}
% \setlength{\belowdisplayskip}{1pt}
% \mathcal{L} = \begin{cases} 
% \mathcal{L}_{\text{cross}} + \lambda \mathcal{L}_{\text{ad}} & \text{(initial epochs)} \\
% \mathcal{L}_{\text{cross}} & \text{(later epochs)}
% \end{cases} 
% \end{equation}

%It is worth noting that $L_{ad}$ can also be applied to the text branch, treating each text description as a separate sample. In subsequent experiments, we found that adding $L_{ad}$ to the text branch also improves performance.
\begin{table*}[!t]
\belowrulesep=0pt
\aboverulesep=0pt
\caption{
Accuracies (\%) of target domain datasets of 5-way 1-shot and 5-shot tasks. Refer to the Appendix for the extended table.}
%\vspace{-0.2cm}
\label{tab:cdfsl}    
\centering
\begin{adjustbox}{max width=0.95\linewidth}
\begin{tabular}{*{8}{c|c c|c c c c c|} }  
\toprule 
 \multirow{1}*{Task}  &  \multirow{1}*{Method}  &\multirow{1}*{backbone} & \multirow{1}*{ISIC} & \multirow{1}*{EuroSAT} & \multirow{1}*{CropDisease} & \multirow{1}*{ChestX} & \multirow{1}*{Avg} \\           
 \midrule
  \multirow{14}{*}{\rotatebox{90}{5-way 1-shot}} 
  &StepSTP~\cite{xu2024step} &ViT/CLIP &32.97±0.27 &70.01±0.21 &84.84±0.72 &\textbf{22.84±0.95} &52.68 \\
  &Tip-Adapter~\cite{zhang2021tip}  &ViT/CLIP &32.68±0.37 &75.44±0.51 &77.15±0.66 &22.24±0.26 &51.87 \\
  &AMU-Tuning~\cite{tang2024amu}  &ViT/CLIP &32.29±0.67 &72.24±0.71 &80.20±0.86 &21.56±0.36 &51.57 \\
  &LP++~\cite{huang2024lp++}  &ViT/CLIP &33.63±0.41 &73.05±0.55 &81.84±0.66 &21.72±0.42 &52.56 \\
  &LDC~\cite{li2025logits}  &ViT/CLIP &33.72±0.46 &74.39±0.52 &84.07±0.61 &22.32±0.36 &53.62 \\
  &CoOp~\cite{zhou2022learning}  &ViT/CLIP  &32.86±0.47 &72.08±0.66 &80.50±0.74 &21.65±0.32 &51.77 \\
  &\textbf{CoOp + OURS}  &ViT/CLIP &33.44±0.50 &72.51±0.65 &81.77±0.79 &21.71±0.31 &52.36 \\
  &Maple~\cite{khattak2023maple}  &ViT/CLIP &33.38±0.49 &76.05±0.63 &81.78±0.72 &21.09±0.31 &53.07 \\
  %\rowcolor{cyan!10} \cellcolor{white}
  &\textbf{Maple + OURS}  &ViT/CLIP &35.11±0.51 &76.92±0.65 &82.51±0.69 &21.64±0.34 &54.05 \\
  &CLIP-LoRA-Vision~\cite{zanella2024low}  &ViT/CLIP &36.40±0.42 &81.72±0.52 &84.62±0.62 &21.86±0.32 &56.07 \\
  \rowcolor{cyan!10} \cellcolor{white}
  &\textbf{CLIP-LoRA-Vision + OURS}  &ViT/CLIP &\textbf{38.12±0.48} &\textbf{85.02±0.46} &\textbf{87.20±0.51} &22.68±0.41 &\textbf{58.26} \\
  \cline{2-8}
  &SigLIP2-LoRA~\cite{tschannen2025siglip} &ViT/SigLip2  &33.47 &74.16 & 87.50 &21.44 & 54.14 \\
  \rowcolor{cyan!10} \cellcolor{white}
  &\textbf{SigLIP2-LoRA + OURS} &ViT/SigLip2 &\textbf{36.88} &\textbf{78.04} & \textbf{90.85} &\textbf{22.27} & \textbf{57.01} \\
  \cline{2-8}
  &PE-Core-LoRA~\cite{bolya2025perception} &ViT/PE-Core  &40.89 &84.49 & 91.75 &22.02 & 59.78 \\
  \rowcolor{cyan!10} \cellcolor{white}
  &\textbf{PE-Core-LoRA + OURS} &ViT/PE-Core &\textbf{45.01} &\textbf{86.83} & \textbf{93.03} &\textbf{23.66} & \textbf{62.14} \\
  \midrule
  \midrule
  % &GNN [12] &RN10 &Y &- &25.27±0.46 &43.94±0.67 &83.64±0.77 &87.96±0.67 &60.20 \\
  % &FWT [40] &RN10 &Y &- &25.18±0.45 &43.17±0.70 &83.01±0.79 &87.11±0.67 &59.62 \\
  % &LRP [38] &RN10 &Y &- &24.53±0.30 &44.14±0.40 &77.14±0.40 &86.15±0.40 &57.99 \\
  % &ATA~\cite{wang2021cross} &RN10 &Y &- &24.32±0.40 &44.91±0.40 &83.75±0.40 &90.59±0.30 & 60.89 \\
  % &AFA~\cite{hu2022adversarial} &RN10 &Y &-  &46.01±0.40 &85.58±0.40 &88.06±0.30 &25.02±0.20 &61.17 \\
  % &wave-SAN~\cite{fu2022wave} &RN10 &Y &-  &44.93±0.67 &85.22±0.71 &89.70±0.64 &25.63±0.49 &61.37 \\
  % &StyleAdv~\cite{fu2023styleadv} &RN10 &Y &- &45.77±0.51 &86.58±0.54 &93.65±0.39 &26.07±0.37 &63.02 \\
  % % &Fine-tune [14] &RN10 &Y &Y &25.97±0.41 &48.11±0.64 &79.08±0.61 &89.25±0.51 &60.60 \\
  % % &NSAE [25] &RN10 &Y &Y &27.10±0.44 &54.05±0.63 &83.96±0.57 &93.14±0.47 &64.56 \\
  % % &BSR [26] &RN10 &Y &Y &26.84±0.44 &54.42±0.66 &80.89±0.61 &92.17±0.45 &63.58 \\
  % &ATA-FT~\cite{wang2021cross} &RN10 &Y &Y &25.08±0.20 &49.79±0.40 &89.64±0.30 &95.44±0.20 &64.99 \\
  % &DARA~\cite{zhao2023dual} &RN10 &Y &Y &56.28±0.66 &85.84±0.54 &95.32±0.34 &27.54±0.42 &66.25 \\
  % &StyleAdv-FT~\cite{fu2023styleadv} &RN10 &Y &Y &53.05±0.54 &91.64±0.43 &96.51±0.28 &26.24±0.35 &66.86 \\
  % \cline{2-10}
   \multirow{14}{*}{\rotatebox{90}{5-way 5-shot}} 
  % &PMF~\cite{hu2022pushing} &ViT/DINO &Y &Y &50.12 &85.98 &92.96 &27.27 &64.08 \\
  % &StyleAdv-FT~\cite{fu2023styleadv} &ViT/DINO &Y &Y &51.23±0.51 &90.12±0.33 &95.99±0.27 &26.97±0.33 &66.08 \\
  % &FLoR~\cite{zou2024flatten} &ViT/DINO &Y &Y &53.06 &90.75 &96.47 & 27.02 & 66.83 \\
  % &CD-CLS~\cite{zou2025closer} &ViT/DINO &Y &Y  &54.69 & 91.53 & 96.27 &27.66 & 67.54 \\
  % &AttnTemp~\cite{zouattention} &ViT/DINO &Y &Y  &54.91 &90.82 & 96.66 &28.03 & 67.61 \\  
  % \cline{2-10}
  % &FN+VDB~\cite{yazdanpanah2022visual} &RN18 &- &Y &47.48±0.59 &87.31±0.50 &94.63±0.37 &25.55±0.43 &64.74 \\
  % &IM-DCL~\cite{xu2024enhancing} &RN10 &- &Y &52.74±0.69 &89.47±0.42 &95.73±0.38 &\textbf{28.93±0.41} &66.72 \\
  % \cline{2-10}
  % &SeGD-VPT~\cite{zhuo2024prompt} &ViT/CLIP &- &Y &23.20±0.30 &53.10±0.51 &93.81±0.24 &96.93±0.25 &66.76 \\
  &StepSTP~\cite{xu2024step} &ViT/CLIP &52.12±0.36 &89.40±1.05 &96.01±0.88 &26.36±0.97 &65.97 \\
  &Tip-Adapter~\cite{zhang2021tip}  &ViT/CLIP &46.96±0.59 &87.24±0.33 &94.19±0.39 &24.07±0.44 &63.12 \\
  &AMU-Tuning~\cite{tang2024amu}  &ViT/CLIP &44.60±0.62 &88.47±0.39 &94.26±0.52 &23.34±0.41 &62.66 \\
  &LP++~\cite{huang2024lp++}  &ViT/CLIP &48.49±0.44 &87.48±0.42 &94.47±0.38 &23.89±0.29 &63.58 \\
  &LDC~\cite{li2025logits}  &ViT/CLIP  &49.70±0.33 &90.82±0.22 &96.71±0.34 &25.89±0.21 &65.78 \\
  &CoOp~\cite{zhou2022learning}  &ViT/CLIP &45.78±0.75 &85.88±0.49 &93.31±0.57 &23.35±0.50 &62.08 \\
  &\textbf{CoOp + OURS}  &ViT/CLIP &45.95±0.74 &86.88±0.52 &93.92±0.55 &23.55±0.43 &62.58 \\
  &Maple~\cite{khattak2023maple}  &ViT/CLIP &48.35±0.75 &89.04±0.52 &93.50±0.54 &22.96±0.50 &63.46 \\
  %\rowcolor{cyan!10} \cellcolor{white}
  &\textbf{Maple + OURS}  &ViT/CLIP &50.01±0.78 &91.00±0.50 &94.48±0.50 &23.45±0.49 &64.74 \\
  &CLIP-LoRA-Vision~\cite{zanella2024low}  &ViT/CLIP &52.22±0.71 &93.31±0.47 &95.88±0.42 &24.61±0.47 &66.50 \\
  \rowcolor{cyan!10} \cellcolor{white}
  &\textbf{CLIP-LoRA-Vision + OURS}  &ViT/CLIP &\textbf{56.14±0.46} &\textbf{94.14±0.34} &\textbf{96.64±0.39} &\textbf{26.61±0.43} &\textbf{68.38} \\
  \cline{2-8}
  &SigLIP2-LoRA~\cite{tschannen2025siglip} &ViT/SigLip2 &51.79 &91.39 & 96.43
  &24.24 & 65.96 \\
  \rowcolor{cyan!10} \cellcolor{white}
  &\textbf{SigLIP2-LoRA + OURS} &ViT/SigLip2 &\textbf{55.12} &\textbf{92.10} & \textbf{97.37} &\textbf{26.44} & \textbf{67.43} \\
  \cline{2-8}
  &PE-Core-LoRA~\cite{bolya2025perception} &ViT/PE-Core &58.81 &94.07 & 97.25 &24.44 & 68.64 \\
  \rowcolor{cyan!10} \cellcolor{white}
  &\textbf{PE-Core-LoRA + OURS} &ViT/PE-Core &\textbf{61.41} &\textbf{94.83} & \textbf{98.13} &\textbf{26.77} & \textbf{70.29} \\
\bottomrule
\end{tabular}
\end{adjustbox}
%%\vspace{-0.2cm}
\end{table*}

%\vspace{-0.2cm}
\subsection{Relationship Alignment (RA)}
%\vspace{-0.1cm}
After suppressing visual learning, a further issue arises: It is necessary to provide a new direction for learning relationships within the visual modality ($A^v$) to facilitate cross-modal alignment. To address this, we propose using the semantics relationships extracted by the text branch ($A^t$) to gradually guide the learning of relationships within the visual modality, as illustrated in Fig.~\ref{fig:method_full2}b.

Let the current epoch be $e$ and the total number of epochs be $E$. Let the label list for the current batch of samples be $L$. The fused relationship matrix $A^{fuse}$, which integrates the visual modality similarity $A^v$ and the text modality similarity $A^t$, is defined as follows:
\begin{equation}
\label{eq:fuse_matrix}
\setlength{\abovedisplayskip}{3pt}
\setlength{\belowdisplayskip}{3pt}
A^{fuse} = (1 - \frac{e}{E}) A^v +  \frac{e}{E} A^t[L,L].
\end{equation}
Here, \( A^t[L, L] \) is a matrix derived by slicing \( A^t \) according to the label list \( L \), resulting in a matrix that is the same size as \( A^v \). Each element in this matrix, such as \( (A^t[L, L])_{i,j} \), represents the similarity relationship between the classes of sample \( i \) and sample \( j \) in the text feature space, corresponding to \( A^v_{i,j} \) in the visual feature space. Subsequently, \( A^{\text{fuse}} \) is used to guide the learning of relationships within the visual modality:
\begin{equation}
\label{eq:ra_loss}
\setlength{\abovedisplayskip}{0pt}
\setlength{\belowdisplayskip}{0pt}
\mathcal{L}_{\mathrm{ra}}= D_{KL}(A^v||A^{fuse}).
\end{equation}
Here, \( D_{KL} \) represents the KL Divergence~\cite{ji2020kullback}. At the initial stage of training, when \( e/E \approx 0 \), \( A^{\text{fuse}} \approx A^v \). Therefore, $\mathcal{L}_{\mathrm{ra}}\approx D_{KL}(A^v||A^v)$, which constrains the value of \( A^v \) to remain unchanged. This has a similar effect to $\mathcal{L}_{\mathrm{ad}}$: it suppresses visual learning by inhibiting changes in the relationships between visual features. As the number of training epochs increases, \( A^{\text{fuse}} \) gradually incorporates the relational information from the text features, guiding \( A^v \) to approximate the distribution of \( A^t \). This alignment ensures that the similarity relationships between visual features become consistent with those between text features.
%\vspace{-0.2cm}
\subsection{Two Phase Training}
%\vspace{-0.1cm}
It is important to note that visual learning is restricted only during the early epochs of fine-tuning (see Section 6.5 for a more detailed discussion). Smaller hyperparameters, \(\lambda\) and \(\beta\), are used to control the contributions of \(\mathcal{L}_{\text{ad}}\) and \(\mathcal{L}_{\text{ra}}\), ensuring that the restriction on visual learning acts as a disturbance rather than dominating the entire training procedure. In subsequent epochs, we remove \(\mathcal{L}_{\text{ad}}\) and \(\mathcal{L}_{\text{ra}}\) to allow for visual learning. As a result, the fine-tuning loss function ($\mathcal{L}$) can be expressed as:
% Notably, visual learning is restricted only during the early epochs of fine-tuning (see Section 6.4). Smaller hyperparameters, \(\lambda\) and \(\beta\), control the contributions of \(\mathcal{L}_{\text{ad}}\) and \(\mathcal{L}_{\text{ra}}\), ensuring this restriction acts as a disturbance rather than dominating training. In later epochs, \(\mathcal{L}_{\text{ad}}\) and \(\mathcal{L}_{\text{ra}}\) are removed to allow visual learning. The fine-tuning loss function (\(\mathcal{L}\)) can be expressed as:
\begin{equation}
\setlength{\abovedisplayskip}{3pt}
\setlength{\belowdisplayskip}{3pt}
\mathcal{L}= \begin{cases}\mathcal{L}_{\text {vlm}} + \beta \mathcal{L}_{\text {ra}}+\lambda \mathcal{L}_{\text {ad }} & \text { (initial epochs) } \\ \mathcal{L}_{\text {vlm}} & \text { (later epochs) }\end{cases}
\end{equation}

Our method can be \textbf{easily applied to any model with just a few lines of code.} See pseudocode in the Appendix.

%% file: Tex/05_exp.tex
\section{Experiments}
%\vspace{-0.1cm}
\subsection{Datasets and Implementation Details}
%\vspace{-0.1cm}
\noindent
\textbf{Datasets.}
Following previous works~\cite{yazdanpanah2022visual,zhuo2024prompt,xu2024step,radford2021learning}, we do not utilize source domain datasets and finetune our model directly on the target domain. For the target domain, we utilize CropDisease~\cite{mohanty2016using}, EuroSAT~\cite{helber2019eurosat}, ISIC~\cite{codella2019skin}, and ChestX~\cite{wang2017chestx}, which are cross-domain datasets from the domains of agriculture, remote sensing, and medical data with significant domain gaps.

\noindent
\textbf{Implementation Details.}
We adopt the ViT-Base/16 network as the backbone, with parameters pre-trained by CLIP~\cite{radford2021learning}, SigLIP2~\cite{tschannen2025siglip}, and PE-Core~\cite{bolya2025perception}. 
%For the parameter $\lambda$, we consider two cases: when $\mathcal{L}_{\text{ad}}$ is used for the visual branch (e.g., Maple, CLIP-LoRA), $\lambda$ is set to 0.1; when used for the text branch (e.g., CoOp), $\lambda$ is set to 0.001. 
We set $\lambda$ to 0.1 when $\mathcal{L}_{\text{ad}}$ is used for the visual branch (e.g., Maple, CLIP-LoRA), to 0.001 when $\mathcal{L}_{\text{ad}}$ is used for the text branch (e.g., CoOp).
We set $\beta$ to 3. \textbf{See hyperparameter experiments in the Appendix.}
%The parameter $\beta$ is set to 0.5 in all cases. 
For each episode, following \cite{xu2024step}, we perform data augmentation on the support samples. Subsequently, we train each model for 250 epochs~\cite{zanella2024low}, with the first 3/5 epochs (150 epochs, \textbf{see the related ablation study in the Appendix}) used as the initial training phase.
%We evaluate every model using 15 query samples per class, randomly selecting 800 episodes, reporting the results (\%) with the 95\% confidence interval. 
%A single NVIDIA GeForce RTX 3090 is used for training and testing.

%\vspace{-0.1cm}
\subsection{Comparison with the SOTAs}
%\vspace{-0.1cm}
% GNN, FWT, LRP,  Fine-tune, NSAE, BSR
We present the performance of the most representative CDFSL task models~\cite{hu2022adversarial,fu2022wave,fu2023styleadv,zhao2023dual,hu2022pushing,zou2024flatten,zou2025closer,zouattention,yazdanpanah2022visual,xu2024enhancing,xu2024step,zhou2022learning,khattak2023maple,zanella2024low,zhang2021tip,tang2024amu,huang2024lp++,li2025logits}. 
% These models are under different settings, including varying backbones, the use of a source dataset (Source), and whether fine-tuning on the target domain is applied (FT). 
% Specifically, the methods include AFA~\cite{hu2022adversarial}, wave-SAN~\cite{fu2022wave}, StyleAdv~\cite{fu2023styleadv}, StyleAdv-FT (fine-tuned StyleAdv), DARA~\cite{zhao2023dual}, PMF~\cite{hu2022pushing}, FLoR~\cite{zou2024flatten}, CD-CLS~\cite{zou2025closer}, AttnTemp~\cite{zouattention}, VDB~\cite{yazdanpanah2022visual}, IM-DCL~\cite{xu2024enhancing}, StepSTP~\cite{xu2024step}, CoOp~\cite{zhou2022learning}, Maple~\cite{khattak2023maple}, and CLIP-LoRA~\cite{zanella2024low}, which are introduced as our competitors.

% Our method, as an extension, can be applied to existing models. Therefore, we selected three existing models—CoOp~\cite{zhou2022learning}, Maple~\cite{khattak2023maple}, and CLIP-LoRA~\cite{zanella2024low}—to integrate our method. These models cover all possible scenarios in the fine-tuning process: training only the text branch (CoOp), training only the visual branch (CLIP-LoRA-Vision), and training all branches (Maple). 
Our method, as an extension, can be applied to existing models. We selected three models that cover all possible scenarios in the fine-tuning: training only the text branch (CoOp~\cite{zhou2022learning}), training only the visual branch (CLIP-LoRA-Vision~\cite{zanella2024low}), and training all branches (Maple~\cite{khattak2023maple}) to demonstrate the effectiveness of our approach.

As shown in Tab.~\ref{tab:cdfsl}, our method effectively improves performance across four datasets. Our method achieves new SOTA on CLIP~\cite{radford2021learning}, SigLIP~\cite{tschannen2025siglip}, and PE-Core~\cite{bolya2025perception}. Also, our method can be applied to the text branch, treating each prompt as a sample. Applying our method to CoOp (CoOp + OURS) also improves its performance.
% The CLIP-LoRA-Vision + OURS model achieves new state-of-the-art performance on average results across the four CDFSL datasets in the CLIP baseline. Meanwhile, PE-Core-LoRA + OURS achieves the best CDFSL performance. 

\begin{figure}[!t]
\centering
%\vspace{-0.1cm} 
% \setlength{\abovecaptionskip}{0cm}
\centering
\includegraphics[width=0.95 \linewidth]{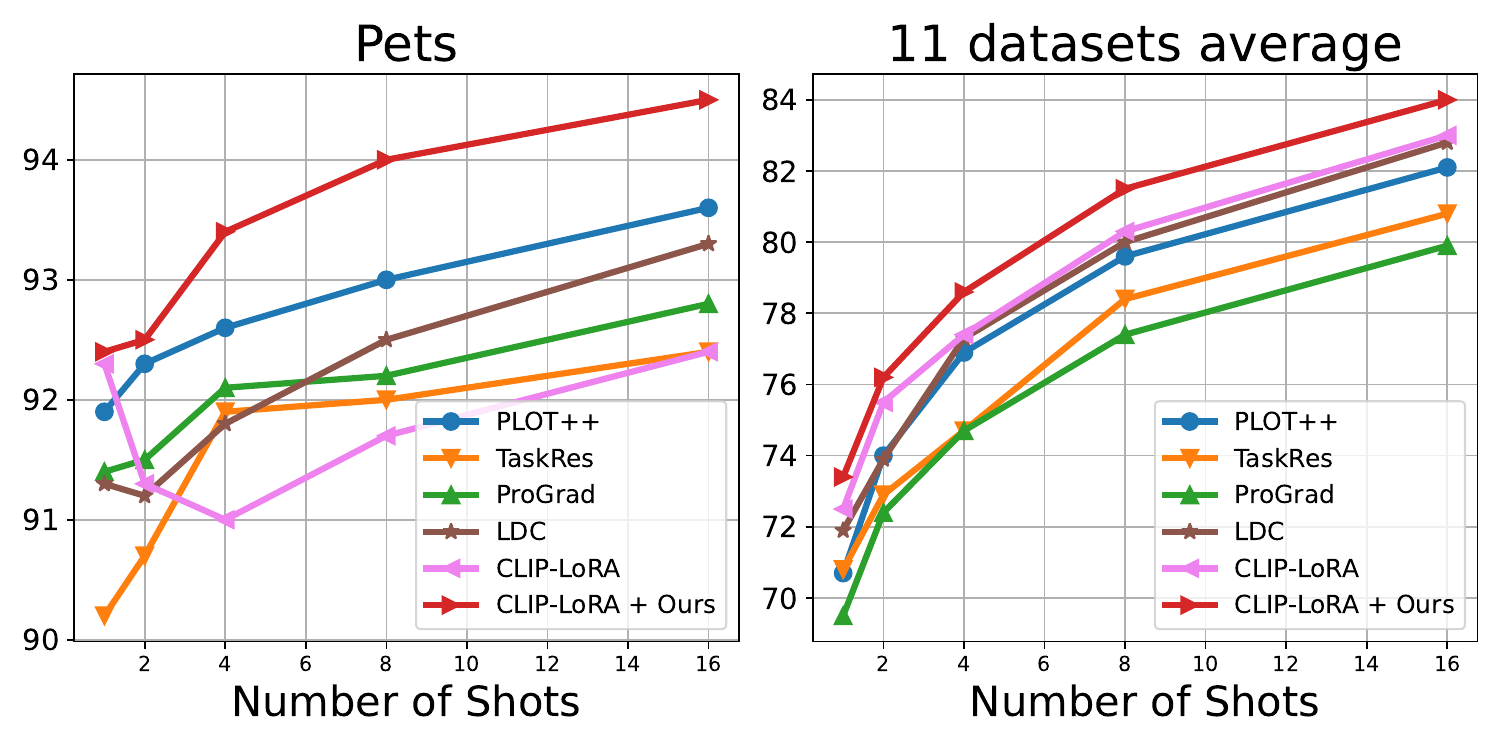}
%%\vspace{-0.3cm} 
\caption{Average few-shot learning result of 11 datasets.}
\label{fig:exp_11_datasets}
%%\vspace{-0.5cm}  
\end{figure}

Further more, following the settings of previous works \cite{zanella2024low,zhu2023prompt,li2025logits}, we test our method on FSL tasks across 11 commonly used classification datasets~\cite{xiao2010sun,maji2013fine,helber2019eurosat,krause20133d,bossard2014food,parkhi2012cats,nilsback2008automated,fei2004learning,cimpoi2014describing,soomro2012ucf101,deng2009imagenet}. As shown in Fig.~\ref{fig:exp_11_datasets}, our method achieves optimal performance across various shot settings. Detailed results are provided in the Appendix.

%%\vspace{-0.1cm}
\subsection{Better Modality Alignment}
%%\vspace{-0.1cm}
The results in Tab~\ref{tab:gap_story} and~\ref{tab:cdfsl} (higher Acc and lower Gap), Fig~\ref{fig:gap_shift_exp} (closer to optimal alignment), and Fig~\ref{fig:heat_map} (better attention scope and higher similarity) confirm that our methods enhance modality alignment. Specifically, Tab.~\ref{tab:gap_story} shows that adding $\mathcal{L}_{\mathrm{ad}}$ and $\mathcal{L}_{\mathrm{ra}}$ progressively improves alignment (higher Acc and lower Gap), validating the effectiveness of both modules for enhancing modality alignment.
%\vspace{-0.05cm}
\subsection{Ablation Study}
%\vspace{-0.1cm}
% We conducted ablation experiments to validate the effectiveness of each module in our method. Our approach consists of two main modules: Suppressing Visual Learning (SVL) and Relationship Alignment (RA). 
% We selected CLIP-LoRA-Vision as the base method and progressively added our modules. The ablation experiments were conducted under the 5-way 1-shot setting on four CDFSL datasets, and the results are presented in Table~\ref{tab:gap_story} and~\ref{tab: ablation}.  As shown, all modules in our method effectively improve the model's performance across the four datasets. When both the SVMD and ISM modules are used together, the model achieves the best performance. \textbf{Also, the modules we proposed effectively facilitate the cross-modal alignment process}, as shown in Table~\ref{tab:gap_story} and Figure~\ref{fig:heat_map}. Using our method improves the cross-modal classification performance while reducing the Loss Gap, indicating that the model fine-tuned with our method achieves good modality alignment. Furthermore, our method enables the model to focus on the appropriate scope. \textbf{ Please refer to the Appendix for additional experiments, including hyperparameter experiments and experiments on alternative strategies for the SVMD and ISM modules.}
We conduct ablation experiments under the 5-way 1-shot setting on four CDFSL datasets, as shown in Tab.~\ref{tab:gap_story} and~\ref{tab:ablation}. All modules in our method improve performance, with the best results achieved when all modules are used together. \textbf{Hyperparameter and alternative strategy tests for the SVL and RA modules, are in the Appendix}.

\begin{table}[t]
\centering
\caption{Ablation study on 5-way 1-shot task.}%\vspace{-0.2cm}
    \label{tab:ablation} 
    \begin{adjustbox}{max width=0.90\linewidth}
    \begin{tabular}{ccccccc}
    \toprule
    SVL & RA  & Cropdisease & EuroSAT & ISIC & ChestX & Avg\\
    \midrule
    \midrule
    - & -  & 84.6 & 81.7 & 36.4 & 21.8 & 56.07\\
    \midrule
    $\checkmark$ & - & 86.4  &83.8  & 37.6 & 22.4  & 57.55\\
    \midrule
    - & $\checkmark$  &85.9  & 83.2 & 37.4 & 22.2 & 57.17\\
    \midrule
    $\checkmark$ & $\checkmark$   & 87.2  & 85.0 & 38.1 & 22.7 &58.26\\
    % \midrule
    % $\checkmark$ & $\checkmark$ & $\checkmark$ & $\checkmark$ & $\checkmark$ & $\mathbf{81.05 \pm 0.58}$ & $\mathbf{87.95 \pm 0.34}$ \\
    \bottomrule
    \end{tabular}
    \end{adjustbox} 
%%\vspace{-0.2cm}
\end{table}

\begin{table}[t]
\centering
    \caption{Add $L_{ad}$ in different finetune phase.}%\vspace{-0.2cm}
    \label{tab:ablation_phase} 
    \begin{adjustbox}{max width=0.90\linewidth}
    \begin{tabular}{cccccc}
    \toprule
    Disturb Phase  & Cropdisease & EuroSAT & ISIC & ChestX & Avg\\
    \midrule
    \midrule
    No  &84.62 &81.72 &36.40 &21.86 &56.07\\
    \midrule
    Begin & 86.40	&83.82	&37.62	&22.38 &57.55\\
    \midrule
    Middle & 86.02	&82.77	&37.36	&22.04 &57.04\\
    \midrule
    Last & 84.63 & 81.27	&36.56	&21.52 &55.99\\
    \midrule
    All &85.73	&83.80	&36.63	&22.24 &57.10\\
    % \midrule
    % $\checkmark$ & $\checkmark$ & $\checkmark$ & $\checkmark$ & $\checkmark$ & $\mathbf{81.05 \pm 0.58}$ & $\mathbf{87.95 \pm 0.34}$ \\
    \bottomrule
    \end{tabular}
    \end{adjustbox} 
%%\vspace{-0.2cm}
\end{table}

\begin{table}[t]
\centering 
\caption{Model overhead.}
\label{tab:overhead}
%\vspace{-0.2cm}
\begin{adjustbox}{max width=0.90\linewidth}
\begin{tabular}{*{4}{c}}
\hline
Method & Parameter &FLOPs &Avg Acc \\
\midrule
\midrule
CLIP-LoRA  & 0.884736 M & 120.149333 G  & 56.07\\
CLIP-LoRA + OURS & 0.884761 M & 120.149359 G  & 58.26 \\
\hline
$\triangle$ & $\uparrow$ 0.0028\% & $\uparrow$ 0.000021\% & $\uparrow$ 3.9\% \\
\hline
\end{tabular}
\end{adjustbox}
%%\vspace{-0.3cm}
\end{table}

%%\vspace{-0.1cm}
\subsection{When to Disturb?}
%%\vspace{-0.2cm}

% In our approach, we suppress visual learning during the fine-tuning process by incorporating $L_{ad}$. 
% An important question is: at which stage of fine-tuning should visual learning be suppressed? To investigate this, we add $L_{ad}$ at different stages of fine-tuning. We fix the total number of fine-tuning epochs at 250 and conduct experiments under the following conditions: 1) \textbf{No}, where $L_{ad}$ is not applied during fine-tuning; 2) \textbf{Begin}, where$L_{ad}$ is introduced during the first 150 epochs (first 3/5 epochs); 3) \textbf{Middle}, where $L_{ad}$ is applied between the 50th and 200th epochs (middle 3/5 epochs); 4) \textbf{Last}, where $L_{ad}$ is added after the 100th epoch (last 3/5 epochs); and 5) \textbf{All}, where $L_{ad}$ is applied throughout the fine-tuning process. 

% The results, presented in Table~\ref{tab:ablation_phase}, demonstrate that the \textbf{Begin} strategy yields the best performance, consistent with the configuration of our method. A comparison of the \textbf{Begin}, \textbf{Middle}, and \textbf{Last} strategies shows that $L_{ad}$ should be added as early as possible, with later application leading to degraded performance. Notably, the \textbf{Last} strategy performs even worse than \textbf{No}, suggesting that discriminative learning remains crucial in the final stages of fine-tuning.

In our approach, we suppress visual learning during fine-tuning using $\mathcal{L}_{\text{ad}}$. To find the best stage for suppression, we try to add $\mathcal{L}_{\text{ad}}$ at different fine-tuning stages over 250 epochs: 1) \textbf{No:} don't use \(L_{ad}\); 2) \textbf{Begin:} the first 150 epochs (first 3/5 epochs); 3) \textbf{Middle:} 50th to 200th epochs (middle 3/5 epochs); 4) \textbf{Last:} after the 100th epoch (last 3/5 epochs); and 5) \textbf{All:} throughout the fine-tuning.

Results in Tab.~\ref{tab:ablation_phase} show \textbf{Begin} performs best, consistent with our method. The \textbf{Last} strategy performs worse than \textbf{No}, highlighting the importance of visual learning in later stages. This aligns with Fig.~\ref{fig:disc_ana}, where visual learning is strongest early in fine-tuning and should be suppressed then.

%%\vspace{-0.1cm}
\subsection{Overhead}
%%\vspace{-0.1cm}
As shown in Tab.~\ref{tab:overhead}, our method significantly improves performance with \textbf{virtually no additional overhead}.

%% file: Tex/06_conclusion.tex
\section{Conclusion}
%%\vspace{-0.25cm}
%In this paper, we observe the misalignment of modalities in CLIP under cross-domain scenarios and reveal that the cross-entropy-based fine-tuning process inherently incorporates strong visual discriminative learning. This learning direction acts as a shortcut, allowing the model to reduce the loss without considering cross-modal relationships, which leads to the suppression of cross-modal relationship learning. Based on these findings, we propose two methods to promote cross-modal alignment: perturbing discriminative learning and mapping domain-specific semantics to general semantics. Extensive experiments on four CDFSL benchmarks validate our rationale and effectiveness.

In this paper, we reveal that visual learning during fine-tuning serves as a shortcut to reduce loss without considering cross-modal relationships. We then propose two methods to suppress visual learning and enhance cross-modal alignment. Extensive experiments confirm the effectiveness of our methods.

% In this work, we observed that promoting discriminative learning leads to a decrease performance. We discovered that the CLIP model exhibits poor alignment between modalities in cross-domain settings, and emphasizing discriminative learning further exacerbates the misalignment between modalities. Through an analysis of the optimization direction of CLIP’s parameters, we provide a qualitative explanation for why focusing on discriminative learning results in poor modality alignment. We argue that emphasizing intra-modality relationships suppresses CLIP’s ability to learn inter-modality relationships. However, in further experiments, we find that the cross-entropy-loss-based learning in CLIP already inherently involves strong discriminative learning. We also observe that discriminative learning within a single modality can effectively reduce the cross-entropy loss on CLIP. This discriminative learning becomes a shortcut, causing the model to neglect the crucial inter-modality alignment under cross-domain conditions. Based on these findings, we propose a method that perturbs intra-modality discriminative learning and constructs generalized category features to reinforce the model’s alignment across modalities. Extensive experiments on four datasets demonstrate that our method achieves state-of-the-art performance on SF-CDFSL.

%% file: Tex/07_supp.tex
% \clearpage
% \setcounter{page}{1}

\appendix
~\\

\section{Detailed Related Work}
\subsection{Source-Free Cross-domain Few shot Learning}
 Cross-Domain Few-Shot Learning (CDFSL) aims to train a model on a source domain that can generalize effectively to a target domain with limited examples. Existing methods are typically categorized into two types: meta-learning-based approaches~\cite{fu2022wave,guo2020broader,hu2022adversarial,wang2021cross,zhang2022metadiff} and transfer learning-based approaches~\cite{guo2020broader,liang2021boosting,zhou2023revisiting,zou2024flatten,zou2025closer,zouattention,Hatano2024MMCDFSL}. Source-Free Cross-Domain Few-Shot Learning (SF-CDFSL) introduces a stronger constraint by making source domain data inaccessible. Current SF-CDFSL methods~\cite{yazdanpanah2022visual,zhuo2024prompt,xu2024step} primarily rely on large models, such as CLIP~\cite{radford2021learning}, leveraging their prior knowledge for classification in the target domain. However, these approaches fail to account for the misalignment between modalities when transferring CLIP to cross-domain settings. Moreover, the influence of visual learning on CLIP-based SF-CDFSL tasks remains underexplored.
\subsection{Parameter-Efficient Fine-Tuning}
Efficiently applying Vision-Language Models (VLMs) to downstream tasks is a key research area. A common strategy is parameter-efficient fine-tuning (PEFT), which uses only a few samples from the target task. PEFT adjusts a small number of the VLM’s parameters, allowing the model to adapt to various applications without modifying all pre-trained parameters. PEFT methods can be grouped into three main types: prompt learning, adapters, and LoRA (and its variants). Prompt learning transforms fixed templates into learnable parameters, such as CoOp~\cite{zhou2022learning} , CoCoOp~\cite{zhou2022conditional}, MaPLe~\cite{khattak2023maple}, PLOT~\cite{chen2022plot}, ProGrad~\cite{zhu2023prompt}, PromptSRC~\cite{khattak2023self}, KgCoOp~\cite{yao2023visual}, PCB~\cite{bang2024active}, DynaPrompt~\cite{zehao2025dynaprompt}, TCP~\cite{yao2024tcp} and ATPrompt~\cite{li2025advancing}. Additionally, Customized Ensemble~\cite{lu2023beyond} combines outputs from multiple models for improved performance, and PromptKD~\cite{li2024promptkd} explores knowledge distillation in prompt learning. Adapter-based methods add trainable modules to the original frozen architecture, making fine-tuning easier, such as CLIP-Adapter~\cite{gao2024clip}, Tip-Adapter~\cite{zhang2022tip}, LP++~\cite{huang2024lp++}, AMU-Tuning~\cite{tang2024amu}, LatHAdapter~\cite{zhao2025fine}, MMA~\cite{Yang_2024_CVPR} and LDC~\cite{li2025logits}. Low-Rank Adaptation (LoRA)~\cite{hu2021lora,zanella2024low} fine-tunes the model by adding learnable low-rank matrices while keeping the original parameters fixed. The new weights can be merged with the original ones, and LoRA does not add extra inference time. Various studies have extended LoRA by adapting the rank for each matrix~\cite{valipour2022dylora,zhang2023adalora}, improving its performance~\cite{chavan2023one,kim2024hydra,zi2023delta}, or reducing memory usage through quantization~\cite{dettmers2024qlora,rajabzadeh2024qdylora}

\subsection{Modality Gap and Misalignment}
\cite{liang2022mind} was the first to identify the existence of a modality gap in multimodal models. Some reserch observed that the performance of multimodal models significantly declines when facing significant domain shifts~\cite{shangguan2025cross,jing2020cross}.~\cite{ji2022information} highlighted the cross-modal bias between semantic-guided samples and nonsemantic-guided samples.~\cite{hao2023uncertainty} emphasized the pairwise misalignment of multimodal uncertainties, addressing the one-to-many alignment issue in multimodal video-text retrieval tasks.~\cite{tian2025mind} pointed out that ProtoNet exhibits a gap between prototypes and instances in cross-domain scenarios.~\cite{wang2022toward} attributed generalization errors to the model learning non-aligned features between the source domain and the test data. These existing works identified the issue of modality misalignment in cross-domain scenarios and considered fine-tuning an effective method for realignment. However, our work demonstrates that fine-tuning alone is insufficient for effective realignment in cross-domain scenarios, especially with large domain gaps in the CDFSL task, such as general domains vs. medical domains. We analyze this issue from the perspective of visual learning, which acts as a shortcut during fine-tuning, and address it in this paper.
%\tableofcontents
\section{Proof of Theorem 4.1}
The loss in CLIP is designed to align image-text pairs while contrasting mismatched pairs. Below is the step-by-step derivation of the gradient for the visual feature $f_i$. 
For a batch of $N$ image-text pairs, the loss for the $i$-th image is:
\begin{equation}
\mathcal{L}_i = -\log \frac{e^{f_i \cdot t_i / \tau}}{\sum_{k=1}^N e^{f_i \cdot t_k / \tau}},
\end{equation}
where:
$f_i$: L2-normalized visual feature of the $i$-th image ($\|f_i\| = 1$).
$t_k$: L2-normalized text feature of the $k$-th text ($\|t_k\| = 1$).
$\tau$: Temperature coefficient (e.g., $\tau = 0.01$).

Expand $\mathcal{L}_i$:
\begin{equation}
\begin{aligned}
    \mathcal{L}_i &= -\underbrace{\log \left( e^{f_i \cdot t_i / \tau} \right)}_{\text{postive}} + \underbrace{\log \left( \sum_{k=1}^N e^{f_i \cdot t_k / \tau} \right)}_{\text{negative}} \\ &= -\frac{f_i \cdot t_i}{\tau} + \log \left( \sum_{k=1}^N e^{f_i \cdot t_k / \tau} \right).
\end{aligned}
\end{equation}

The gradient $\nabla_{f_i} \mathcal{L}_i$ has two terms from the loss components:

\begin{equation}
\begin{aligned}
    \nabla_{f_i} \mathcal{L}_i &= -\nabla_{f_i} \left( \frac{f_i \cdot t_i}{\tau} \right) + \nabla_{f_i} \left( \log \left( \sum_{k=1}^N e^{f_i \cdot t_k / \tau} \right) \right) \\
    &= - \frac{t_i}{\tau} + \frac{1}{\sum_{m=1}^N e^{f_i \cdot t_m / \tau}} \cdot \sum_{k=1}^N \nabla_{f_i} \left( e^{f_i \cdot t_k / \tau} \right) \\
    &= - \frac{t_i}{\tau} + \frac{\sum_{k=1}^N e^{f_i \cdot t_k / \tau} \cdot \frac{t_k}{\tau}}{\sum_{m=1}^N e^{f_i \cdot t_m / \tau}} \\
    &= - \frac{t_i}{\tau} + \frac{1}{\tau} \sum_{k=1}^N p_{ik} t_k
\end{aligned}
\end{equation}
where $p_{ik}$ is the softmax probability: $p_{ik} = \frac{e^{f_i \cdot t_k / \tau}}{\sum_{m=1}^N e^{f_i \cdot t_m / \tau}}$. Let $\mathbf{f}_i^{\text{new}}$ and $\mathbf{f}_k^{\text{new}}$ represent the visual features after parameter updates. The term $\mathbf{f}_i^{\text{new}}\cdot \mathbf{f}_k^{\text{new}}$ denotes the similarity between samples $i$ and $k$ after the parameter update, then:

\begin{equation}
    \mathbf{f}_i^{\text{new}} = \mathbf{f}_i + \eta \frac{1}{\tau} \left( \mathbf{t}_i - \sum_{j=1}^{N} p_{ij} \mathbf{t}_j \right),   \mathbf{f}_k^{\text{new}} = \mathbf{f}_k + \eta \frac{1}{\tau} \left( \mathbf{t}_k - \sum_{j=1}^{N} p_{kj} \mathbf{t}_j \right)
\end{equation}

\begin{equation}
\begin{aligned}
\mathbf{f}_i^{\text{new}} \cdot \mathbf{f}_k^{\text{new}} &= (\mathbf{f}_i + \eta \frac{1}{\tau} \left( \mathbf{t}_i - \sum_{j=1}^{N} p_{ij} \mathbf{t}_j \right)) \cdot (\mathbf{f}_k + \eta \frac{1}{\tau} \left( \mathbf{t}_k - \sum_{j=1}^{N} p_{kj} \mathbf{t}_j \right)) \\
&=\mathbf{f}_i \cdot \mathbf{f}_k + \eta \frac{1}{\tau} \left( \mathbf{f}_i \cdot (\mathbf{t}_k - \sum_{j=1}^{N} p_{kj} \mathbf{t}_j) + \mathbf{f}_k \cdot (\mathbf{t}_i - \sum_{j=1}^{N} p_{ij} \mathbf{t}_j) \right) + O(\eta^2),
\end{aligned}
\end{equation}

where $\eta$ is the learning rate and $\tau$ is the temperature coefficient. Let the difference in cosine similarity between two samples, $\mathbf{f}_i$ and $\mathbf{f}_k$, before and after one step of training is $\Delta \cos(\theta_{ik}) = \mathbf{f}_i^{\text{new}}\cdot\mathbf{f}_k^{\text{new}} -  \mathbf{f}_i \cdot \mathbf{f}_k$, there is:
\begin{equation}
\begin{aligned}
    \Delta \cos(\theta_{ik}) &= \mathbf{f}_i^{\text{new}} \cdot \mathbf{f}_k^{\text{new}} -\mathbf{f}_i \cdot \mathbf{f}_k \\
    &=  \eta \frac{1}{\tau} \left( \mathbf{f}_i \cdot \mathbf{t}_k - \sum_{j=1}^{N} p_{kj} \mathbf{f}_i \cdot \mathbf{t}_j + \mathbf{f}_k \cdot \mathbf{t}_i - \sum_{j=1}^{N} p_{ij} \mathbf{f}_k \cdot\mathbf{t}_j \right) + O(\eta^2)
\end{aligned}
\end{equation}

\textbf{ When $\mathbf{f}_i$ and $\mathbf{f}_k$ belong to the same class}, $\mathbf{t}_i = \mathbf{t}_k$. Considering that the learning objective of the cross-entropy loss function satisfies: $\mathbf{f}_i\cdot\mathbf{t}_i > \mathbf{f}_i\cdot\mathbf{t}_j, \forall j \neq i$. there is:
\begin{equation}
\begin{aligned}
\Delta \cos(\theta_{ik}) &=  \eta \frac{1}{\tau} \left( \mathbf{f}_i \cdot \mathbf{t}_i - \sum_{j=1}^{N} p_{kj} \mathbf{f}_i \cdot \mathbf{t}_j + \mathbf{f}_k \cdot \mathbf{t}_k - \sum_{j=1}^{N} p_{ij} \mathbf{f}_k \cdot\mathbf{t}_j \right) + O(\eta^2) \\
&> \eta \frac{1}{\tau} \left( \mathbf{f}_i \cdot \mathbf{t}_i - (\sum_{j=1}^{N} p_{kj}) \mathbf{f}_i \cdot \mathbf{t}_i + \mathbf{f}_k \cdot \mathbf{t}_k - (\sum_{j=1}^{N} p_{ij}) \mathbf{f}_k \cdot\mathbf{t}_k \right) \\
&= \eta \frac{1}{\tau} \left( \mathbf{f}_i \cdot \mathbf{t}_i - 1\cdot \mathbf{f}_i \cdot \mathbf{t}_i + \mathbf{f}_k \cdot \mathbf{t}_k - 1\cdot\mathbf{f}_k \cdot\mathbf{t}_k \right) =0.
\end{aligned}
\end{equation}
That is, the visual features between samples $i$ and $k$ of the same class will become increasingly similar.

\textbf{When $\mathbf{f}_i$ and $\mathbf{f}_k$ belong to different classes}, we analyzed the values of $\Delta \cos(\theta_{ik})$ during the training process across 200 episodes, as shown in Figure~\ref{fig:delta_cos}. As seen, in different VLMs (CLIP~\cite{radford2021learning} and PE-Core~\cite{bolya2025perception}), the $\Delta \cos(\theta_{ik})$ is always less than 0., which means that the visual features of samples $i$ and $k$ from different classes will become increasingly dissimilar.

\textbf{Summary:} during fine-tuning, visual features of the same category cluster together while those of different categories are pushed apart, which means visual learning is consistently present in the fine-tuning process.

\begin{figure*}[!t]
%%%\vspace{-1.2cm}
%\centering
%\includegraphics[width=3in]{fig5}
\subfloat[$\Delta \cos(\theta_{ik})$ in fine-tuning]{
\label{fig:delta_cos}
%\centering
\includegraphics[width=0.25\linewidth]{./Figs/disc_in_clip/clip_deta_cos.pdf}
\includegraphics[width=0.25\linewidth]{./Figs/disc_in_clip/pe_deta_cos.pdf}
} 
\subfloat[Hyperparameter study]{
\label{fig:hyper}
%\centering
\includegraphics[width=0.235\linewidth]{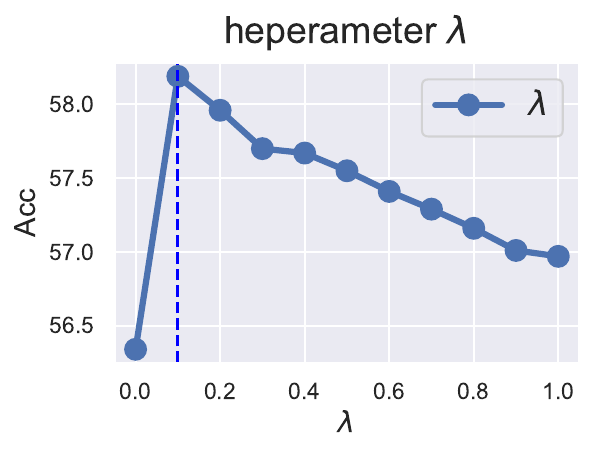}
\includegraphics[width=0.235\linewidth]{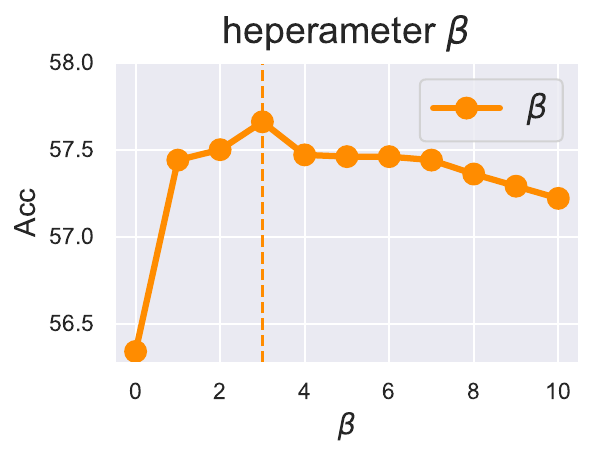}
}
\caption{(a) When sample $i$ and sample $k$ belong to different classes, $\Delta \cos(\theta_{ik})$ in 5-way 1-shot fine-tuning is always less than 0, which means that the visual features of samples $i$ and $k$ from different classes will become increasingly dissimilar. (b) Hyperparameter study.}
\label{fig:delta_cos}
%\vspace{-0.4cm}
\end{figure*}

% \begin{figure}[htbp]
% %%\vspace{-1.2cm}
% \centering
% %\includegraphics[width=3in]{fig5}
% \subfloat{
%         \centering
% 	\label{fig:delta_cos}\includegraphics[width=0.45\linewidth]{./Figs/disc_in_clip/delta_cos.jpg}}
% \subfloat{
%         \centering
% 	\label{fig:hyper}\includegraphics[width=0.48\linewidth]{./Figs/supp/hyper.jpg}}
% \caption{(When sample $i$ and sample $k$ belong to different classes, $\Delta \cos(\theta_{ik})$ in 5-way 1-shot fine-tuning is always less than 0.}
% %%\vspace{-0.4cm}
% \end{figure}
% \begin{figure}[htbp]
% %%\vspace{-1.2cm}
% \centering
% %\includegraphics[width=3in]{fig5}
% \subfloat[]{
%         \centering
% 	\label{fig:delta_cos}\includegraphics[width=0.45\linewidth]{./Figs/disc_in_clip/delta_cos.jpg}}
% \subfloat[]{
%         \centering
% 	\label{fig:hyper}\includegraphics[width=0.48\linewidth]{./Figs/supp/hyper.jpg}}
% \caption{(a) When sample $i$ and sample $k$ belong to different classes, $\Delta \cos(\theta_{ik})$ in 5-way 1-shot fine-tuning is always less than 0. (b) Hyperparameter study.}
% %%\vspace{-0.4cm}
% \end{figure}

\section{Hyperparameter Study}

Our method involves two main hyperparameters: $\lambda$, corresponding to the Suppressing Visual Learning (SVL) module, and $\beta$, corresponding to the Relationship Alignment (RA) module. Here, we present the 5-way 1-shot performance corresponding to different values of $\lambda$ and $\beta$ on the four CDFSL datasets, as shown in Figure~\ref{fig:hyper}. As observed, the model achieves peak performance when $\lambda$ is set to 0.1 and $\beta$ to 3. A smaller value for $\lambda$ is preferred, as $\mathcal{L}_{ad}$ functions as a perturbation and should be assigned a small weight to prevent it from dominating the training process.

\section{Alternative Strategies for the SVL Module}
In this section, we discuss several alternative strategies for the SVL modules. The SVL module is designed to perturb the visual learning, thereby guiding the model to focus more on learning cross-modal relationships. Our approach involves randomly sampling support samples to create class-shuffle weights and performing the classification task on the visual branch (as described in Section 5.1 of the main text). To demonstrate the effectiveness of this strategy, we also test three other approaches: \textbf{NO}, which does not use the SVL module and serves as the baseline, and \bm{$-\mathcal{L}_v$}, which use the negative of $\mathcal{L}_v$ as $\mathcal{L}_{ad}$, where $\mathcal{L}_{ad} = - \mathcal{L}_{v}$ (as defined in Equation 3 of the paper). Additionally, \textbf{Noise Proto} generates class-shuffle weights using random initialization (non-true visual features) and then performs the classification task on the visual branch.

%\vspace{0.10cm}
\begin{minipage}[t]{\textwidth}
%\centering
\begin{minipage}[t]{0.48\textwidth}
\makeatletter\def\@captype{table}
%\centering
    \caption{Alternative strategies for the SVL module.}
    \label{tab:abl_svmd} 
    \begin{adjustbox}{max width=0.95\linewidth}
    \begin{tabular}{cccccc}
    \toprule
    Disturb Strategy  & Cropdisease & EuroSAT & ISIC & ChestX & Avg\\
    \midrule
    No  &84.62 &81.72 &36.40 &21.86 &56.07\\
    \midrule
    $-\mathcal{L}_v$ & 82.41	&78.84	&34.72	&21.37 &54.33\\
    \midrule
    Noise Proto & 84.31	&82.42	&35.96	&21.80 &56.12\\
    \midrule
    \textbf{Ours} & \textbf{86.41} &\textbf{83.80}	&\textbf{37.63}	&\textbf{22.38} &\textbf{57.55}\\
    \bottomrule
    \end{tabular}
    \end{adjustbox} 
\end{minipage}
\medskip
\begin{minipage}[t]{0.48\textwidth}
\makeatletter\def\@captype{table}
%\centering
\caption{Alternative strategies for the RA module.}
    \label{tab:abl_ra} 
    \begin{adjustbox}{max width=0.98\linewidth}
    \begin{tabular}{cccccc}
    \toprule
    Alignment Strategy  & Cropdisease & EuroSAT & ISIC & ChestX & Avg\\
    \midrule
    No  &84.62 &81.72 &36.40 &21.86 &56.07\\
    \midrule
    only vision & 84.87	&82.61	&36.94	&22.11 &56.63\\
    \midrule
    only text & 84.14	&81.36	&36.87	&21.34 &55.92\\
    \midrule
    \textbf{Ours} & \textbf{85.92} &\textbf{83.21}	&\textbf{37.42}	&\textbf{22.17} &\textbf{57.17}\\
    % \midrule
    % $\checkmark$ & $\checkmark$ & $\checkmark$ & $\checkmark$ & $\checkmark$ & $\mathbf{81.05 \pm 0.58}$ & $\mathbf{87.95 \pm 0.34}$ \\
    \bottomrule
    \end{tabular}
    \end{adjustbox} 
\end{minipage}
%%\vspace{-0.15cm}
\end{minipage}

The results, shown in Table~\ref{tab:abl_svmd}, indicate that the \bm{$-\mathcal{L}_v$} strategy actually harms the model’s performance. Moreover, the \textbf{Noise Proto} strategy brings little or no improvement in performance. This is understandable, as a fundamental requirement for visual learning is that the learning occurs within a valid visual feature distribution. The class-shuffle weights generated by the \textbf{Noise Proto} strategy may not be within this valid distribution. In contrast, our approach ensures the simulation of an effective visual feature distribution while simultaneously suppressing discriminative learning, leading to a significant improvement in model performance.

\section{Alternative Strategies for the RA Module}
In this section, we discuss several alternative strategies for the RA modules. The RA module is designed to replace the discriminative visual feature learning direction in the fine-tuning process, providing a new learning direction that facilitates alignment between modalities. Our approach promotes cross-modal alignment by gradually aligning the internal relationships of visual features with those of textual features (as described in Section 5.2 of the main text). To demonstrate the effectiveness of this strategy, we also test three other approaches: \textbf{NO}, which does not use the RA module and serves as the baseline, and \textbf{only vision}, which does not incorporate the internal relationships of the text modality and only maintains the relationships between visual features, i.e., $\mathcal{L}_{\mathrm{ra}}= D_{KL}(A^v||A^v)$ (see Equation 10 in the main text). Additionally, \textbf{only text} does not use the gradual fusion approach (Equation 9 in the main text) but directly aligns the internal relationships of visual features with those of textual features, i.e., $\mathcal{L}_{\mathrm{ra}}= D_{KL}(A^v||A^t)$.

The results, shown in Table~\ref{tab:abl_ra}, indicate that the \textbf{only text} strategy harms the model’s performance. In contrast, our method, which gradually fuses the relationships between visual features \(A_v\) and textual features \(A_t\) to guide the learning of internal visual feature relationships, most effectively promotes alignment between modalities and achieves optimal performance.

\section{Division Between Initial Epochs and Later Epochs}
\begin{wrapfigure}{r}{7.0cm}
%\vspace{-0.6cm}
\centering
\begin{adjustbox}{max width=1.1\linewidth}
\includegraphics[width=1.0\linewidth]{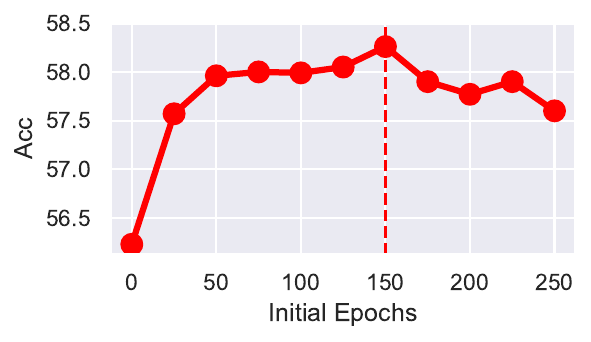}
\end{adjustbox}
%\vspace{-0.8cm}
\caption{Results for different initial epochs settings.}
\label{fig:ad_lenth}
%\vspace{-0.5cm}
\end{wrapfigure}
During the fine-tuning process, we train the model for 250 epochs, with the first 3/5 epochs used as the initial training phase. In this section, we conduct further experiments and analysis on the initial training phase. We test different proportions for the initial training phase division, as shown in Figure~\ref{fig:ad_lenth}. The results indicate that when 150 epochs (the first 3/5 ) is used as the initial training phase, the model achieves optimal performance across four datasets. It is important to note that in this section, we focus only on using the initial epochs for the initial training phase, as Section 6.5 of the main text has already demonstrated the need to suppress visual learning in the early stages.
%\vspace{-0.2cm}
\section{Better Modality Alignment}
\begin{figure*}[t]
\centering
\begin{minipage}[b]{.49\linewidth}
    \centering
    \centering
    \subfloat["a photo of a Strawberry\_healthy."]{
    \includegraphics[width=0.23\linewidth]{./Figs/heatmap/raw_0_61_n.png} 
    \includegraphics[width=0.23\linewidth]{./Figs/heatmap/base_0_61_n.png} 
    \includegraphics[width=0.23\linewidth]{./Figs/heatmap/cls_0_61_n.png} 
    \includegraphics[width=0.23\linewidth]{./Figs/heatmap/ours_0_61_n.png} 
    } 
\end{minipage}
\medskip
\begin{minipage}[b]{.48\linewidth}
    \centering
    \centering
    \subfloat["a photo of a Benign Keratosis."]{
    \includegraphics[width=0.23\linewidth]{./Figs/heatmap/raw_1_43_n.png} 
    \includegraphics[width=0.23\linewidth]{./Figs/heatmap/base_1_43_n.png} 
    \includegraphics[width=0.23\linewidth]{./Figs/heatmap/cls_1_43_n.png} 
    \includegraphics[width=0.23\linewidth]{./Figs/heatmap/ours_1_43_n.png} 
    } 
\end{minipage} \\

\begin{minipage}[b]{.49\linewidth}
    \centering
    \centering
    \subfloat["a photo of a Melanoma."]{
    \includegraphics[width=0.23\linewidth]{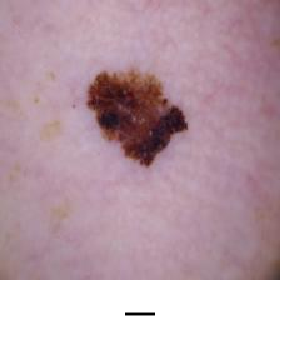} 
    \includegraphics[width=0.23\linewidth]{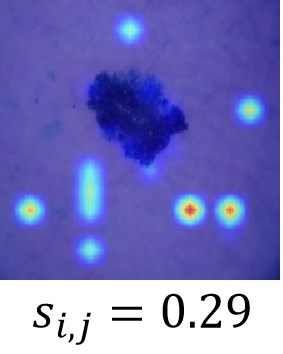} 
    \includegraphics[width=0.23\linewidth]{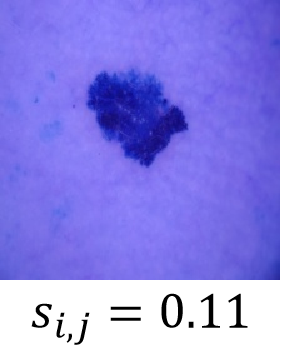} 
    \includegraphics[width=0.23\linewidth]{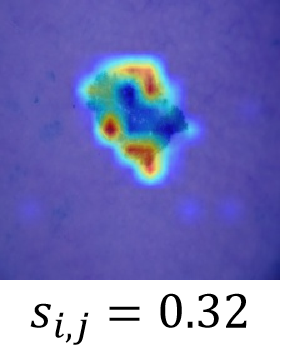} 
    } 
\end{minipage}
\medskip
\begin{minipage}[b]{.49\linewidth}
    \centering
    \centering
    \subfloat["a photo of a Basal Cell Carcinoma."]{
    \includegraphics[width=0.23\linewidth]{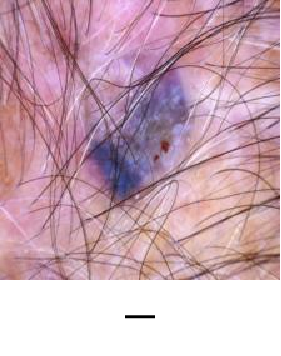} 
    \includegraphics[width=0.23\linewidth]{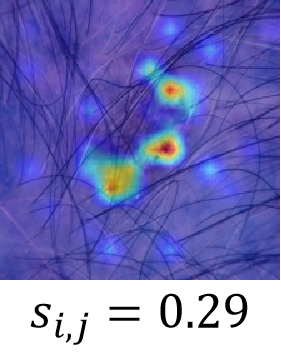} 
    \includegraphics[width=0.23\linewidth]{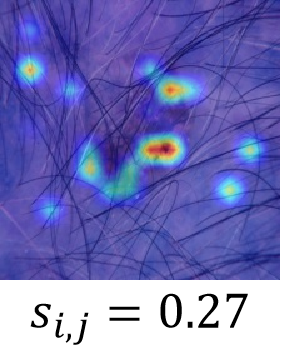} 
    \includegraphics[width=0.23\linewidth]{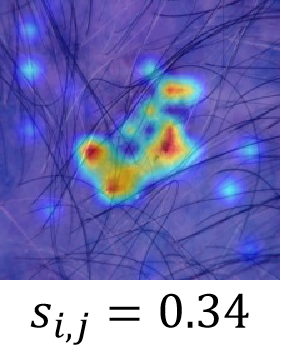} 
    }  
\end{minipage} \\

\begin{minipage}[b]{.49\linewidth}
    \centering
    \centering
    \subfloat["a photo of a Industrial Buildings."]{
    \includegraphics[width=0.23\linewidth]{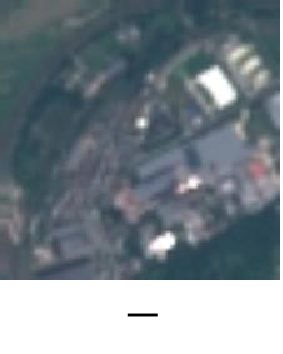} 
    \includegraphics[width=0.23\linewidth]{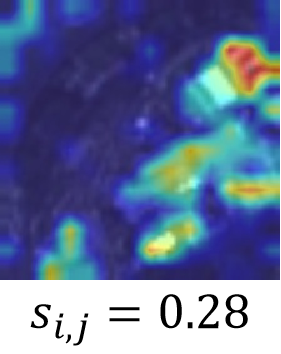} 
    \includegraphics[width=0.23\linewidth]{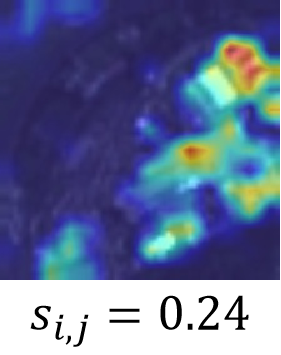} 
    \includegraphics[width=0.23\linewidth]{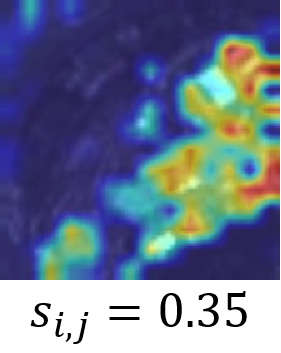} 
    } 
\end{minipage}
\medskip
\begin{minipage}[b]{.49\linewidth}
    \centering
    \centering
    \subfloat["a photo of a Industrial Buildings."]{
    \includegraphics[width=0.23\linewidth]{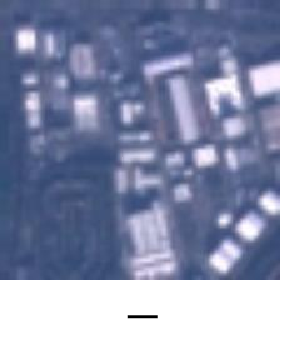} 
    \includegraphics[width=0.23\linewidth]{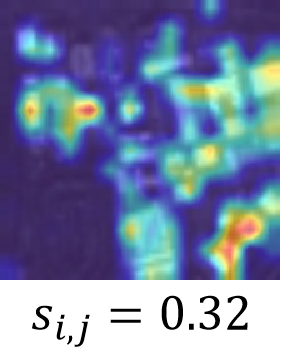} 
    \includegraphics[width=0.23\linewidth]{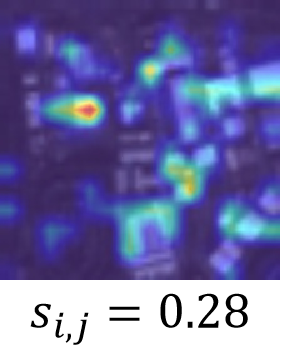} 
    \includegraphics[width=0.23\linewidth]{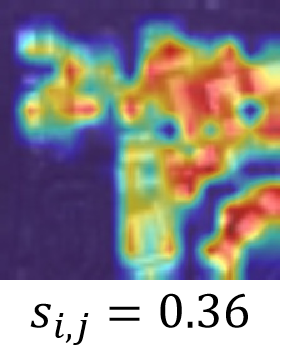} 
    }  
\end{minipage} \\

\begin{minipage}[b]{.49\linewidth}
    \centering
    \centering
    \subfloat["a photo of a Atelectasis."]{
    \includegraphics[width=0.23\linewidth]{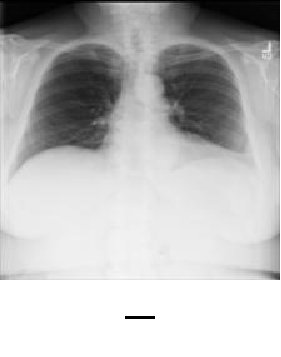} 
    \includegraphics[width=0.23\linewidth]{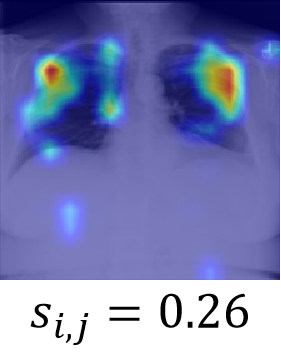} 
    \includegraphics[width=0.23\linewidth]{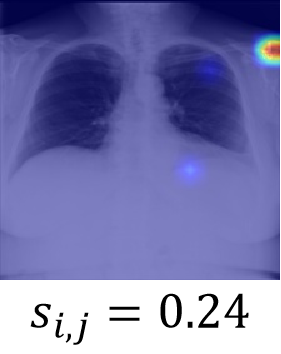} 
    \includegraphics[width=0.23\linewidth]{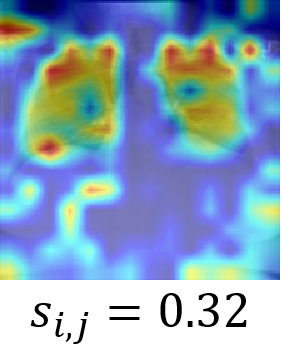} 
    } 
\end{minipage}
\medskip
\begin{minipage}[b]{.49\linewidth}
    \centering
    \centering
    \subfloat["a photo of a Nodule."]{
    \includegraphics[width=0.23\linewidth]{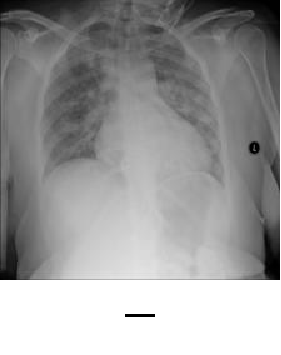} 
    \includegraphics[width=0.23\linewidth]{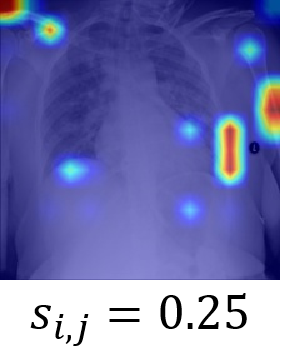} 
    \includegraphics[width=0.23\linewidth]{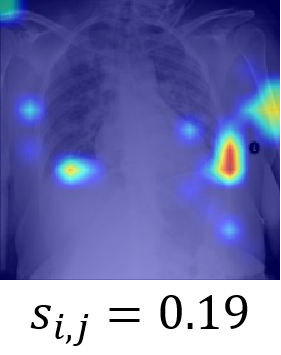} 
    \includegraphics[width=0.23\linewidth]{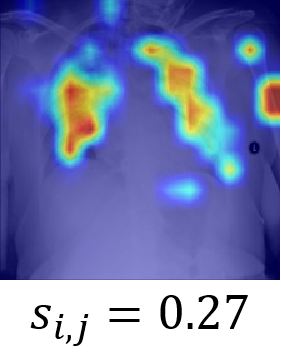} 
    }  
\end{minipage}
%%\vspace{-0.4cm}
\caption{The attention maps of the three models are shown. From left to right: the original image, the Baseline result (trained with $\mathcal{L}_{\mathrm{vlm}}$), the result of the model with enhanced visual learning (trained with $\mathcal{L}_{\mathrm{vlm}} + \mathcal{L}_{\mathrm{v}}$), and the result of our method (trained with $\mathcal{L}_{\mathrm{vlm}} + \mathcal{L}_{\mathrm{ad}}$). Here, $s_{i, j}$ represents the cosine similarity between the image features and the text features. A higher similarity indicates better alignment. It is evident that our model has the most appropriate attention scope and achieves the highest similarity between sample pairs.
}
\label{fig:heat_map2}
%\vspace{-0.3cm}
\end{figure*}

Additionally, the modules we proposed effectively facilitate the cross-modal alignment process. Firstly, as shown in Table~\ref{tab:gap_story}, both of our proposed modules improves cross-modal classification performance while reducing the loss gap, indicating that models fine-tuned with our approach achieve good modality alignment. Secondly, as illustrated in Figure~\ref{fig:heat_map} and Figure~\ref{fig:heat_map2}, our method enables the model to focus on the appropriate scope. In the baseline model, due to its strong visual learning, it tends to focus on the most discriminative small features in the image (second column), which is detrimental to modality alignment. In this case, the visual features are dominated by a small, discriminative part of the image. For example, in the case of "a photo of a Strawberry\_healthy," the most discriminative feature of the strawberry leaf is its serrated edges, but this feature alone cannot fully represent the semantic information of "strawberry leaf." Our method, by suppressing this visual learning, leads to a more generalized model. As seen in the fourth column, after applying our method, the model is able to focus correctly on the image features corresponding to the semantic information. This promotes cross-modal alignment, resulting in a significantly higher similarity between the image and text features.

% \section{Future work}
% In this work, we observe that visual-modal discriminative learning becomes a shortcut during the CLIP fine-tuning process. As a result, we introduce visual-modal perturbations to hinder discriminative learning, encouraging the model's fine-tuning process to focus more on learning cross-modal knowledge, thereby improving overall performance. However, during our experiments, we also found that introducing similar perturbations on the text branch can also enhance performance (CoOp + ours). This discovery motivates us to further explore,  whether a similar phenomenon exists in the text branch.

\section{Extended Results on CDFSL Task}
Table~\ref{tab:cdfsl2} is an extended version of Table~\ref{tab:cdfsl} in the main text. It includes the performance of various models on four CDFSL datasets under different settings. These settings involve different backbones (CLIP~\cite{radford2021learning}, SigLIP2~\cite{tschannen2025siglip}, and PE-Core~\cite{bolya2025perception}), the use of a source dataset (Source), and whether fine-tuning on the target domain is applied (FT).
Specifically, the methods include ATA~\cite{wang2021cross}, AFA~\cite{hu2022adversarial}, wave-SAN~\cite{fu2022wave}, StyleAdv~\cite{fu2023styleadv}, StyleAdv-FT (fine-tuned StyleAdv), DARA~\cite{zhao2023dual}, PMF~\cite{hu2022pushing}, FLoR~\cite{zou2024flatten}, CD-CLS~\cite{zou2025closer}, AttnTemp~\cite{zouattention}, VDB~\cite{yazdanpanah2022visual}, IM-DCL~\cite{xu2024enhancing}, StepSTP~\cite{xu2024step}, CoOp~\cite{zhou2022learning}, Tip-Adapter~\cite{zhang2021tip}, AMU-Tuning~\cite{tang2024amu}, LP++~\cite{huang2024lp++}, LDC~\cite{li2025logits}, Maple~\cite{khattak2023maple}, and CLIP-LoRA~\cite{zanella2024low}, which are introduced as our competitors.

\section{Few Shot Learning Results}
We further evaluate the effectiveness of our method on 11 commonly used classification datasets. Following the setup of previous work~\cite{zhou2022learning,zanella2024low}, we evaluate our approach on 11 datasets covering various fine-grained classification tasks: scenes (SUN397~\cite{xiao2010sun}), aircraft types (Aircraft~\cite{maji2013fine}), satellite imagery (EuroSAT~\cite{helber2019eurosat}), automobiles (StanfordCars~\cite{krause20133d}), food items (Food101~\cite{bossard2014food}), pet breeds (OxfordPets~\cite{parkhi2012cats}), flowers (Flower102~\cite{nilsback2008automated}), general objects (Caltech101~\cite{fei2004learning}), textures (DTD~\cite{cimpoi2014describing}), and human actions (UCF101~\cite{soomro2012ucf101}), in addition to ImageNet~\cite{deng2009imagenet}. These datasets provide a comprehensive benchmarking framework for evaluating few-shot visual classification tasks. 

We compare our model against several prompt-based methods, including CoOp~\cite{zhou2022learning} with 4 learnable tokens, CoOp~\cite{zhou2022learning} with 16 learnable tokens, CoCoOp~\cite{zhou2022conditional}, PLOT++~\cite{chen2022plot} (an adaptation of the original PLOT designed for transformer architectures), KgCoOp~\cite{yao2023visual}, MaPLe~\cite{khattak2023maple}, and ProGrad~\cite{zhu2023prompt} with 16 tokens. Additionally, we evaluate adapter-based methods such as Tip-Adapter-F~\cite{zhang2022tip} and TaskRes~\cite{yu2023task} and LDC~\cite{li2025logits}, as well as the LoRA-based method CLIP-LoRA~\cite{zanella2024low}. These comparative methods provide a comprehensive benchmark to assess the effectiveness of our proposed approach in few-shot visual classification tasks. 

Our method is plug-and-play, allowing for quick integration into existing approaches. As shown in Table~\ref{tab:fewshot_shull}, applying our method to the existing CLIP-LoRA~\cite{zanella2024low} model effectively enhances its few-shot classification performance across 11 datasets. In addition, it achieves the highest average classification accuracy under various shot settings.

\begin{table*}[!htbp]
\belowrulesep=0pt
\aboverulesep=0pt
\caption{
The accuracy(\%) of four target domain datasets under 5-way 1-shot and 5-way 5-shot tasks. The use of a source dataset (Source), and whether fine-tuning on the target domain is applied (FT)}
%\vspace{-0.1cm}
\label{tab:cdfsl2}    
\centering
\begin{adjustbox}{max width=0.90\linewidth}
\begin{tabular}{*{10}{c|c c|c c c c c c c|} }  
\toprule 
 \multirow{1}*{Task}  &  \multirow{1}*{Method}  &\multirow{1}*{backbone} &  \multirow{1}*{Source} &  \multirow{1}*{FT}  & \multirow{1}*{ISIC} & \multirow{1}*{EuroSAT} & \multirow{1}*{CropDisease} & \multirow{1}*{ChestX} & \multirow{1}*{Avg} \\           
 \midrule
%\multirow{14}{*}{\rotatebox{90}{5-way 1-shot}} 
  % &GNN  & RN10 & Y & - & 22.00±0.46 & 32.02±0.66 & 63.69±1.03 & 64.48±1.08 & 45.55 \\
  % &FWT  &RN10 &Y &- &22.04±0.44 &31.58±0.67 &62.36±1.05 &66.36±1.04 &45.59 \\
  % &LRP  &RN10 &Y &- &22.11±0.20 &30.94±0.30 &54.99±0.50 &59.23±0.50 &41.82 \\
  &ATA~\cite{wang2021cross}  &RN10 &Y &-  &33.21±0.40 &61.35±0.50 &67.47±0.50 &22.10±0.20 &46.03 \\
  &AFA~\cite{hu2022adversarial}  &RN10 &Y &-  &33.21±0.30 &63.12±0.50 &67.61±0.50 &22.92±0.20 &46.72 \\
  &wave-SAN~\cite{fu2022wave}  &RN10 &Y &- &33.35±0.71 &69.64±1.09 &70.80±1.06 &22.93±0.49 &49.18 \\
  &StyleAdv~\cite{fu2023styleadv}  &RN10 &Y &- &33.96±0.57 &70.94±0.82 &74.13±0.78 &22.64±0.35 &50.42 \\
  \cline{2-10}
  &ATA-FT~\cite{wang2021cross}  &RN10 &Y &Y &34.94±0.40 &68.62±0.50 &75.41±0.50 &22.15±0.20 &50.28 \\
  &DARA~\cite{zhao2023dual}  &RN10 &Y &Y &36.42±0.64 &67.42±0.80 &80.74±0.76 &22.92±0.40 &51.88 \\
  &StyleAdv-FT~\cite{fu2023styleadv}  &RN10 &Y &Y &35.76±0.52 &72.92±0.75 &80.69±0.28 &22.64±0.35 &53.00 \\
  \cline{2-10}
  \multirow{14}{*}{\rotatebox{90}{5-way 1-shot}} 
  &PMF~\cite{hu2022pushing}  &ViT/DINO &Y &Y &30.36±0.36 &70.74±0.63 &80.79±0.62 &21.73±0.30 &50.91 \\
  &StyleAdv-FT~\cite{fu2023styleadv} &ViT/DINO &Y &Y &33.99±0.46 &74.93±0.58 &84.11±0.57 &22.92±0.32 &53.99 \\
  &FLoR~\cite{zou2024flatten} &ViT/DINO &Y &Y &35.49 &73.09 &83.55 &23.26 & 53.85 \\
  &CD-CLS~\cite{zou2025closer} &ViT/DINO &Y &Y  &35.56 & 74.97 & 84.53 &23.39 & 54.62 \\
  &AttnTemp~\cite{zouattention} &ViT/DINO &Y &Y  &38.05 &75.09 & 84.78 &23.63 & 55.39 \\
  \cline{2-10}
  &FN+VDB~\cite{yazdanpanah2022visual}  &RN18 &- &Y &32.96±0.57 &69.67±0.80 &79.68±0.74 &22.64±0.41 &51.24 \\
  &IM-DCL~\cite{xu2024enhancing}  &RN10 &- &Y &38.13±0.57 &77.14±0.71 &84.37±0.99 &23.98±0.79 &55.91 \\
  \cline{2-10}
  % &SeGD-VPT~\cite{zhuo2024prompt}  &ViT/CLIP &- &Y &22.03±0.32 &37.18±0.50 &83.58±0.52 &90.45±0.52 &58.31 \\
  &StepSTP~\cite{xu2024step} &ViT/CLIP &- &Y &32.97±0.27 &70.01±0.21 &84.84±0.72 &\textbf{22.84±0.95} &52.68 \\
  &CoOp~\cite{zhou2022learning}  &ViT/CLIP &- &Y  &32.86±0.47 &72.08±0.66 &80.50±0.74 &21.65±0.32 &51.77 \\
  &Tip-Adapter~\cite{zhang2021tip}  &ViT/CLIP &- &Y  &32.68±0.37 &75.44±0.51 &77.15±0.66 &22.24±0.26 &51.87 \\
  &PromptSRC~\cite{Khattak_2023_ICCV}  &ViT/CLIP &- &Y  &31.86±0.57 &73.44±0.71 &76.15±0.89 &21.16±0.36 &50.65 \\
  &PDA~\cite{bai2024prompt} &ViT/CLIP &- &Y &31.45±0.44 &69.68±0.71 &79.20±0.83 &20.66±0.28 &50.25\\
  &AMU-Tuning~\cite{tang2024amu}  &ViT/CLIP &- &Y  &32.29±0.67 &72.24±0.71 &80.20±0.86 &21.56±0.36 &51.57 \\
  &LP++~\cite{huang2024lp++}  &ViT/CLIP &- &Y  &33.63±0.41 &73.05±0.55 &81.84±0.66 &21.72±0.42 &52.56 \\
  &LDC~\cite{li2025logits}  &ViT/CLIP &- &Y  &33.72±0.46 &74.39±0.52 &84.07±0.61 &22.32±0.36 &53.62 \\
  % &\textbf{CoOp + OURS}  &ViT/CLIP &- &Y &33.44±0.50 &72.51±0.65 &81.77±0.79 &21.71±0.31 &52.36 \\
  &Maple~\cite{khattak2023maple}  &ViT/CLIP &- &Y &33.38±0.49 &76.05±0.63 &81.78±0.72 &21.09±0.31 &53.07 \\
  \rowcolor{cyan!10} \cellcolor{white}
  &\textbf{Maple + OURS}  &ViT/CLIP &- &Y &35.11±0.51 &76.92±0.65 &82.51±0.69 &21.64±0.34 &54.05 \\
  &CLIP-LoRA-Vision~\cite{zanella2024low}  &ViT/CLIP &- &Y &36.40±0.42 &81.72±0.52 &84.62±0.62 &21.86±0.32 &56.07 \\
  \rowcolor{cyan!10} \cellcolor{white}
  &\textbf{CLIP-LoRA-Vision + OURS}  &ViT/CLIP &- &Y &\textbf{38.12±0.48} &\textbf{85.02±0.46} &\textbf{87.20±0.51} &22.68±0.41 &\textbf{58.26} \\
  \cline{2-10}
  &SigLIP2-LoRA~\cite{tschannen2025siglip} &ViT/SigLip2 &- &Y  &33.47 &74.16 & 87.50 &21.44 & 54.14 \\
  \rowcolor{cyan!10} \cellcolor{white}
  &\textbf{SigLIP2-LoRA + OURS} &ViT/SigLip2 &- &Y  &\textbf{36.88} &\textbf{78.04} & \textbf{90.85} &\textbf{22.27} & \textbf{57.01} \\
  \cline{2-10}
  &PE-Core-LoRA~\cite{bolya2025perception} &ViT/PE-Core &- &Y  &40.89 &84.49 & 91.75 &22.02 & 59.78 \\
  \rowcolor{cyan!10} \cellcolor{white}
  &\textbf{PE-Core-LoRA + OURS} &ViT/PE-Core &- &Y  &\textbf{45.01} &\textbf{86.83} & \textbf{93.03} &\textbf{23.66} & \textbf{62.14} \\
  \midrule
  \midrule
  % &GNN [12] &RN10 &Y &- &25.27±0.46 &43.94±0.67 &83.64±0.77 &87.96±0.67 &60.20 \\
  % &FWT [40] &RN10 &Y &- &25.18±0.45 &43.17±0.70 &83.01±0.79 &87.11±0.67 &59.62 \\
  % &LRP [38] &RN10 &Y &- &24.53±0.30 &44.14±0.40 &77.14±0.40 &86.15±0.40 &57.99 \\
  &ATA~\cite{wang2021cross} &RN10 &Y &- &44.91±0.40 &83.75±0.40 &90.59±0.30 &24.32±0.40 & 60.89 \\
  &AFA~\cite{hu2022adversarial} &RN10 &Y &-  &46.01±0.40 &85.58±0.40 &88.06±0.30 &25.02±0.20 &61.17 \\
  &wave-SAN~\cite{fu2022wave} &RN10 &Y &-  &44.93±0.67 &85.22±0.71 &89.70±0.64 &25.63±0.49 &61.37 \\
  &StyleAdv~\cite{fu2023styleadv} &RN10 &Y &- &45.77±0.51 &86.58±0.54 &93.65±0.39 &26.07±0.37 &63.02 \\
  \cline{2-10}
  % &Fine-tune [14] &RN10 &Y &Y &25.97±0.41 &48.11±0.64 &79.08±0.61 &89.25±0.51 &60.60 \\
  % &NSAE [25] &RN10 &Y &Y &27.10±0.44 &54.05±0.63 &83.96±0.57 &93.14±0.47 &64.56 \\
  % &BSR [26] &RN10 &Y &Y &26.84±0.44 &54.42±0.66 &80.89±0.61 &92.17±0.45 &63.58 \\
  &ATA-FT~\cite{wang2021cross} &RN10 &Y &Y &49.79±0.40 &89.64±0.30 &95.44±0.20 &25.08±0.20 &64.99 \\
  &DARA~\cite{zhao2023dual} &RN10 &Y &Y &56.28±0.66 &85.84±0.54 &95.32±0.34 &27.54±0.42 &66.25 \\
  &StyleAdv-FT~\cite{fu2023styleadv} &RN10 &Y &Y &53.05±0.54 &91.64±0.43 &96.51±0.28 &26.24±0.35 &66.86 \\
  \cline{2-10}
  \multirow{14}{*}{\rotatebox{90}{5-way 5-shot}} 
  &PMF~\cite{hu2022pushing} &ViT/DINO &Y &Y &50.12 &85.98 &92.96 &27.27 &64.08 \\
  &StyleAdv-FT~\cite{fu2023styleadv} &ViT/DINO &Y &Y &51.23±0.51 &90.12±0.33 &95.99±0.27 &26.97±0.33 &66.08 \\
  &FLoR~\cite{zou2024flatten} &ViT/DINO &Y &Y &53.06 &90.75 &96.47 & 27.02 & 66.83 \\
  &CD-CLS~\cite{zou2025closer} &ViT/DINO &Y &Y  &54.69 & 91.53 & 96.27 &27.66 & 67.54 \\
  &AttnTemp~\cite{zouattention} &ViT/DINO &Y &Y  &54.91 &90.82 & 96.66 &28.03 & 67.61 \\  
  \cline{2-10}
  &FN+VDB~\cite{yazdanpanah2022visual} &RN18 &- &Y &47.48±0.59 &87.31±0.50 &94.63±0.37 &25.55±0.43 &64.74 \\
  &IM-DCL~\cite{xu2024enhancing} &RN10 &- &Y &52.74±0.69 &89.47±0.42 &95.73±0.38 &28.93±0.41 &66.72 \\
  \cline{2-10}
  % &SeGD-VPT~\cite{zhuo2024prompt} &ViT/CLIP &- &Y &23.20±0.30 &53.10±0.51 &93.81±0.24 &96.93±0.25 &66.76 \\
  &StepSTP~\cite{xu2024step} &ViT/CLIP &- &Y &52.12±0.36 &89.40±1.05 &96.01±0.88 &26.36±0.97 &65.97 \\
  &CoOp~\cite{zhou2022learning}  &ViT/CLIP &- &Y &45.78±0.75 &85.88±0.49 &93.31±0.57 &23.35±0.50 &62.08 \\
  &Tip-Adapter~\cite{zhang2021tip}  &ViT/CLIP &- &Y  &46.96±0.59 &87.24±0.33 &94.19±0.39 &24.07±0.44 &63.12 \\
  &PromptSRC~\cite{Khattak_2023_ICCV}  &ViT/CLIP &- &Y  &46.09±0.48 &86.54±0.49 &89.97±0.41 &23.51±0.47 &61.52 \\
  & PDA~\cite{bai2024prompt} &ViT/CLIP &- &Y &45.19±0.62 &86.21±0.44 &92.67±0.39 &21.87±0.33 &61.48 \\
  &AMU-Tuning~\cite{tang2024amu}  &ViT/CLIP &- &Y  &44.60±0.62 &88.47±0.39 &94.26±0.52 &23.34±0.41 &62.66 \\
  &LP++~\cite{huang2024lp++}  &ViT/CLIP &- &Y  &48.49±0.44 &87.48±0.42 &94.47±0.38 &23.89±0.29 &63.58 \\
  &LDC~\cite{li2025logits}  &ViT/CLIP &- &Y  &49.70±0.33 &90.82±0.22 &96.71±0.34 &25.89±0.21 &65.78 \\
  % &\textbf{CoOp + OURS}  &ViT/CLIP &- &Y &45.95±0.74 &86.88±0.52 &93.92±0.55 &23.55±0.43 &62.58 \\
  &Maple~\cite{khattak2023maple}  &ViT/CLIP &- &Y &48.35±0.75 &89.04±0.52 &93.50±0.54 &22.96±0.50 &63.46 \\
  \rowcolor{cyan!10} \cellcolor{white}
  &\textbf{Maple + OURS}  &ViT/CLIP &- &Y &50.01±0.78 &91.00±0.50 &94.48±0.50 &23.45±0.49 &64.74 \\
  &CLIP-LoRA-Vision~\cite{zanella2024low}  &ViT/CLIP &- &Y &52.22±0.71 &93.31±0.47 &95.88±0.42 &24.61±0.47 &66.50 \\
  \rowcolor{cyan!10} \cellcolor{white}
  &\textbf{CLIP-LoRA-Vision + OURS}  &ViT/CLIP &- &Y &\textbf{56.14±0.46} &\textbf{94.14±0.34} &\textbf{96.64±0.39} &\textbf{26.61±0.43} &\textbf{68.38}\\
  \cline{2-10}
  &SigLIP2-LoRA~\cite{tschannen2025siglip} &ViT/SigLip2 &- &Y  &51.79 &91.39 & 96.43
  &24.24 & 65.96 \\
  \rowcolor{cyan!10} \cellcolor{white}
  &\textbf{SigLIP2-LoRA + OURS} &ViT/SigLip2 &- &Y  &\textbf{55.12} &\textbf{92.10} & \textbf{97.37} &\textbf{26.44} & \textbf{67.43} \\
  \cline{2-10}
  &PE-Core-LoRA~\cite{bolya2025perception} &ViT/PE-Core &- &Y  &58.81 &94.07 & 97.25 &24.44 & 68.64 \\
  \rowcolor{cyan!10} \cellcolor{white}
  &\textbf{PE-Core-LoRA + OURS} &ViT/PE-Core &- &Y  &\textbf{61.41} &\textbf{94.83} & \textbf{98.13} &\textbf{26.77} & \textbf{70.29} \\
\bottomrule
\end{tabular}
\end{adjustbox}
%%\vspace{-0.3cm}
\end{table*}

\begin{table*}[!htbp]
\setlength{\abovecaptionskip}{0.cm}
\setlength{\belowcaptionskip}{0.cm}
\caption{
Detailed results for 11 datasets using ViT-B/16 as the visual backbone are presented. Top-1 accuracy, averaged over 3 random seeds, is reported. The highest value is highlighted in bold, and the second-highest performance is underlined.
}
%\vspace{0.05cm}
\label{tab:fewshot_shull}    
\centering
\begin{adjustbox}{max width=0.89\linewidth}
\begin{tabular}{*{15}{c}}  
\toprule 
 \multirow{1}*{Shots} & \multirow{1}*{Method}  &  \multirow{1}*{ImageNet}  & \multirow{1}*{SUN}
 & \multirow{1}*{Aircraft}  & \multirow{1}*{EuroSAT}  & \multirow{1}*{Cars}  
  & \multirow{1}*{Food}   & \multirow{1}*{Pets}  & \multirow{1}*{Flowers}  
  & \multirow{1}*{Caltech}  & \multirow{1}*{DTD} & \multirow{1}*{UCF} & \multirow{1}*{Average}\\
\midrule
& CoOp(4) & 68.0 & 67.3 & 26.2 & 50.9  & 67.1 & 82.6  & 90.3 & 72.7 & 93.2 & 50.1 & 70.7 & 67.2\\
& CoOp(16) & 65.7 & 67.0 & 20.8 & 56.4  & 67.5 & 84.3  & 90.2 & 78.3 & 92.5 & 50.1 & 71.2 & 67.6\\
& CoCoOp & 69.4 & 68.7 & 28.1 & 55.4  & 67.6 & 84.9  & 91.9 & 73.4 & 94.1 & 52.6 & 70.4 & 68.8\\
& TIP-Adapter-F & 69.4 & 67.2 & 28.8 & 67.8 & 67.1 & 85.8 & 90.6 & 83.8 & 94.0 & 51.6 & 73.4 & 70.9\\
& CLIP-Adapter & 67.9 & 65.4 & 25.2 & 49.3 & 65.7 & 86.1 & 89.0 & 71.3 & 92.0 & 44.2 & 66.9 & 65.7\\
& PLOT++ & 66.5 & 66.8 & 28.6 & 65.4 & 68.8 & 86.2 & 91.9 & 80.5 & \textbf{94.3} & 54.6 & 74.3 & 70.7\\
\textbf{1} & KgCoOp & 68.9 & 68.4 & 26.8 & 61.9 & 66.7 & 86.4 & 92.1 & 74.7 & 94.2 & 52.7 & 72.8 & 69.6\\
& TaskRes & 69.6 & 68.1 & 31.3 & 65.4 & 68.8 & 84.6 & 90.2 & 81.7 & 93.6 & 53.8 & 71.7 & 70.8\\
& MaPLe & 69.7 & 69.3 & 28.1 & 29.1 & 67.6 & 85.4 & 91.4 & 74.9 & 93.6 & 50.0 & 71.1 & 66.4\\
& ProGrad & 67.0 & 67.0 & 28.8 & 57.0 & 68.2 & 84.9 & 91.4 & 80.9 & 93.5 & 52.8 & 73.3 & 69.5\\
& LDC & 69.5	& 68.0	& 27.5	& 78.4	& 68.2	& 85.8	& 91.3	& 83.6	& 93.8	& \textbf{58.2}	& 73.2	& 72.5 \\
& CLIP-LoRA & 70.4 & 70.4 & 30.2 & 72.3  & 70.1 & 84.3  & 92.3 & 83.2 & 93.7 & 54.3 & 76.3 & 72.5\\
\rowcolor{cyan!10}
& \textbf{CLIP-LoRA + Ours} & \textbf{70.4} & \textbf{70.7} & \textbf{30.9} & \textbf{78.4}  & \textbf{70.7} & \textbf{86.5}  & \textbf{92.4} & \textbf{83.9} & 93.9 & \underline{54.8} &\textbf{76.4} & \textbf{73.5}\\
\bottomrule

&CoOp(4) &68.7&68.0&28.1&66.2&70.5&82.6&89.9&80.9&93.0&53.7&73.5&70.5\\
&CoOp(16) &67.0&67.0&25.9&65.1&70.4&84.4&89.9&88.0&93.1&54.1&74.1&70.8\\
&CoCoOp &70.1&69.4&29.3&61.8&68.4&85.9&91.9&77.8&94.4&52.3&73.4&70.4\\
&TIP-Adapter-F &70.0&68.6&32.8&73.2&70.8&86.0&91.6&90.1&93.9&57.8&76.2&73.7\\
&CLIP-Adapter &68.2&67.2&27.0&51.2&66.6&86.2&89.7&71.7&93.4&45.4&68.4&66.8\\
&PLOT++ &68.3&68.1&31.1&76.8&73.2&86.3&92.3&89.8&94.7&56.7&76.8&74.0\\
\textbf{2} &KgCoOp &69.6&69.6&28.0&69.2&68.2&86.6&92.3&79.8&94.5&55.3&74.6&71.6\\
&TaskRes &70.2&70.5&32.7&70.2&72.1&85.6&90.7&84.4&94.3&55.6&75.2&72.9\\
&MaPLe &70.0&70.7&29.5&59.4&68.5&86.5&91.8&79.8&\textbf{94.9}&50.6&74.0&70.5\\
&ProGrad &69.1&69.0&31.1&66.3&72.4&84.8&91.5&87.5&93.6&56.0&75.6&72.4\\
&LDC &69.8	&69.6	&30.0 	&81.7	&70.75	&86.1	&91.2	&88.7	&94.3	&\textbf{62.2}	&75.9	&74.5 \\
&CLIP-LoRA &70.8&71.3&33.2&82.7&73.2&83.2&91.3&89.8&94.6&59.9&\textbf{80.0}&75.5\\
\rowcolor{cyan!10}
& \textbf{CLIP-LoRA + Ours} & \textbf{70.8} & \textbf{72.3} & \textbf{33.4}	& \textbf{83.1}	& \textbf{73.5}	& \textbf{86.6}	& \textbf{92.5}	& \textbf{90.4}	& \underline{94.7}	& \underline{60.8} & \underline{79.5}	& \textbf{76.2} \\
% & \textbf{OURS-new} & \textbf{70.5} & 70.3 & \textbf{32.2} & \textbf{78.4}  & \textbf{70.6} & 85.3  & \textbf{92.9} & 83.3 & \textbf{94.8} & \textbf{54.5} & 76.0 & \textbf{73.5}\\

\bottomrule
&CoOp (4) & 69.7 & 70.6 & 29.7 & 65.8 & 73.4 & 83.5 & 92.3 & 86.6 & 94.5 & 58.5 & 78.1 & 73.0\\
&CoOp (16)  & 68.8 & 69.7 & 30.9 & 69.7 & 74.4 & 84.5 & 92.5 & 92.2 & 94.5 & 59.5 & 77.6 & 74.0\\
&CoCoOp  & 70.6 & 70.4 & 30.6 & 61.7 & 69.5 & 86.3 & 92.7 & 81.5 & 94.8 & 55.7 & 75.3 & 71.7\\
&TIP-Adapter-F  & 70.7 & 70.8 & 35.7 & 76.8 & 74.1 & 86.5 & 91.9 & 92.1 & 94.8 & 59.8 & 78.1 & 75.6\\
&CLIP-Adapter  & 68.6 & 68.0 & 27.9 & 51.2 & 67.5 & 86.5 & 90.8 & 73.1 & 94.0 & 46.1 & 70.6 & 67.7\\
&PLOT++  & 70.4 & 71.7 & 35.3 & 83.2 & 76.3 & 86.5 & 92.6 & 92.9 & 95.1 & 62.4 & 79.8 & 76.9\\
\textbf{4} &KgCoOp  & 69.9 & 71.5 & 32.2 & 71.8 & 69.5 & 86.9 & 92.6 & 87.0 & 95.0 & 58.7 & 77.6 & 73.9\\
&TaskRes  & 71.0 & 72.7 & 33.4 & 74.2 & 76.0 & 86.0 & 91.9 & 85.0 & 95.0 & 60.1 & 76.2 & 74.7\\
&MaPLe  & 70.6 & 71.4 & 30.1 & 69.9 & 70.1 & 86.7 & 93.3 & 84.9 & 95.0 & 59.0 & 77.1 & 73.5\\
&ProGrad & 70.2 & 71.7 & 34.1 & 69.6 & 75.0 & 85.4 & 92.1 & 91.1 & 94.4 & 59.7 & 77.9 & 74.7\\
&LDC & 71.0& 72.9& 31.9	& 86.3& 75.1& 86.7& 91.8& 93.9& 95.2& \textbf{66.4}& 79.7& 77.3\\
&CLIP-LoRA & 71.4 & 72.8 & 37.9 & 84.9  & 77.4 & 82.7  & 91.0 & 93.7 & 95.2 & 63.8 & 81.1 & 77.4\\
% &\textbf{OURS} & \textbf{71.6} & 73.6 & 38.7 & 87.2  & 77.5 & 84.2  & 92.0 & 93.8 & 95.6 & 64.5 & 81.7 & 78.2\\
\rowcolor{cyan!10}
&\textbf{CLIP-LoRA + Ours} & \textbf{71.4} & \textbf{73.7} & \textbf{39.2} & \textbf{86.4}  & \textbf{78.0} & \textbf{87.0}  & \textbf{93.4 }& \textbf{94.3} & \textbf{95.8} & \underline{65.2} & \textbf{81.3} & \textbf{78.7}\\
\bottomrule
&CoOp(4) & 70.8 & 72.4 & 37.0 & 74.7 & 76.8 & 83.3 & 92.1 & 95.0 & 94.7 & 63.7 & 79.8 & 76.4\\
 &CoOp(16) & 70.6 & 71.9 & 38.5 & 77.1 & 79.0 & 82.7 & 91.3 & 94.9 & 94.5 & 64.8 & 80.0 & 76.8\\
 &CoCoOp & 70.8 & 71.5 & 32.4 & 69.1 & 70.4 & 87.0 & 93.3 & 86.3 & 94.9 & 60.1 & 75.9 & 73.8\\
 &TIP-Adapter-F & 71.7 & 73.5 & 39.5 & 81.3 & 78.3 & 86.9 & 91.8 & 94.3 & 95.2 & 66.7 & 82.0 & 78.3\\
 &CLIP-Adapter & 69.1 & 71.7 & 30.5 & 61.6 & 70.7 & 86.9 & 91.9 & 83.3 & 94.5 & 50.5 & 76.2 & 71.5\\
 &PLOT++ & 71.3 & 73.9 & 41.4 & 88.4 & 81.3 & 86.6 & 93.0 & 95.4 & 95.5 & 66.5 & 82.8 & 79.6\\
 \textbf{8} &KgCoOp & 70.2 & 72.6 & 34.8 & 73.9 & 72.8 & 87.0 & 93.0 & 91.5 & 95.1 & 65.6 & 80.0 & 76.0\\
 &TaskRes & 72.3 & 74.6 & 40.3 & 77.5 & 79.6 & 86.4 & 92.0 & 96.0 & 95.3 & 66.7 & 81.6 & 78.4\\
 &MaPLe & 71.3 & 73.2 & 33.8 & 82.8 & 71.3 & 87.2 & 93.1 & 90.5 & 95.1 & 63.0 & 79.5 & 76.4\\
 &ProGrad & 71.3 & 73.0 & 37.7 & 77.8 & 78.7 & 86.1 & 92.2 & 95.0 & 94.8 & 63.9 & 80.5 & 77.4\\
 &LDC & 72.4	& 75.5 & 38.0 & 90.8 & 79.7 & 86.9 & 92.5 & 96.0 & 95.9 & \textbf{71.5} & 80.9 & 80.0\\
 &CLIP-LoRA & 72.3 & 74.7 & 45.7 & 89.7 & 82.1 & 83.1 & 91.7 & 96.3 & 95.6 & 67.5 & 84.1 & 80.3\\
 % &\textbf{OURS } & \textbf{72.3} & 75.1 & 47.4 & \textbf{91.2} & \textbf{83.0} & 84.0 & 92.4 & 96.4 & \textbf{95.9} & 67.2 & \textbf{85.2} & 80.9\\
 \rowcolor{cyan!10}
&\textbf{CLIP-LoRA + Ours} & \textbf{72.4} & \textbf{75.5} & \textbf{48.6} & \textbf{90.9} & \textbf{82.5} & \textbf{87.3} & \textbf{94.0} & \textbf{96.9} & \textbf{95.9} & \underline{69.5} & \textbf{84.1} & \textbf{81.6}\\
\bottomrule
&CoOp(4)&71.5&74.6&40.1&83.5&79.1&85.1&92.4&96.4&95.5&69.2&81.9&79.0\\
&CoOp(16)&71.9&74.9&43.2&85.0&82.9&84.2&92.0&96.8&95.8&69.7&83.1&80.0\\
&CoCoOp&71.1&72.6&33.3&73.6&72.3&87.4&93.4&89.1&95.1&63.7&77.2&75.4\\
&TIP-Adapter-F&73.4&76.0&44.6&85.9&82.3&86.8&92.6&96.2&95.7&70.8&83.9&80.7\\
&CLIP-Adapter&69.8&74.2&34.2&71.4&74.0&87.1&92.3&92.9&94.9&59.4&80.2&75.5\\
&PLOT++&72.6&76.0&46.7&92.0&84.6&87.1&93.6&97.6&96.0&71.4&85.3&82.1\\
\textbf{16} &KgCoOp&70.4&73.3&36.5&76.2&74.8&87.2&93.2&93.4&95.2&68.7&81.7&77.3\\
&TaskRes&73.0&76.1&44.9&82.7&83.5&86.9&92.4&97.5&95.8&71.5&84.0&80.8\\
&MaPLe&71.9&74.5&36.8&87.5&74.3&87.4&93.2&94.2&95.4&68.4&81.4&78.6\\
&ProGrad&72.1&75.1&43.0&83.6&82.9&85.8&92.8&96.6&95.9&68.8&82.7&79.9\\
&LDC&\textbf{73.8}&76.9&47.8&92.1&84.1&87.3&93.3&97.8&96.4&\textbf{77.0}&84.4&82.8\\
&CLIP-LoRA&73.6&76.1&54.7&92.1&86.3&84.2&92.4&98.0&96.4&72.0&\textbf{86.7}&83.0\\
\rowcolor{cyan!10}
& \textbf{CLIP-LoRA + Ours} & 73.4 & \textbf{76.9} & \textbf{57.3}	& \textbf{92.5}	& \textbf{86.6}	& \textbf{87.7} & \textbf{94.5}	& \textbf{98.4}	& \textbf{96.5}	& \underline{73.2} & \underline{86.1}	& \textbf{84.0} \\

\bottomrule
\end{tabular}
\end{adjustbox}
%\vspace{-0.3cm}
\end{table*}

\section{Datasets}
 Following previous works~\cite{yazdanpanah2022visual,zhuo2024prompt,xu2024step,radford2021learning}, we do not utilize source domain datasets and finetune our model directly on the target domain. For the target domain we utilize CropDisease~\cite{mohanty2016using}, EuroSAT~\cite{helber2019eurosat}, ISIC~\cite{codella2019skin}, and ChestX~\cite{wang2017chestx}, which are cross-domain datasets from the domain of agriculture, remote sensing, and medical data with significant domain gaps.

\section{Implementation Details}
In the experiments, we adopt the ViT-Base/16 network as the primary feature extraction network, with parameters pre-trained by CLIP~\cite{radford2021learning}, SigLIP2~\cite{tschannen2025siglip} and PE-Core~\cite{bolya2025perception}. For the parameter $\lambda$, we consider two cases: when $\mathcal{L}_{\text{ad}}$ is used for the visual branch (e.g., Maple, CLIP-LoRA), $\lambda$ is set to 0.1; when used for the text branch (e.g., CoOp), $\lambda$ is set to 0.001. The parameter $\beta$ is set to 0.5 in all cases. For each episode, following~\cite{xu2024step}, we first perform data augmentation on the support samples. Subsequently, for each method, we train for 250 epochs, and the first 150 epochs are used as the initial epochs, employing $\mathcal{L}_{\text{ad}}$ and $\mathcal{L}_{\text{ra}}$. We evaluate every network using 15 query samples per class, randomly selecting 800 tasks, and report the average results (\%) with the 95\% confidence interval. We use an NVIDIA GeForce RTX 3090 GPU for training and testing.

\section{Broader Impact}
In this paper, we observe the misalignment of modalities in CLIP under cross-domain scenarios and reveal that the cross-entropy-based fine-tuning process inherently incorporates strong visual learning. This learning direction acts as a shortcut, allowing the model to reduce the loss without considering cross-modal relationships, leading to suppression of cross-modal relationship learning. Based on these findings, We then propose two methods to suppress visual learning and enhance cross-modal alignment. Our research is crucial for future studies on fine-tuning VLM models in cross-domain scenarios. It highlights the impact of visual learning on fine-tuning, a factor that has been overlooked in previous work. While our method has been evaluated across four distinct target domains, offering a promising initial assessment of its cross-domain applicability, the diversity of these domains may not fully capture all potential real-world scenarios. Future work will focus on expanding our evaluations to include a broader range of target domains to better understand the method's performance in diverse real-world contexts.

\section{Pseudocode}
Below is the pseudocode for fine-tuning the model in a few-shot learning setting. The parts highlighted in red correspond to the implementation of our method. Our method can be implemented with just a few lines of code.
\begin{algorithm}[h]
\caption{Suppressing Visual Learning and Relationship Alignment}
\label{alg:template_correction}
\begin{algorithmic}[1]
\REQUIRE Pretrained VLMs with visual encoder $\theta_v$ and text encoder $\theta_t$, tokenized prompts for all classes  $\{r_k\}$, few-shot labeled samples $\{(\mathbf{x}_i, y_i)\}$, and hyperparameters $\lambda$ and $\beta$.
% \STATE \textbf{Stage1: Auto-Construct Empty Prompts:}
% \STATE Use a language model to generate a set of words and phrases which mean “empty”: $T_{\text{null}}$
%\STATE \textbf{Stage1: Pretraining Initialization:}
\FOR{each labeled sample $(\mathbf{x}_i, y_i)$}
    \STATE Obtain visual feature $f_i = \theta_v(\mathbf{x}_i)$
\ENDFOR
\FOR{each prompt $r_j \in \{r_k\}$}
    \STATE Obtain text feature \( t_j = \theta_t(r_j) \)
\ENDFOR
\STATE Compute logits $\mathcal{S} = \mathcal{F}\mathcal{T}^T$.
\STATE Using $\mathcal{S}$ to compute cross-entropy loss $\mathcal{L}_{\text {vlm}}$.
\textcolor{red}{
\STATE Compute visual feature similarity matrix $A^v = \mathcal{F}\mathcal{F}^T$.
\STATE Compute text feature similarity matrix $A^t = \mathcal{T}\mathcal{T}^T$.
%\STATE \textbf{Suppressing Visual Learnin}
\STATE Randomly generate index $I_{rand}$.
\STATE Compute class-shffle logits as $\mathcal{S}_{rand} = A^v[:,I_{rand}]$.
\STATE Using $\mathcal{S}_{rand}$ as new logits to compute cross-entropy loss as the anti-visual loss $\mathcal{L}_{\text {ad }}$.
%\STATE \textbf{Relationship Alignment}
\STATE Fusing $A^v$ and $A^t$ to get $A^{fuse}$.
\STATE Compute relationship alignment loss $\mathcal{L}_{\mathrm{ra}}$.
\STATE Total loss $\mathcal{L}= \mathcal{L}_{\text {vlm}} + \beta \mathcal{L}_{\text {ra}}+\lambda \mathcal{L}_{\text {ad }}$.
}
\end{algorithmic}
\end{algorithm}

%% file: main.bib
@String(CVPR= {IEEE Conf. Comput. Vis. Pattern Recog.})

@String(ICCV= {Int. Conf. Comput. Vis.})

@String(ECCV= {Eur. Conf. Comput. Vis.})

@String(AAAI = {AAAI})

@String(CVPR  = {CVPR})

@String(ICCV  = {ICCV})

@String(ECCV  = {ECCV})

@article{fu2022wave,
  title={Wave-san: Wavelet based style augmentation network for cross-domain few-shot learning},
  author={Fu, Yuqian and Xie, Yu and Fu, Yanwei and Chen, Jingjing and Jiang, Yu-Gang},
  journal={arXiv preprint arXiv:2203.07656},
  year={2022}
}

@inproceedings{guo2020broader,
  title={A broader study of cross-domain few-shot learning},
  author={Guo, Yunhui and Codella, Noel C and Karlinsky, Leonid and Codella, James V and Smith, John R and Saenko, Kate and Rosing, Tajana and Feris, Rogerio},
  booktitle={Computer vision--ECCV 2020: 16th European conference, glasgow, UK, August 23--28, 2020, proceedings, part XXVII 16},
  pages={124--141},
  year={2020},
  organization={Springer}
}

@inproceedings{hu2022adversarial,
  title={Adversarial feature augmentation for cross-domain few-shot classification},
  author={Hu, Yanxu and Ma, Andy J},
  booktitle={European conference on computer vision},
  pages={20--37},
  year={2022},
  organization={Springer}
}

@article{wang2021cross,
  title={Cross-domain few-shot classification via adversarial task augmentation},
  author={Wang, Haoqing and Deng, Zhi-Hong},
  journal={arXiv preprint arXiv:2104.14385},
  year={2021}
}

@inproceedings{liang2021boosting,
  title={Boosting the generalization capability in cross-domain few-shot learning via noise-enhanced supervised autoencoder},
  author={Liang, Hanwen and Zhang, Qiong and Dai, Peng and Lu, Juwei},
  booktitle={Proceedings of the IEEE/CVF international conference on computer vision},
  pages={9424--9434},
  year={2021}
}

@inproceedings{zhou2023revisiting,
  title={Revisiting prototypical network for cross domain few-shot learning},
  author={Zhou, Fei and Wang, Peng and Zhang, Lei and Wei, Wei and Zhang, Yanning},
  booktitle={Proceedings of the IEEE/CVF conference on computer vision and pattern recognition},
  pages={20061--20070},
  year={2023}
}

@inproceedings{zou2024flatten,
  title={Flatten long-range loss landscapes for cross-domain few-shot learning},
  author={Zou, Yixiong and Liu, Yicong and Hu, Yiman and Li, Yuhua and Li, Ruixuan},
  booktitle={Proceedings of the IEEE/CVF Conference on Computer Vision and Pattern Recognition},
  pages={23575--23584},
  year={2024}
}

@article{zou2025closer,
  title={A Closer Look at the CLS Token for Cross-Domain Few-Shot Learning},
  author={Zou, Yixiong and Yi, Shuai and Li, Yuhua and Li, Ruixuan},
  journal={Advances in Neural Information Processing Systems},
  volume={37},
  pages={85523--85545},
  year={2025}
}

@inproceedings{zouattention,
  title={Attention Temperature Matters in ViT-Based Cross-Domain Few-Shot Learning},
  author={Zou, Yixiong and Ma, Ran and Li, Yuhua and Li, Ruixuan},
  booktitle={The Thirty-eighth Annual Conference on Neural Information Processing Systems}
}

@inproceedings{yazdanpanah2022visual,
  title={Visual domain bridge: A source-free domain adaptation for cross-domain few-shot learning},
  author={Yazdanpanah, Moslem and Moradi, Parham},
  booktitle={Proceedings of the IEEE/CVF conference on computer vision and pattern recognition},
  pages={2868--2877},
  year={2022}
}

@article{zhuo2024prompt,
  title={Prompt as Free Lunch: Enhancing Diversity in Source-Free Cross-domain Few-shot Learning through Semantic-Guided Prompting},
  author={Zhuo, Linhai and Wang, Zheng and Fu, Yuqian and Qian, Tianwen},
  journal={arXiv preprint arXiv:2412.00767},
  year={2024}
}

@article{xu2024step,
  title={Step-wise Distribution Alignment Guided Style Prompt Tuning for Source-free Cross-domain Few-shot Learning},
  author={Xu, Huali and Liu, Yongxiang and Liu, Li and Zhi, Shuaifeng and Sun, Shuzhou and Liu, Tianpeng and Cheng, MingMing},
  journal={arXiv preprint arXiv:2411.10070},
  year={2024}
}

@inproceedings{radford2021learning,
  title={Learning transferable visual models from natural language supervision},
  author={Radford, Alec and Kim, Jong Wook and Hallacy, Chris and Ramesh, Aditya and Goh, Gabriel and Agarwal, Sandhini and Sastry, Girish and Askell, Amanda and Mishkin, Pamela and Clark, Jack and others},
  booktitle={International conference on machine learning},
  pages={8748--8763},
  year={2021},
  organization={PmLR}
}

@article{zhou2022learning,
  title={Learning to prompt for vision-language models},
  author={Zhou, Kaiyang and Yang, Jingkang and Loy, Chen Change and Liu, Ziwei},
  journal={International Journal of Computer Vision},
  volume={130},
  number={9},
  pages={2337--2348},
  year={2022},
  publisher={Springer}
}

@inproceedings{zhou2022conditional,
  title={Conditional prompt learning for vision-language models},
  author={Zhou, Kaiyang and Yang, Jingkang and Loy, Chen Change and Liu, Ziwei},
  booktitle={Proceedings of the IEEE/CVF conference on computer vision and pattern recognition},
  pages={16816--16825},
  year={2022}
}

@inproceedings{khattak2023maple,
  title={Maple: Multi-modal prompt learning},
  author={Khattak, Muhammad Uzair and Rasheed, Hanoona and Maaz, Muhammad and Khan, Salman and Khan, Fahad Shahbaz},
  booktitle={Proceedings of the IEEE/CVF Conference on Computer Vision and Pattern Recognition},
  pages={19113--19122},
  year={2023}
}

@article{chen2022plot,
  title={Plot: Prompt learning with optimal transport for vision-language models},
  author={Chen, Guangyi and Yao, Weiran and Song, Xiangchen and Li, Xinyue and Rao, Yongming and Zhang, Kun},
  journal={arXiv preprint arXiv:2210.01253},
  year={2022}
}

@inproceedings{yao2023visual,
  title={Visual-language prompt tuning with knowledge-guided context optimization},
  author={Yao, Hantao and Zhang, Rui and Xu, Changsheng},
  booktitle={Proceedings of the IEEE/CVF conference on computer vision and pattern recognition},
  pages={6757--6767},
  year={2023}
}

@inproceedings{zhu2023prompt,
  title={Prompt-aligned gradient for prompt tuning},
  author={Zhu, Beier and Niu, Yulei and Han, Yucheng and Wu, Yue and Zhang, Hanwang},
  booktitle={Proceedings of the IEEE/CVF International Conference on Computer Vision},
  pages={15659--15669},
  year={2023}
}

@inproceedings{khattak2023self,
  title={Self-regulating prompts: Foundational model adaptation without forgetting},
  author={Khattak, Muhammad Uzair and Wasim, Syed Talal and Naseer, Muzammal and Khan, Salman and Yang, Ming-Hsuan and Khan, Fahad Shahbaz},
  booktitle={Proceedings of the IEEE/CVF International Conference on Computer Vision},
  pages={15190--15200},
  year={2023}
}

@inproceedings{yu2023task,
  title={Task residual for tuning vision-language models},
  author={Yu, Tao and Lu, Zhihe and Jin, Xin and Chen, Zhibo and Wang, Xinchao},
  booktitle={Proceedings of the IEEE/CVF Conference on Computer Vision and Pattern Recognition},
  pages={10899--10909},
  year={2023}
}

@article{lu2023beyond,
  title={Beyond sole strength: Customized ensembles for generalized vision-language models},
  author={Lu, Zhihe and Bai, Jiawang and Li, Xin and Xiao, Zeyu and Wang, Xinchao},
  journal={arXiv preprint arXiv:2311.17091},
  year={2023}
}

@inproceedings{li2024promptkd,
  title={Promptkd: Unsupervised prompt distillation for vision-language models},
  author={Li, Zheng and Li, Xiang and Fu, Xinyi and Zhang, Xin and Wang, Weiqiang and Chen, Shuo and Yang, Jian},
  booktitle={Proceedings of the IEEE/CVF Conference on Computer Vision and Pattern Recognition},
  pages={26617--26626},
  year={2024}
}

@article{gao2024clip,
  title={Clip-adapter: Better vision-language models with feature adapters},
  author={Gao, Peng and Geng, Shijie and Zhang, Renrui and Ma, Teli and Fang, Rongyao and Zhang, Yongfeng and Li, Hongsheng and Qiao, Yu},
  journal={International Journal of Computer Vision},
  volume={132},
  number={2},
  pages={581--595},
  year={2024},
  publisher={Springer}
}

@inproceedings{zhang2022tip,
  title={Tip-adapter: Training-free adaption of clip for few-shot classification},
  author={Zhang, Renrui and Zhang, Wei and Fang, Rongyao and Gao, Peng and Li, Kunchang and Dai, Jifeng and Qiao, Yu and Li, Hongsheng},
  booktitle={European conference on computer vision},
  pages={493--510},
  year={2022},
  organization={Springer}
}

@article{hu2021lora,
  title={Lora: Low-rank adaptation of large language models},
  author={Hu, Edward J and Shen, Yelong and Wallis, Phillip and Allen-Zhu, Zeyuan and Li, Yuanzhi and Wang, Shean and Wang, Lu and Chen, Weizhu},
  journal={arXiv preprint arXiv:2106.09685},
  year={2021}
}

@inproceedings{zanella2024low,
  title={Low-Rank Few-Shot Adaptation of Vision-Language Models},
  author={Zanella, Maxime and Ben Ayed, Ismail},
  booktitle={Proceedings of the IEEE/CVF Conference on Computer Vision and Pattern Recognition},
  pages={1593--1603},
  year={2024}
}

@article{valipour2022dylora,
  title={Dylora: Parameter efficient tuning of pre-trained models using dynamic search-free low-rank adaptation},
  author={Valipour, Mojtaba and Rezagholizadeh, Mehdi and Kobyzev, Ivan and Ghodsi, Ali},
  journal={arXiv preprint arXiv:2210.07558},
  year={2022}
}

@article{zhang2023adalora,
  title={AdaLoRA: Adaptive budget allocation for parameter-efficient fine-tuning},
  author={Zhang, Qingru and Chen, Minshuo and Bukharin, Alexander and Karampatziakis, Nikos and He, Pengcheng and Cheng, Yu and Chen, Weizhu and Zhao, Tuo},
  journal={arXiv preprint arXiv:2303.10512},
  year={2023}
}

@article{chavan2023one,
  title={One-for-all: Generalized lora for parameter-efficient fine-tuning},
  author={Chavan, Arnav and Liu, Zhuang and Gupta, Deepak and Xing, Eric and Shen, Zhiqiang},
  journal={arXiv preprint arXiv:2306.07967},
  year={2023}
}

@article{kim2024hydra,
  title={Hydra: Multi-head low-rank adaptation for parameter efficient fine-tuning},
  author={Kim, Sanghyeon and Yang, Hyunmo and Kim, Yunghyun and Hong, Youngjoon and Park, Eunbyung},
  journal={Neural Networks},
  pages={106414},
  year={2024},
  publisher={Elsevier}
}

@article{zi2023delta,
  title={Delta-lora: Fine-tuning high-rank parameters with the delta of low-rank matrices},
  author={Zi, Bojia and Qi, Xianbiao and Wang, Lingzhi and Wang, Jianan and Wong, Kam-Fai and Zhang, Lei},
  journal={arXiv preprint arXiv:2309.02411},
  year={2023}
}

@article{dettmers2024qlora,
  title={Qlora: Efficient finetuning of quantized llms},
  author={Dettmers, Tim and Pagnoni, Artidoro and Holtzman, Ari and Zettlemoyer, Luke},
  journal={Advances in Neural Information Processing Systems},
  volume={36},
  year={2024}
}

@article{rajabzadeh2024qdylora,
  title={Qdylora: Quantized dynamic low-rank adaptation for efficient large language model tuning},
  author={Rajabzadeh, Hossein and Valipour, Mojtaba and Zhu, Tianshu and Tahaei, Marzieh and Kwon, Hyock Ju and Ghodsi, Ali and Chen, Boxing and Rezagholizadeh, Mehdi},
  journal={arXiv preprint arXiv:2402.10462},
  year={2024}
}

@article{chen2019closer,
  title={A closer look at few-shot classification},
  author={Chen, Wei-Yu and Liu, Yen-Cheng and Kira, Zsolt and Wang, Yu-Chiang Frank and Huang, Jia-Bin},
  journal={arXiv preprint arXiv:1904.04232},
  year={2019}
}

@article{mistretta2025cross,
  title={Cross the Gap: Exposing the Intra-modal Misalignment in CLIP via Modality Inversion},
  author={Mistretta, Marco and Baldrati, Alberto and Agnolucci, Lorenzo and Bertini, Marco and Bagdanov, Andrew D},
  journal={arXiv preprint arXiv:2502.04263},
  year={2025}
}

@article{liang2022mind,
  title={Mind the gap: Understanding the modality gap in multi-modal contrastive representation learning},
  author={Liang, Victor Weixin and Zhang, Yuhui and Kwon, Yongchan and Yeung, Serena and Zou, James Y},
  journal={Advances in Neural Information Processing Systems},
  volume={35},
  pages={17612--17625},
  year={2022}
}

@article{tian2025mind,
  title={Mind the Gap Between Prototypes and Images in Cross-domain Finetuning},
  author={Tian, Hongduan and Liu, Feng and Zhou, Zhanke and Liu, Tongliang and Zhang, Chengqi and Han, Bo},
  journal={Advances in Neural Information Processing Systems},
  volume={37},
  pages={11251--11289},
  year={2025}
}

@article{mohanty2016using, 
  title={Using deep learning for image-based plant disease detection},
  author={Mohanty, Sharada P and Hughes, David P and Salath{\'e}, Marcel},
  journal={Frontiers in plant science},
  volume={7},
  pages={215232},
  year={2016},
  publisher={frontiers}
}

@article{helber2019eurosat,
  title={Eurosat: A novel dataset and deep learning benchmark for land use and land cover classification},
  author={Helber, Patrick and Bischke, Benjamin and Dengel, Andreas and Borth, Damian},
  journal={IEEE Journal of Selected Topics in Applied Earth Observations and Remote Sensing},
  volume={12},
  number={7},
  pages={2217--2226},
  year={2019},
  publisher={IEEE}
}

@article{codella2019skin,
  title={Skin lesion analysis toward melanoma detection 2018: A challenge hosted by the international skin imaging collaboration (isic)},
  author={Codella, Noel and Rotemberg, Veronica and Tschandl, Philipp and Celebi, M Emre and Dusza, Stephen and Gutman, David and Helba, Brian and Kalloo, Aadi and Liopyris, Konstantinos and Marchetti, Michael and others},
  journal={arXiv preprint arXiv:1902.03368},
  year={2019}
}

@inproceedings{wang2017chestx,
  title={Chestx-ray8: Hospital-scale chest x-ray database and benchmarks on weakly-supervised classification and localization of common thorax diseases},
  author={Wang, Xiaosong and Peng, Yifan and Lu, Le and Lu, Zhiyong and Bagheri, Mohammadhadi and Summers, Ronald M},
  booktitle={Proceedings of the IEEE conference on computer vision and pattern recognition},
  pages={2097--2106},
  year={2017}
}

@inproceedings{fu2023styleadv,
  title={Styleadv: Meta style adversarial training for cross-domain few-shot learning},
  author={Fu, Yuqian and Xie, Yu and Fu, Yanwei and Jiang, Yu-Gang},
  booktitle={Proceedings of the IEEE/CVF conference on computer vision and pattern recognition},
  pages={24575--24584},
  year={2023}
}

@article{zhao2023dual,
  title={Dual adaptive representation alignment for cross-domain few-shot learning},
  author={Zhao, Yifan and Zhang, Tong and Li, Jia and Tian, Yonghong},
  journal={IEEE Transactions on Pattern Analysis and Machine Intelligence},
  volume={45},
  number={10},
  pages={11720--11732},
  year={2023},
  publisher={IEEE}
}

@inproceedings{hu2022pushing,
  title={Pushing the limits of simple pipelines for few-shot learning: External data and fine-tuning make a difference},
  author={Hu, Shell Xu and Li, Da and St{\"u}hmer, Jan and Kim, Minyoung and Hospedales, Timothy M},
  booktitle={Proceedings of the IEEE/CVF Conference on Computer Vision and Pattern Recognition},
  pages={9068--9077},
  year={2022}
}

@article{xu2024enhancing,
  title={Enhancing information maximization with distance-aware contrastive learning for source-free cross-domain few-shot learning},
  author={Xu, Huali and Liu, Li and Zhi, Shuaifeng and Fu, Shaojing and Su, Zhuo and Cheng, Ming-Ming and Liu, Yongxiang},
  journal={IEEE Transactions on Image Processing},
  year={2024},
  publisher={IEEE}
}

@inproceedings{shangguan2025cross,
  title={Cross-domain multi-modal few-shot object detection via rich text},
  author={Shangguan, Zeyu and Seita, Daniel and Rostami, Mohammad},
  booktitle={2025 IEEE/CVF Winter Conference on Applications of Computer Vision (WACV)},
  pages={6570--6580},
  year={2025},
  organization={IEEE}
}

@inproceedings{jing2020cross,
  title={Cross-modal cross-domain moment alignment network for person search},
  author={Jing, Ya and Wang, Wei and Wang, Liang and Tan, Tieniu},
  booktitle={Proceedings of the IEEE/CVF Conference on Computer Vision and Pattern Recognition},
  pages={10678--10686},
  year={2020}
}

@article{ji2022information,
  title={Information symmetry matters: a modal-alternating propagation network for few-shot learning},
  author={Ji, Zhong and Hou, Zhishen and Liu, Xiyao and Pang, Yanwei and Han, Jungong},
  journal={IEEE Transactions on Image Processing},
  volume={31},
  pages={1520--1531},
  year={2022},
  publisher={IEEE}
}

@article{hao2023uncertainty,
  title={Uncertainty-aware alignment network for cross-domain video-text retrieval},
  author={Hao, Xiaoshuai and Zhang, Wanqian},
  journal={Advances in Neural Information Processing Systems},
  volume={36},
  pages={38284--38296},
  year={2023}
}

@inproceedings{wang2022toward,
  title={Toward learning human-aligned cross-domain robust models by countering misaligned features},
  author={Wang, Haohan and Huang, Zeyi and Zhang, Hanlin and Lee, Yong Jae and Xing, Eric P},
  booktitle={Uncertainty in Artificial Intelligence},
  pages={2075--2084},
  year={2022},
  organization={PMLR}
}

@inproceedings{deng2009imagenet,
  title={Imagenet: A large-scale hierarchical image database},
  author={Deng, Jia and Dong, Wei and Socher, Richard and Li, Li-Jia and Li, Kai and Fei-Fei, Li},
  booktitle={2009 IEEE conference on computer vision and pattern recognition},
  pages={248--255},
  year={2009},
  organization={Ieee}
}

@article{zhang2021tip,
  title={Tip-adapter: Training-free clip-adapter for better vision-language modeling},
  author={Zhang, Renrui and Fang, Rongyao and Zhang, Wei and Gao, Peng and Li, Kunchang and Dai, Jifeng and Qiao, Yu and Li, Hongsheng},
  journal={arXiv preprint arXiv:2111.03930},
  year={2021}
}

@inproceedings{tang2024amu,
  title={Amu-tuning: Effective logit bias for clip-based few-shot learning},
  author={Tang, Yuwei and Lin, Zhenyi and Wang, Qilong and Zhu, Pengfei and Hu, Qinghua},
  booktitle={Proceedings of the IEEE/CVF Conference on Computer Vision and Pattern Recognition},
  pages={23323--23333},
  year={2024}
}

@inproceedings{huang2024lp++,
  title={Lp++: A surprisingly strong linear probe for few-shot clip},
  author={Huang, Yunshi and Shakeri, Fereshteh and Dolz, Jose and Boudiaf, Malik and Bahig, Houda and Ben Ayed, Ismail},
  booktitle={Proceedings of the IEEE/CVF Conference on Computer Vision and Pattern Recognition},
  pages={23773--23782},
  year={2024}
}

@inproceedings{li2025logits,
  title={Logits DeConfusion with CLIP for Few-Shot Learning},
  author={Li, Shuo and Liu, Fang and Hao, Zehua and Wang, Xinyi and Li, Lingling and Liu, Xu and Chen, Puhua and Ma, Wenping},
  booktitle={Proceedings of the Computer Vision and Pattern Recognition Conference},
  pages={25411--25421},
  year={2025}
}

@article{tschannen2025siglip,
  title={Siglip 2: Multilingual vision-language encoders with improved semantic understanding, localization, and dense features},
  author={Tschannen, Michael and Gritsenko, Alexey and Wang, Xiao and Naeem, Muhammad Ferjad and Alabdulmohsin, Ibrahim and Parthasarathy, Nikhil and Evans, Talfan and Beyer, Lucas and Xia, Ye and Mustafa, Basil and others},
  journal={arXiv preprint arXiv:2502.14786},
  year={2025}
}

@article{bolya2025perception,
  title={Perception encoder: The best visual embeddings are not at the output of the network},
  author={Bolya, Daniel and Huang, Po-Yao and Sun, Peize and Cho, Jang Hyun and Madotto, Andrea and Wei, Chen and Ma, Tengyu and Zhi, Jiale and Rajasegaran, Jathushan and Rasheed, Hanoona and others},
  journal={arXiv preprint arXiv:2504.13181},
  year={2025}
}

@article{li2020prototypical,
  title={Prototypical contrastive learning of unsupervised representations},
  author={Li, Junnan and Zhou, Pan and Xiong, Caiming and Hoi, Steven CH},
  journal={arXiv preprint arXiv:2005.04966},
  year={2020}
}

@inproceedings{wu2018unsupervised,
  title={Unsupervised feature learning via non-parametric instance discrimination},
  author={Wu, Zhirong and Xiong, Yuanjun and Yu, Stella X and Lin, Dahua},
  booktitle={Proceedings of the IEEE conference on computer vision and pattern recognition},
  pages={3733--3742},
  year={2018}
}

@inproceedings{yang2025clip,
  title={Clip-cid: Efficient clip distillation via cluster-instance discrimination},
  author={Yang, Kaicheng and Gu, Tiancheng and An, Xiang and Jiang, Haiqiang and Dai, Xiangzi and Feng, Ziyong and Cai, Weidong and Deng, Jiankang},
  booktitle={Proceedings of the AAAI Conference on Artificial Intelligence},
  volume={39},
  number={20},
  pages={21974--21982},
  year={2025}
}

@article{geirhos2020shortcut,
  title={Shortcut learning in deep neural networks},
  author={Geirhos, Robert and Jacobsen, J{\"o}rn-Henrik and Michaelis, Claudio and Zemel, Richard and Brendel, Wieland and Bethge, Matthias and Wichmann, Felix A},
  journal={Nature Machine Intelligence},
  volume={2},
  number={11},
  pages={665--673},
  year={2020},
  publisher={Nature Publishing Group UK London}
}

@article{song2024shortcut,
  title={Shortcut learning in in-context learning: A survey},
  author={Song, Rui and Li, Yingji and Shi, Lida and Giunchiglia, Fausto and Xu, Hao},
  journal={arXiv preprint arXiv:2411.02018},
  year={2024}
}

@article{yuan2024llms,
  title={Do llms overcome shortcut learning? an evaluation of shortcut challenges in large language models},
  author={Yuan, Yu and Zhao, Lili and Zhang, Kai and Zheng, Guangting and Liu, Qi},
  journal={arXiv preprint arXiv:2410.13343},
  year={2024}
}

@article{ma2023rectify,
  title={Rectify vit shortcut learning by visual saliency},
  author={Ma, Chong and Zhao, Lin and Chen, Yuzhong and Guo, Lei and Zhang, Tuo and Hu, Xintao and Shen, Dinggang and Jiang, Xi and Liu, Tianming},
  journal={IEEE Transactions on Neural Networks and Learning Systems},
  year={2023},
  publisher={IEEE}
}

@inproceedings{xiao2010sun,
  title={Sun database: Large-scale scene recognition from abbey to zoo},
  author={Xiao, Jianxiong and Hays, James and Ehinger, Krista A and Oliva, Aude and Torralba, Antonio},
  booktitle={2010 IEEE computer society conference on computer vision and pattern recognition},
  pages={3485--3492},
  year={2010},
  organization={IEEE}
}

@article{maji2013fine,
  title={Fine-grained visual classification of aircraft},
  author={Maji, Subhransu and Rahtu, Esa and Kannala, Juho and Blaschko, Matthew and Vedaldi, Andrea},
  journal={arXiv preprint arXiv:1306.5151},
  year={2013}
}

@inproceedings{krause20133d,
  title={3d object representations for fine-grained categorization},
  author={Krause, Jonathan and Stark, Michael and Deng, Jia and Fei-Fei, Li},
  booktitle={Proceedings of the IEEE international conference on computer vision workshops},
  pages={554--561},
  year={2013}
}

@inproceedings{bossard2014food,
  title={Food-101--mining discriminative components with random forests},
  author={Bossard, Lukas and Guillaumin, Matthieu and Van Gool, Luc},
  booktitle={Computer vision--ECCV 2014: 13th European conference, zurich, Switzerland, September 6-12, 2014, proceedings, part VI 13},
  pages={446--461},
  year={2014},
  organization={Springer}
}

@inproceedings{parkhi2012cats,
  title={Cats and dogs},
  author={Parkhi, Omkar M and Vedaldi, Andrea and Zisserman, Andrew and Jawahar, CV},
  booktitle={2012 IEEE conference on computer vision and pattern recognition},
  pages={3498--3505},
  year={2012},
  organization={IEEE}
}

@inproceedings{nilsback2008automated,
  title={Automated flower classification over a large number of classes},
  author={Nilsback, Maria-Elena and Zisserman, Andrew},
  booktitle={2008 Sixth Indian conference on computer vision, graphics \& image processing},
  pages={722--729},
  year={2008},
  organization={IEEE}
}

@inproceedings{fei2004learning,
  title={Learning generative visual models from few training examples: An incremental bayesian approach tested on 101 object categories},
  author={Fei-Fei, Li and Fergus, Rob and Perona, Pietro},
  booktitle={2004 conference on computer vision and pattern recognition workshop},
  pages={178--178},
  year={2004},
  organization={IEEE}
}

@inproceedings{cimpoi2014describing,
  title={Describing textures in the wild},
  author={Cimpoi, Mircea and Maji, Subhransu and Kokkinos, Iasonas and Mohamed, Sammy and Vedaldi, Andrea},
  booktitle={Proceedings of the IEEE conference on computer vision and pattern recognition},
  pages={3606--3613},
  year={2014}
}

@article{soomro2012ucf101,
  title={UCF101: A dataset of 101 human actions classes from videos in the wild},
  author={Soomro, K},
  journal={arXiv preprint arXiv:1212.0402},
  year={2012}
}

@article{ji2020kullback,
  title={Kullback--Leibler divergence metric learning},
  author={Ji, Shuyi and Zhang, Zizhao and Ying, Shihui and Wang, Liejun and Zhao, Xibin and Gao, Yue},
  journal={IEEE transactions on cybernetics},
  volume={52},
  number={4},
  pages={2047--2058},
  year={2020},
  publisher={IEEE}
}

@article{tang2021augmented,
  title={Augmented shortcuts for vision transformers},
  author={Tang, Yehui and Han, Kai and Xu, Chang and Xiao, An and Deng, Yiping and Xu, Chao and Wang, Yunhe},
  journal={Advances in Neural Information Processing Systems},
  volume={34},
  pages={15316--15327},
  year={2021}
}

@inproceedings{li2025advancing,
  title={Advancing textual prompt learning with anchored attributes},
  author={Li, Zheng and Song, Yibing and Cheng, Ming-Ming and Li, Xiang and Yang, Jian},
  booktitle={Proceedings of the IEEE/CVF International Conference on Computer Vision},
  pages={3618--3627},
  year={2025}
}

@inproceedings{bang2024active,
  title={Active Prompt Learning in Vision Language Models},
  author={Bang, Jihwan and Ahn, Sumyeong and Lee, Jae-Gil},
  booktitle={Proceedings of the IEEE/CVF Conference on Computer Vision and Pattern Recognition},
  pages={27004--27014},
  year={2024}
}

@inproceedings{
zehao2025dynaprompt,
title={DynaPrompt: Dynamic Test-Time Prompt Tuning},
author={Xiao, Zehao and Yan, Shilin and Hong, Jack and Cai, Jiayin and Jiang, Xiaolong and Hu, Yao and Shen, Jiayi and Wang, Qi and Snoek, Cees GM},
booktitle={The Thirteenth International Conference on Learning Representations},
year={2025},
}

@inproceedings{yao2024tcp,
  title={Tcp: Textual-based class-aware prompt tuning for visual-language model},
  author={Yao, Hantao and Zhang, Rui and Xu, Changsheng},
  booktitle={Proceedings of the IEEE/CVF Conference on Computer Vision and Pattern Recognition},
  pages={23438--23448},
  year={2024}
}

@article{zhao2025fine,
  title={Fine-Grained VLM Fine-tuning via Latent Hierarchical Adapter Learning},
  author={Zhao, Yumiao and Jiang, Bo and Ding, Yuhe and Wang, Xiao and Tang, Jin and Luo, Bin},
  journal={arXiv preprint arXiv:2508.11176},
  year={2025}
}

@InProceedings{Yang_2024_CVPR,
    author    = {Yang, Lingxiao and Zhang, Ru-Yuan and Wang, Yanchen and Xie, Xiaohua},
    title     = {MMA: Multi-Modal Adapter for Vision-Language Models},
    booktitle = {Proceedings of the IEEE/CVF Conference on Computer Vision and Pattern Recognition (CVPR)},
    month     = {June},
    year      = {2024},
    pages     = {23826-23837}
}

@inproceedings{Hatano2024MMCDFSL,
  author = {Hatano, Masashi and Hachiuma, Ryo and Fujii, Ryo and Saito, Hideo},
  title = {Multimodal Cross-Domain Few-Shot Learning for Egocentric Action Recognition},
  booktitle = {European Conference on Computer Vision (ECCV)},
  year = {2024},
}

@inproceedings{zhang2022metadiff,
	author    = {Zhang, Baoquan and Luo, Chuyao and Yu, Demin and Lin, Huiwei and Li, Xutao and Ye, Yunming and Zhang, Bowen},
	title     = {MetaDiff: Meta-Learning with Conditional Diffusion for Few-Shot Learning},
	booktitle = {AAAI},
	year      = {2024},
}

@InProceedings{Khattak_2023_ICCV,
    author    = {Khattak, Muhammad Uzair and Wasim, Syed Talal and Naseer, Muzammal and Khan, Salman and Yang, Ming-Hsuan and Khan, Fahad Shahbaz},
    title     = {Self-regulating Prompts: Foundational Model Adaptation without Forgetting},
    booktitle = {Proceedings of the IEEE/CVF International Conference on Computer Vision (ICCV)},
    month     = {October},
    year      = {2023},
    pages     = {15190-15200}
}

@inproceedings{bai2024prompt,
  title={Prompt-based distribution alignment for unsupervised domain adaptation},
  author={Bai, Shuanghao and Zhang, Min and Zhou, Wanqi and Huang, Siteng and Luan, Zhirong and Wang, Donglin and Chen, Badong},
  booktitle={Proceedings of the AAAI conference on artificial intelligence},
  volume={38},
  number={2},
  pages={729--737},
  year={2024}
}
